\documentclass{article}

\usepackage{amsmath,amsthm,amssymb,amsfonts}
\usepackage{graphicx}
\usepackage{arxiv}

\usepackage[utf8]{inputenc} 
\usepackage[T1]{fontenc}    
\usepackage{hyperref}       
\usepackage{url}            
\usepackage{booktabs}       
\usepackage{amsfonts}       
\usepackage{nicefrac}       
\usepackage{microtype}      
\usepackage{lipsum}

\usepackage{xcolor}         
\usepackage{multirow}
\usepackage{caption}
\usepackage{subcaption}
\usepackage{graphicx}
\usepackage{amsfonts, amsmath, amssymb} 
\usepackage{algorithm}
\usepackage{algorithmic}

\usepackage{wrapfig}
\usepackage{tikz}
\usepackage{array}

\newtheorem{definition}{Definition}
\newtheorem{theorem}{Theorem}
\newtheorem{lemma}[theorem]{Lemma}

\newcommand{\modelname}{\textbf{CG-FedLLM}}
\newcommand{\stageone}{\textbf{TGAP}}
\newcommand{\stagetwo}{\textbf{FAF}}
\newcommand{\auto}{\mathcal{V}}

\title{CG-FedLLM: How to Compress Gradients in Federated Fune-tuning for Large Language Models \\ Extended Version}

\author{
  Huiwen Wu \\
  Zhejiang Laboratory\\
  Hangzhou, Zhejiang, China \\
  \texttt{huiwen0820@outlook.com} \\
   \And
    Xiaogang Xu~ \thanks{corresponding author} \\
    The Chinese University of Hong Kong.\\
    Hong Kong, China \\
    \texttt{xiaogangxu00@gmail.com} \\
   \And
    Deyi Zhang \\
    Zhejiang Laboratory\\
    Hangzhou, Zhejiang, China \\
    \texttt{xiaohan@zhejianglab.com} \\
   \And
    Xiaohan Li \\
    Zhejiang Laboratory\\
    Hangzhou, Zhejiang, China \\
    \texttt{xiaohan@zhejianglab.com} \\
   \And
    Jiafei Wu~ \thanks{corresponding author} \\
    Zhejiang Laboratory\\
    Hangzhou, Zhejiang, China \\
    \texttt{wujiafei@zhejianglab.com} \\
   \And
    Zhe Liu \\
    Zhejiang Laboratory\\
    Hangzhou, Zhejiang, China \\
    \texttt{zhe.liu@zhejianglab.com} \\
}

\begin{document}
\maketitle
\footnotetext{This paper has been published in the Proceedings of the 26th European Conference on Artificial Intelligence (ECAI 2025).}

\begin{abstract}
The success of current Large-Language Models (LLMs) hinges on extensive training data that is collected and stored centrally, called Centralized Learning (CL).
However, such a collection manner poses a privacy threat, and one potential solution is Federated Learning (FL), which transfers gradients, not raw data, among clients.
Unlike traditional networks, FL for LLMs incurs significant communication costs due to their tremendous parameters. In this study, we introduce an innovative approach to compress gradients to improve communication efficiency during LLM FL, formulating the new FL pipeline named \modelname. 
This approach integrates an encoder on the client side to acquire the compressed gradient features and a decoder on the server side to reconstruct the gradients. We also develop a novel training strategy that comprises Temporal-ensemble Gradient-Aware Pre-training (\stageone)~to identify characteristic gradients of the target model and Federated AutoEncoder-Involved Fine-tuning (\stagetwo)~to compress gradients adaptively. Extensive experiments confirm that our approach reduces communication costs and improves performance (e.g., average \textbf{3} points increment compared with traditional CL- and FL-based fine-tuning with \textbf{LlaMA} on a well-recognized benchmark, C-Eval). This is because our encoder-decoder, trained via \stageone~and \stagetwo,~can filter gradients while selectively preserving critical features. Furthermore, we present a series of experimental analyses that focus on the signal-to-noise ratio, compression rate, and robustness within this privacy-centric framework, providing insight into the development of more efficient and secure LLMs.
\end{abstract}

\begin{figure*}[h]
     \begin{subfigure}[b]{0.33\textwidth}
         \centering
         \includegraphics[width=\textwidth]{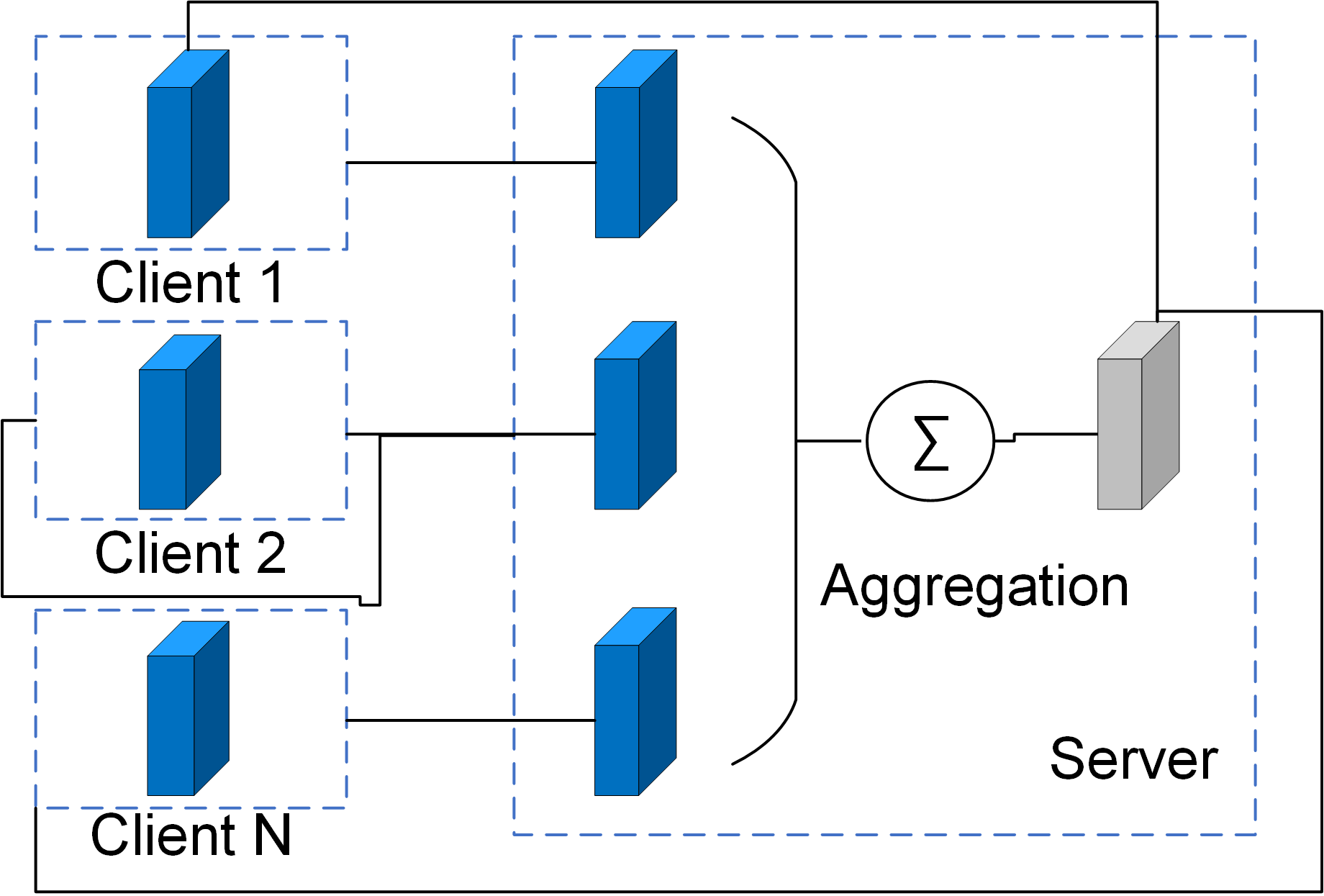}
         \caption{\textbf{Plain-FedLLM}}
         \label{fig:plain_fedllm}
     \end{subfigure}
    \begin{subfigure}[b]{0.33\textwidth}
         \centering
         \includegraphics[width=\textwidth]{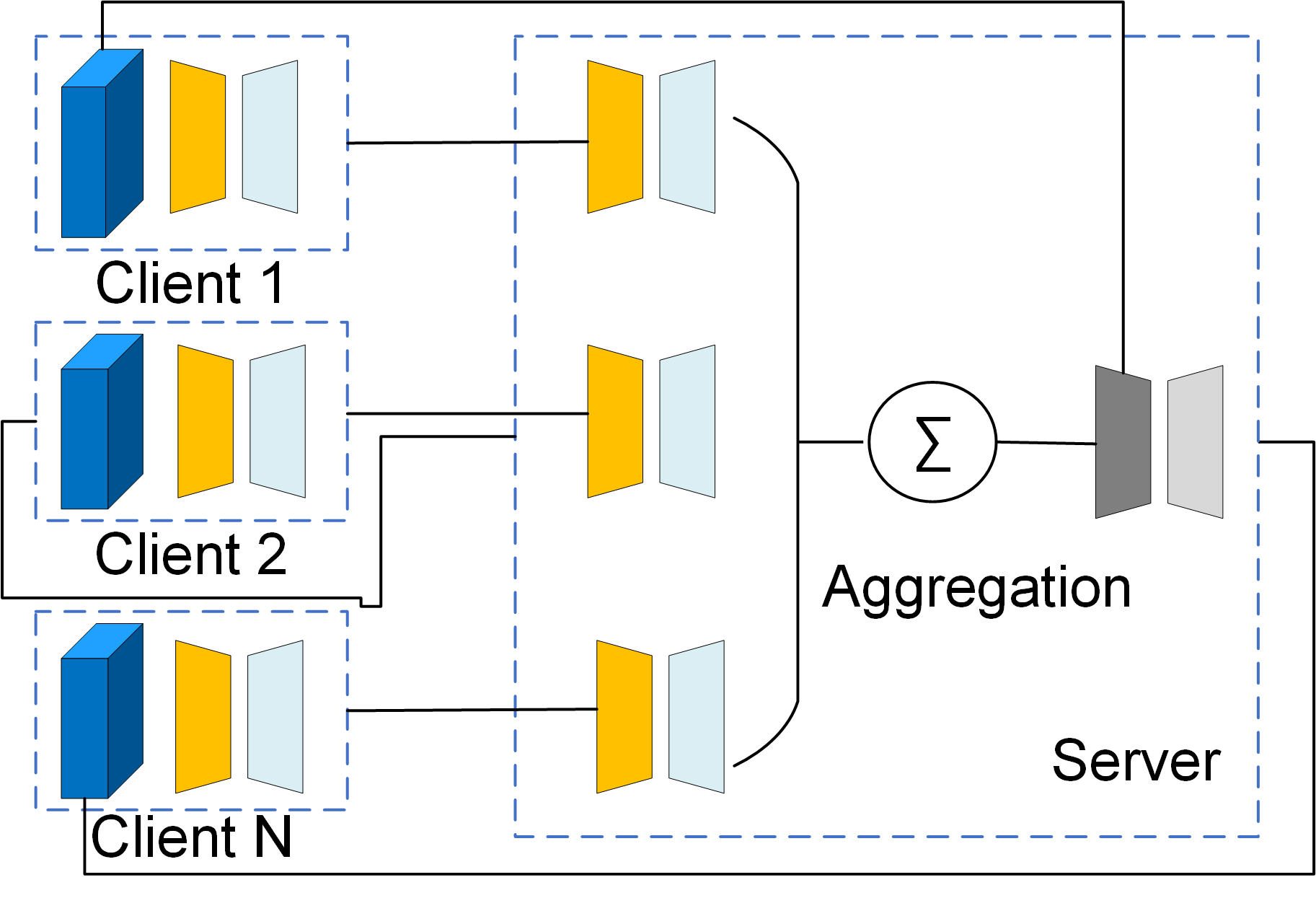}
         \caption{\textbf{LoRA-FedLLM}}
         \label{fig:lora_fedllm}
     \end{subfigure}
     \begin{subfigure}[b]{0.33\textwidth}
         \centering
         \includegraphics[width=\textwidth]{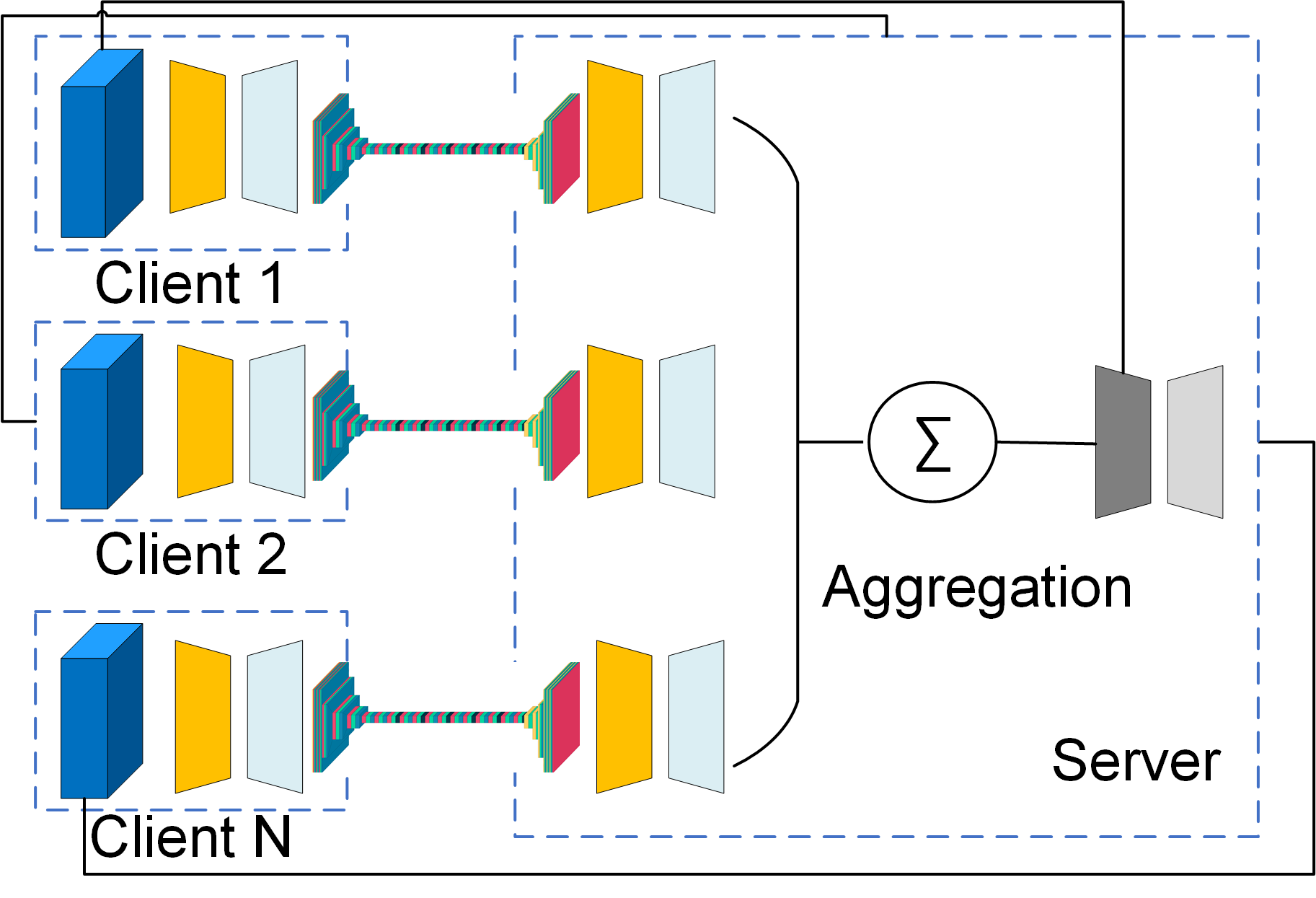}
         \caption{\modelname~}
         \label{fig:cg_fedllm}
     \end{subfigure}
\vspace{0.1cm}
\caption{Design Variants of FL Approaches for LLMs.}
\label{fig:compare}
\end{figure*}

\section{Introduction}

The development of Large Language Models (LLMs) has been on the rise, especially since the introduction of ChatGPT~\cite{lund2023chatting}. These models have shown strong capabilities in natural language understanding~\cite{karanikolas2023large}, common sense reasoning~\cite{sun2024benchmarking,wang2024can}, and even mathematics reasoning~\cite{imani2023mathprompter}. However, their practical application remains challenging due to inefficiency and privacy issues in training strategies and data management.
The large number of parameters in LLMs makes it impractical to manage the entire parameter set on a standard low-cost GPU, such as the RTX 4090, even for models like LLaMa-7B~\cite{touvron2023llama}. Furthermore, finetuning LLMs requires substantial amounts of high-quality data, while data privacy laws such as the General Data Protection Regulation (GDPR)~\cite{voigt2017eu} and the California Consumer Privacy Act (CCPA)~\cite{de2018guide} restrict data sharing across platforms.
\textbf{Therefore, there is an immediate need to establish a strategy to efficiently and privately finetune the LLM while ensuring the performance of the finetuned model.}

In Federated Learning (FL), clients perform local stochastic gradient descent on their datasets and transmit gradients or incremental model updates to a central server, as shown in Fig.~\ref{fig:plain_fedllm}.
This server aggregates these local gradients and redistributes the averaged gradients to the clients, thus incurring communication costs between local and global entities in FL.
The communication costs of applying FL to LLMs could be prohibitive due to the contradiction between limited bandwidth and extensive parameters.
Even if we employ LoRA (Low-Rank Adaptation) to finetune partial layers (as displayed in Fig.~\ref{fig:lora_fedllm}). 
Consequently, enhancing communication efficiency requires further compression. However, typical strategies, such as predefined quantization pipelines, inevitably lead to information loss. Thus, selecting an appropriate compression technique is crucial to ensure a satisfactory compression ratio and minimal distortion in transmitted gradients.

In this paper, we introduce a new method for compressing gradients called Compress Gradients Federated Finetuning of LLM (\textbf{CG-FedLLM}). In our model, we use an AutoEncoder: clients compress their gradients into compact features using the encoder $\mathbf{Enc}$, and the server reconstructs these features into accurate gradients using the decoder $\mathbf{Dec}$, as shown in Fig.~\ref{fig:cg_fedllm}. 

A direct approach is to train the AutoEncoder alongside the LLM to ensure compression accuracy. However, this method incurs substantial additional training costs. Furthermore, the \textit{dynamic} nature of the training samples, due to the varying gradients of the target model during training, complicates the fitting process.
Therefore, we identify the characteristics of the target model's gradients and design a time-ensemble pretraining scheme to train $\mathbf{Enc}$ and $\mathbf{Dec}$, called \stageone.
Observations show that local gradients before FL are consistent with those during FL, as the target model remains unchanged.
Therefore, by recording the gradients of the target model in various training stages, we can create a \textit{consistent and homogeneous} data set $\mathcal{A}$ to train our AutoEncoder effectively.
This data set can be compiled through the cooperation of various clients before FL by aggregating their local gradients.
Empirical evidence suggests that by pretraining $\mathbf{Enc}$ and $\mathbf{Dec}$ with such a gradient data set before FL, the AutoEncoder can efficiently reconstruct gradients during FL without the need for further training.
The additional computational cost is tiny with the details in Sect.~\ref{sec:cost}. 

In addition to efficient communication, extensive experiments on established LLM benchmarks show that finetuning LLMs with our \modelname\ leads to better performance than traditional FL and centralized learning (CL). The trained AutoEncoder can accurately reconstruct functional gradients for training and correct abnormal gradients with different distributions. This ability is derived from the inherent nature of deep neural networks in adapting abnormal signals~\cite{ulyanov2018deep}.
Moreover, our \modelname~is compatible with differential privacy mechanisms, which enhances privacy protection. We have empirically demonstrated that the encoder can filter out noise added to the gradient and reconstructed by the decoder.

To summarize, the main goal of this paper is to develop a learning-based gradient compression method to enhance communication efficiency and privacy guarantee for federated finetuning (\textbf{FedLLM}). In order to accomplish this goal, (1) we develop a new federated finetuning framework, \modelname, which incorporates an AutoEncoder into FL to enhance communication efficiency significantly. (2) We design \stageone~to pre-train the AutoEncoder before FL implementation to ensure that the AutoEncoder performs well without incurring additional training costs in \stagetwo. (3) We conduct extensive evaluations, including the multilevel C-Eval and MMLU benchmarks, to confirm that the model finetuned with \modelname~performs better than traditional FL and CL.
This paper is an extended version of our earlier work ‘CG-FedLLM: How to Compress Gradients in Federated Fine-Tuning for Large Language Models,’~\cite{Wu2025CGFedLLM} with significant enhancements including extensive experiments across diverse LLMs and datasets, 
extensive evaluations via diverse metrics, 
and detailed ablation studies.


\section{Related Work}

\noindent{\textbf{Parameter-Efficient Fine-Tuning (PEFT).}}
PEFT methods tackle the challenges of memory and computational demands when adapting extensive language models for specific applications~\cite{fu2023effectiveness,liu2022few,ding2023parameter}.
Popular techniques in PEFT involve the adjustment of LLM parameters within a confined subspace of low rank, such as LoRA~\cite{hu2021lora}, achieving considerable savings in resources while preserving efficacy.
PEPT\cite{houlsby2019parameter} mitigates elevated communication expenses by fine-tuning a minor segment of the overall parameters in conventional NLP tasks.
DoRA~\cite{liu2024dora} divides the initial model weights into magnitude and direction to facilitate fine-tuning, employing LoRA to alter the aspect of direction.
QLoRA~\cite{dettmers2024qlora} shows that fully precision adapters can be efficiently trained on lower precision models while ensuring sustained performance.
Although these methods significantly reduce parameter updates, communication costs remain considerable for LLMs in FL with a large number of clients and limited bandwidth.

\begin{figure*}[t!]
    \centering
    \includegraphics[width=0.95\textwidth]{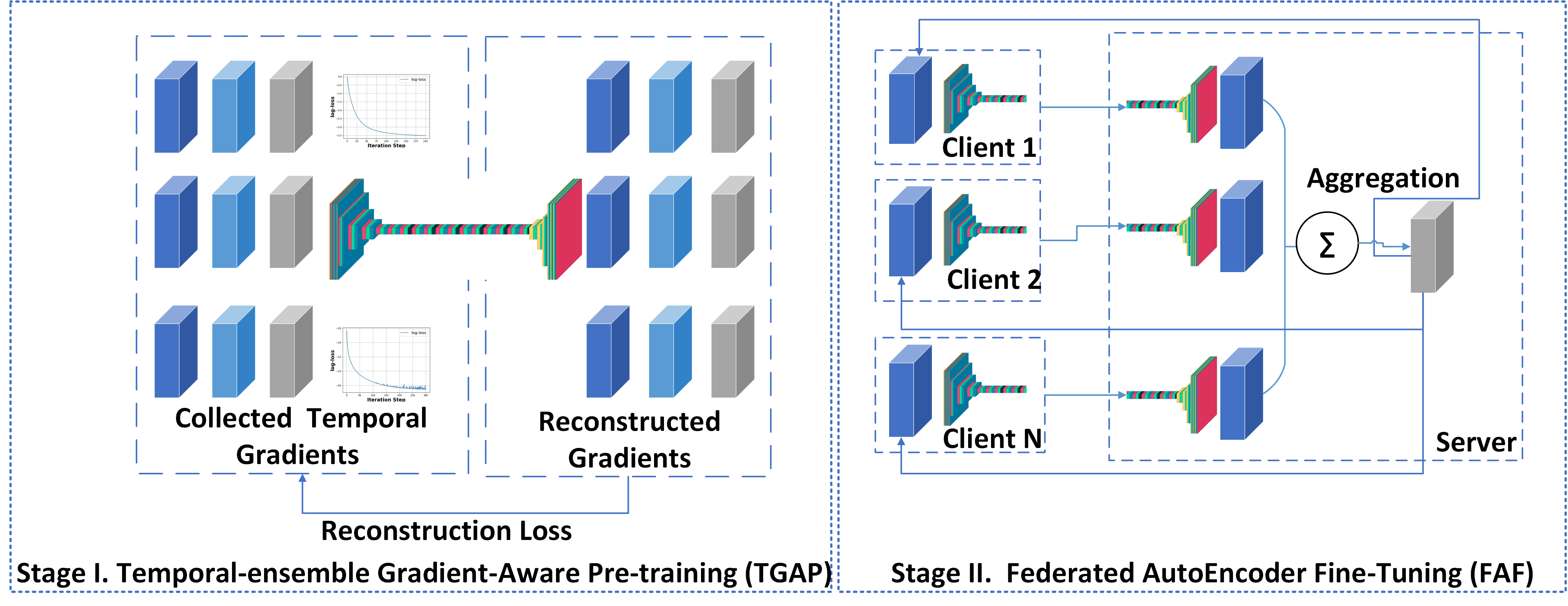}
    \vspace{0.1cm}
    \caption{\modelname: Compress Gradients in Federated Fune-tuning for Large Language Models. The AutoEncoder $\auto$ involved in \modelname~is completed by \stageone~(left), and the illustration of utilizing $\auto$ in FL is implemented by \stagetwo~(right).}
    \label{fig:cgfedgpt}
    \vspace{0.35cm}
\end{figure*}

\noindent{\textbf{Federated Learning (FL).}}
FL is a general training scheme to collaboratively train a deep learning model without sharing raw data~\cite{mcmahan2017communication,wu2020theoretical,li2020review}.
In the FL setting, the client train its own model with a local dataset and send the increment gradients to the server. The server then aggregates the partial gradients and synchronizes with the local clients. 
Existing approaches to ensure privacy in FL including differential privacy~\cite{wei2020federated,truex2020ldp}, secret sharing~\cite{rathee2024mpc,mansouri2023sok}, and homomorphic encryption~\cite{rovida2024transformer,fang2021privacy}. 
However, differential privacy causes the degradation of FL training due to the added random noise. 
Secret sharing exerts an additional cost of communication while homomorphic encryption increases the computation complexity trendingly due to the encryption and decryption. 

Besides the privacy guarantee, another important issue in adapting FL for LLMs is the communication bottleneck~\cite{chen2021communication,tang2024z}.
Although some researchers attempt to learn a gradient compression model with CNN networks of 1 dimension, these networks does not scale to LLM due to the huge size of model parameters~\cite{li2019end,abrahamyan2021learned}.

Recently, several initiatives have focused on training and fine-tuning LLMs using federated learning principles. 
For example, FederatedScope-LLM~\cite{kuang2024federatedscope} offers a comprehensive package, while OpenFedLLM~\cite{ye2024openfedllm} provides a concise and research-friendly framework. TITANIC~\cite{su2024titanic} allows for privacy-preserving, on-device fine-tuning, and FedBiOT~\cite{wu2024fedbiot} enables local fine-tuning without sharing the entire model. 
However, these methods mainly address training paradigms or local fine-tuning and often overlook communication bottlenecks and privacy challenges in FedLLMs.

\section{Methodology}

\subsection{Overview of FedLLM}
\label{sec:over}

\begin{table}
  \caption{Compression Ratio (CR) and Signal Noise Ratio (SNR) with Various AutoEncoder Architectures.}
  \label{tab:ratio_compress}
  \centering
  \begin{tabular}{c|c|c|c}
    \toprule 
 & \textbf{1-D CNN~\cite{abrahamyan2021learned}} & \textbf{ResNet}~\cite{he2016deep}  &  \textbf{U-Former}~\cite{wang2022uformer} \\
$\mathbf{G}$ and $\tilde{\mathbf{G}}$ &  $[128, 1, 32768]$ &{$[1, 4096, 2048]$} & $[8, 1024, 1024]$\\
$\mathbf{Enc}({\mathbf{G}})$ &  $[128, 4, 256]$ & $[64, 64, 32]$  &  $[1, 256, 512]$ \\
CR  & 3.21 \% &  1.56 \% &  1.56 \% \\
\midrule
$\|\mathbf{G} \|^2_2$ &  $14.29$ &  $14.29$ &   $14.29$  \\
$\| \mathbf{G} - \tilde{\mathbf{G}}  \|^2_2 $ & $8.94 \cdot 10^{-4}$ & $5.06\cdot 10^{-12}$ & $1.04 \cdot 10^{-11}$  \\
SNR & $1.60 \cdot 10^{4} $ &  $ 2.82 \cdot 10^{12}$ &   $ 1.37 \cdot 10^{12}$ \\
\bottomrule
\end{tabular}
\end{table}

Given a target LLM $\mathbf{M}$, there are $N$ clients $\mathbf{C}_i$ and a global server $\mathbf{S}$ in FL. 
Specifically, each client $\mathbf{C}_i$ will have its training dataset $\mathcal{D}_i$ and a copy of model $\mathbf{W}_{i}$, and training $\mathbf{W}_{i}$ with $\mathcal{D}_i$ can result in a set of gradients to update the model, as $\mathbf{G}_i$.
Gradients $\mathbf{G}_{i=1,...,N}$ will be propagated on the server $\mathbf{S}$ for ensemble, as $\tilde{\mathbf{G}}_{i=1,...,N}=\mathbf{T}_i(\mathbf{G}_{i=1,...,N})$, where $\mathbf{T}_i$ is the transmission process for the $i$-th client. $\tilde{\mathbf{G}}_{i=1,...,N}$ can be combined to produce the polymerized gradient $\mathbf{G}$ to update the central model weight, as $\mathbf{W}_{\rm{global}}$. 
In addition, the global update can be delivered to the selected clients $I_{t}$ in the iteration step $t$, updating the weight of each client with $\mathbf{W}_{i}$. This is a cyclic process until the training is completed.
In traditional FL, $\mathbf{W}$ and $\mathbf{W}_{i=1,..., N}$ are small in size, resulting in a low communication cost in $\mathbf{T}_{i=1,..., N}$.
However, this burden becomes overwhelming when FL is applied to LLMs due to the large number of parameters that require fine-tuning. Even with techniques like LoRA, transmitting numerous transformer layers remains challenging for typical LLMs.

In our \modelname, we integrate an adaptive compression model $\auto$~into the FL with the AutoEncoder structure, consisting of the encoder $\mathbf{Enc}$ and the decoder $\mathbf{Dec}$.
To transmit $\mathbf{G}_{i=1,..., N}$ (the gradient can be first compressed by other methods like Low-rank decomposition in \textbf{LoRA}), it will be first transferred into a feature with a tiny size, as 
\begin{equation}
    \mathbf{F}_{i=1,...,N}=\mathbf{Enc}(\mathbf{G}_{i=1,...,N}).
    \label{eq:encoder}
\end{equation} 
The server side can receive these features efficiently and decode them into the desired gradients as 
\begin{equation}
    \tilde{\mathbf{G}}_{i=1,...,N}=\mathbf{Dec}(\mathbf{F}_{i=1,...,N}).
        \label{eq:decoder}
\end{equation}
The differences between \modelname~and traditional FL can be seen in Figure~\ref{fig:compare}.
Training of $\auto$ is completed by our designed Temporal-ensemble Gradient-Aware Pre-training (\stageone)~before FL, and FL with $\auto$ is implemented via our designed Federated AutoEncoder Fine-Tuning (\stagetwo)~algorithm, which will be described in the following two sections.

\subsection{TGAP}
\label{sec:tgap}



To ensure compression accuracy, a straightforward approach is to simultaneously train the AutoEncoder $\auto$ with the LLM. However, this strategy incurs significant additional training costs. Furthermore, the AutoEncoder training set is \textit{dynamic}, as the gradients of the target model continuously evolve during the training process. This variability complicates fitting, as newly acquired knowledge can disrupt previously established patterns. Empirical evidences also indicate that existing continuous learning approaches can lead to failures. To avoid the dynamic training of the AutoEncoder, we utilize a pre-training strategy, consisting the following steps. 

\subsubsection{Construction of Gradients Data}
We created two distinct datasets with similar domain knowledge for two training stages of \modelname.
The first $\mathcal{D}_1$ is utilized for the pre-training of FedLLM to collect the intermediate gradients as inputs for \stageone. The second $\mathcal{D}_2$ is employed for \stagetwo. 
During the first training procedure with dataset $\mathcal{D}_1$, we can collect the gradient sequences as $\mathbf{G} = 
\begin{bmatrix} \mathbf{G}_{i}^t \end{bmatrix}_{i \in [N], t \in [T]}$, where $i$ is the client index, and $t$ is the iteration step to record.
The distributions of the gradients from varying time steps are also similar, resulting in the homogeneous property for the set of $\mathbf{G}_{i}^t, i \in [N], t \in [T]$. 
Training $\auto$ with $\mathbf{G}$ avoids the suboptimal issues caused by dynamic training data while effectively capturing the characteristics of the target model's gradients.
It is empirically validated that $\auto$ with such pre-training before FL, can have satisfied reconstruction ability during FL.
We present the detailed examples of $\mathcal{D}_1$ and $\mathcal{D}_2$ in Sect.~\ref{sec:generalization}.

\subsubsection{Structure and Training of AutoEnocder}

The AutoEncoder, comprising $\mathbf{Enc}$ and $\mathbf{Dec}$, processes 2D features and can incorporate various structures, e.g., ResNet~\cite{targ2016resnet} and UFormer~\cite{wang2022uformer}.
The encoder compresses the initial input into a compact latent representation, while the decoder uses this representation to reconstruct the original input, as shown in Eqs.~\eqref{eq:encoder} and \eqref{eq:decoder}.
Our main goal is to explore how to train satisfied $\mathbf{Enc}$ and $\mathbf{Dec}$, without restrictions on network types. 
The training of $\auto$ is completed by measuring the error between the inputs and outputs, as 
\begin{equation}
    \mathcal{L}_{\rm{recons}} =  
    \| \mathbf{G} - \tilde{\mathbf{G}} \|_2^2.
\end{equation}
The optimizer we adopt in \stageone~is Adam~\cite{kingma2014adam} and the learning rate of $2 \times 10^{-4}$, detailed in the Supplementary Material, Sect.B.~\cite{wu2024cg}

\begin{algorithm}[t!]
\caption{Federated AutoEncoder Fine-tuning~(\stagetwo)}
\label{alg:CG_FedLLM}
\textbf{Input: }
number of all clients $N$, 
local partitioned clients ratio $\alpha$, 
foundational model $\mathbf{W}_{\rm{base}}$, 
step size $\eta$,
pre-trained encoder $\rm{Enc}$, 
pre-trained decoder $\rm{Dec}$, 
number of communication rounds $T$; \\
\textbf{Output: }
federated fine-tuned model $\mathbf{W}^T$; \\
\textbf{Initialize with foundational model:} 
$\mathbf{W}^{0}_{\rm{global}} = \mathbf{W}_{\rm{base}}$ for all $i \in [1,2,\cdots,N]$\\
\vspace{-0.2in}
\begin{algorithmic}
\FOR{$t := 0$ to $T-1$}{
	\STATE Server samples a client subset $I_t$ with sampling ratio $\alpha$ \\
	\STATE Server sends $\mathbf{W}^{t}_{\rm{global}}$ to all selected clients \\
	\STATE \textbf{Local Training:} \\
	\FOR{each chosen client $i \in I_t$}{
		\STATE Each client perform LoRA training locally and obtain 
      $\mathbf{G}_{i}^t = \mathbf{B_i A_i}$; \\  
        \STATE Each client performs the gradient compression with the pre-trained encoder \vspace{-0.1in} $$[\bar{\mathbf{A}}_i, \bar{\mathbf{B}}_i] = \mathbf{Enc} [\mathbf{A}_i, \mathbf{B}_i];$$ \\
        \vspace{-0.1in} 
    \STATE Each client sends the compressed local gradients $[\bar{\mathbf{A}}_i, \bar{\mathbf{B}}_i]$ to the server;}
 \ENDFOR
	\STATE \textbf{Server Aggregation:}\\
	\STATE Server collects the local compressed low rank gradients
    $[\bar{\mathbf{A}}_i, \bar{\mathbf{B}}_i]$;
    \STATE Server decompresses local compressed gradients
    \vspace{-0.1in}
    $$[\tilde{\mathbf{A}}_i, \tilde{\mathbf{B}}_i] = \mathbf{Dec}[\bar{\mathbf{A}}_i, \bar{\mathbf{B}}_i];$$ \\
    \vspace{-0.1in}
	\STATE Server aggregates in the \textbf{LoRA} subspace
    \vspace{-0.1in}
	$$[\tilde{\mathbf{A}}, \tilde{\mathbf{B}}] = [\sum_i \tilde{\mathbf{A}}_i, \sum_i \tilde{\mathbf{B}}_i];$$\\
    \vspace{-0.2in}
    \STATE Server updates the global model
    \vspace{-0.1in}
    $$\mathbf{W}_{\rm{global}}^{t+1} = \mathbf{W}_{\rm{global}}^t + \eta \tilde{\mathbf{G}}^t = \mathbf{W}_{\rm{global}}^t +\eta \mathbf{\tilde{B} \tilde{A}}.$$
\vspace{-0.2in}
	}
\ENDFOR

\end{algorithmic}
\end{algorithm}

\subsubsection{Communication Efficiency}
During the fine-tuning phase of the LLaMA-7B foundation model, even after employing low-rank decomposition, the challenge of transmitting large model parameters persists. For example, with a chosen rank number of $8$, the parameters to be transmitted in federated fine-tuning is completely $4096 \times 8 \times 2 \times 4 \times 32 = 8,388,608$. When the number of clients $N$ is large, the computational efficiency is limited by the communication bandwidth. In the second phase of our proposed approach (\modelname), we utilize the AutoEncoder developed in Stage I (\stageone) to reduce the communication burden during federated fine-tuning, as only the encoder features $\mathbf{Enc}({\mathbf{G}})$ need to be propagated.
Our compression strategy offers advantages over previous methods that use neural networks to reduce communication costs, such as the 1-D CNN described in~\cite{abrahamyan2021learned}. As detailed in Table~\ref{tab:ratio_compress}, we analyze the compression ratio by comparing the parameter sizes before encoding, after encoding, and after decoding in three different AutoEncoders. 
The compression ratio is calculated based on the proportion of the total parameters in the encoder’s output relative to its input. A lower compression ratio signifies improved communication efficiency in federated settings. Through a detailed analysis, we found that AutoEncoders using ResNet and U-Former both achieve a compression ratio of $1.56\%$, compared to the $3.21\%$ achieved by the 1-D CNN. This demonstrates that the 2-D architectures, ResNet and U-Former, are more effective in reducing communication overhead during federated fine-tuning for LLMs.

\subsection{FAF}
\label{sec:faf}


This section outlines the federated fine-tuning pipeline involving $\auto$. 
The global model is initially set up as the foundational model already specified. During iteration $t$, following the sampling strategy, the server selects a subset of clients $I_t$ according to the sampling ratio $\alpha$ and distributes the global model $\mathbf{W}^t_{\rm{global}}$ to these clients. Subsequently, we split the process into two phases. The initial phase occurs on the client side, where each selected client conducts LoRA training locally and updates the local model parameters in a low-rank format, specifically $ \mathbf{G}_{i}^t = \mathbf{B}_i \mathbf{A}_i, i\in I_t$. Following this, each client compresses the low-rank matrices using a pre-trained encoder, $[\bar{\mathbf{A}}_i, \bar{\mathbf{B}}_i] = \mathbf{Enc} [\mathbf{A}_i, \mathbf{B}_i]$, to minimize the uplink communication overhead. These compressed local gradients $[\bar{\mathbf{A}}_i, \bar{\mathbf{B}}_i]$ are then transmitted to the server. On the server side, it gathers the compressed data $[\bar{\mathbf{A}}_i, \bar{\mathbf{B}}_i]$ sent by the clients. The server processes these matrices independently using the pre-trained decoder to revert them to their original forms, denoted as $[\tilde{\mathbf{A}}_i, \tilde{\mathbf{B}}_i] = \mathbf{Dec}[\bar{\mathbf{A}}_i, \bar{\mathbf{B}}_i]$. After decompression, an aggregation of the local matrices within the subspace is performed, represented as $[\tilde{\mathbf{A}}, \tilde{\mathbf{B}}] = [\sum_i \tilde{\mathbf{A}}_i, \sum_i \tilde{\mathbf{B}}_i]$. Ultimately, the global model parameters are refreshed using these decompressed and aggregated low-rank matrices $[\tilde{\mathbf{A}}, \tilde{\mathbf{B}}]$.
The updated global model then send the update of the model to the clients, and this procedure between the clients and the server will be cyclic until convergence. The overall procedure is summarized in Algorithm~\ref{alg:CG_FedLLM}.

\subsection{Comprehensive Security within CG-FedLLM}
\label{sec:adv}
Beyond communication efficiency, our \modelname~also provides enhanced privacy protection. Even within FL contexts, studies~\cite{NEURIPS2019_60a6c400,geiping2020inverting} have shown that local gradients shared during federated aggregation can be used to reconstruct the original fine-tuning data. In our FL system, we assume the server is honest, while the client is semi-honest and may be curious about other clients' gradients. The only potential information leakage occurs during gradient transmission between clients and the server, where a robust security guarantee is essential.

After completing the first stage, \stageone, we split the encoder and decoder between the clients and the server. The data transmission involves encrypted gradients, which are compressed by the AutoEncoder. Moreover, since only the server holds the decoder, this setup ensures that neither attackers during data transmission nor other clients can recover the transmitted gradients.

Moreover, we apply differential privacy to local clients' gradients before transmission to further enhance local data privacy. The privacy guarantee and utility preservation provided by \modelname~are discussed in Sect.~\ref{sec:privacy}. Additionally, to ensure information security during the downlink process when the server transmits the aggregated gradients to the clients, clients are equipped with both the encoder and decoder. Consequently, the only data transmitted during both the uplink and downlink procedures are the encrypted gradients.

\begin{figure*}[t!]
     \begin{subfigure}[b]{0.245\textwidth}
         \centering
         \includegraphics[width=\textwidth]{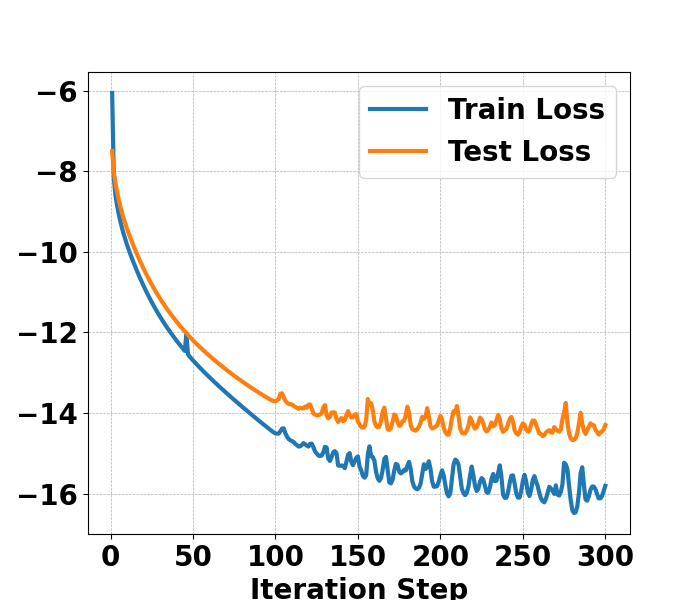}
         \label{fig:ae_conv1d}
     \end{subfigure}
    \begin{subfigure}[b]{0.245\textwidth}
         \centering
         \includegraphics[width=\textwidth]{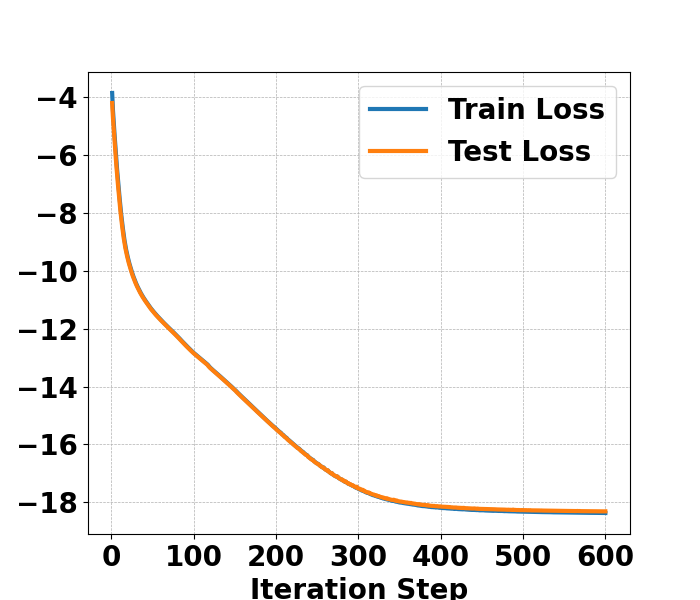}
         \label{fig:ae_uformer0}
     \end{subfigure}
     \begin{subfigure}[b]{0.245\textwidth}
         \centering
         \includegraphics[width=\textwidth]{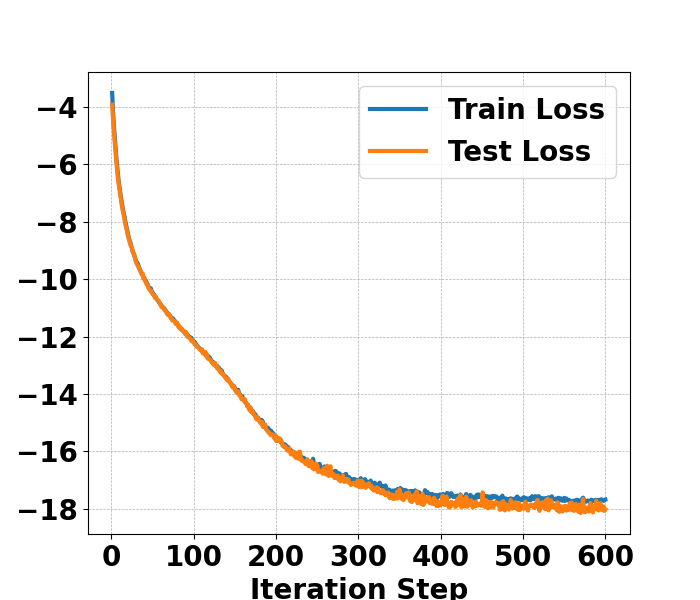}
         \label{fig:ae_uformer1}
     \end{subfigure}
     \begin{subfigure}[b]{0.245\textwidth}
         \centering
         \includegraphics[width=\textwidth]{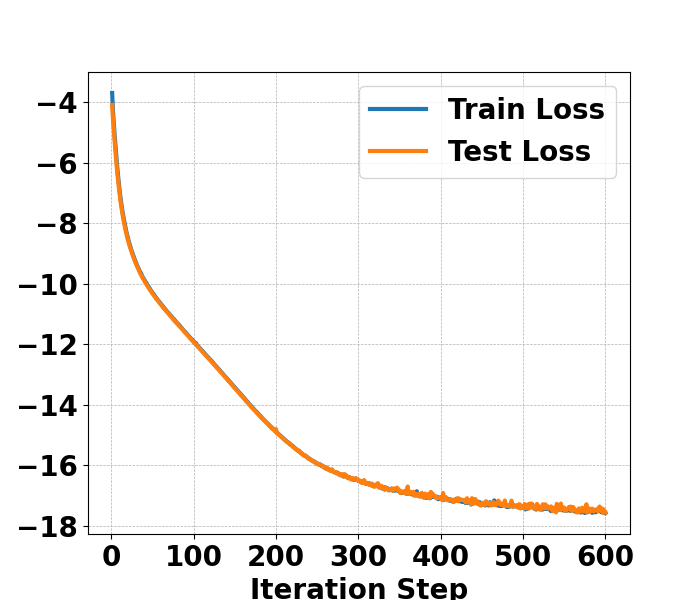}
         \label{fig:ae_uformer2}
     \end{subfigure}
\caption{Convergence of training and test loss in \stageone~using different architectures: 1D-CNN (left), zero-block U-former, one-block U-former, and two-block U-former (right). The x-axis shows iteration steps; the y-axis shows log MSE reconstruction loss. Blue and yellow lines denote training and test loss, respectively.
}
\label{fig:auto_encoder_conv}
\vspace{0.3cm}
\end{figure*}

\section{Experiments}

\label{sec:exp}

In this section, we conduct comprehensive experiments to address the following research questions.

\begin{itemize}
\item{\textbf{RQ1}} How do the proposed methods \modelname~improve the communication efficiency in FedLLM? 
\item{\textbf{RQ2}} Does the additional compression procedure cause performance degradation in LLM benchmark tests?
\item{\textbf{RQ3}} What is the advantage of the proposed method in the privacy-preserving training scenario? 
\item{\textbf{RQ4}} How does the architecture of the Auto-Encoder influence the overall FedLLM fine-tuning? 
\item{\textbf{RQ5}} Does an Auto-Encoder trained on a dataset compress gradients generated from an unseen dataset? 
\end{itemize}

\begin{table*}[t!]
\centering
\caption{Evaluation based on the C-Eval benchmark~(Rows 1 to 7) and 
the MMLU benchmark~(Rows 8 to 10). 
The C-Eval benchmark encompasses 6 subjects, while the MMLU benchmark omits the `Avg(hard)'.}
\label{tab:C-eval Results}
\begin{tabular}{cccccccc} 
\toprule 
 & Methods & Stem & Social Sciences & Humanities & Others & Average & Avg(hard) \\
\midrule 
\multirow{4}{*}{1} & 
\textbf{Cent-LLaMA} & 24.5 & 25.6 & 25.5 & 24.4 & 24.9 & 23.4 \\ 
&\textbf{Compress-FT-LLaMA} (ours) & \textbf{26.8} & 26 & \textbf{26.8} & \textbf{26.5} & \textbf{26.6} & \textbf{26.9} \\
&\textbf{LoRA-FT-LLaMA} & 25.9 & \textbf{27.6} & 25.2 & 24.5 & 25.8 & 24.8 \\
&\textbf{Base-LLaMA} & 21.6 & 23.4 & 23.9 & 23.3 & 22.8 & 20.3 \\ 
\midrule  
\multirow{4}{*}{2} & 
\textbf{Cent-Alpaca} & 25.9 & 24.8 & \textbf{27.3} & 24.5 & 25.7 & 25.7 \\
& \textbf{Compress-FT-Alpaca} (ours) & \textbf{27.5} & 25.1 & 27.1 & \textbf{26.8} & \textbf{26.8} & \textbf{28}  \\
& \textbf{LoRA-FT-Alpaca} & 25.5 & \textbf{27.1} & 25.3 & 25.8 & 25.8 & 24.2 \\
& \textbf{Base-Alpaca} & 25.5 & \textbf{27.1} & 25.4 & 25.9 & 25.9 & 24.1 \\ 
\midrule 
\multirow{4}{*}{3} & 
\textbf{Cent-LLaMA-CEval} & 25.6 & \textbf{26.3} & \textbf{27.3} & 25.2 & 26.0 & 25.7 \\ 
& \textbf{Compress-FT-LLaMA-CEval} (ours) & \textbf{26.8} & 25.3 & 26.4 & \textbf{26.6} & \textbf{26.4} & \textbf{27.2} \\ 
& \textbf{LoRA-FT-LLaMA-CEval}  &  24.4 &  24.8& 26.5 &  25.4& 25.1 & 25.0  \\ 
& \textbf{Base-LLaMA-CEval} & 21.6 & 23.4 & 23.9 & 23.3 & 22.8 & 20.3 \\ 
\midrule 
\multirow{4}{*}{4} & 
\textbf{Cent-Qwen} & 38.4	 & 57.2 & 	\textbf{56.5} & \textbf{43.9} & 	47 & 27.8 \\
& \textbf{Compress-Qwen} (ours) & \textbf{39.8} & \textbf{64.3} & 54.7 & 42.9 & \textbf{48.3} & \textbf{30.4} \\
& \textbf{LoRA-Qwen} & 38 & 57.2 & 53.9 & 40.4 & 45.6 & 27.7 \\
& \textbf{Base-Qwen} & 28.5 & 42.4 & 36.7 & 33.1 & 33.9 & 26.8 \\
\midrule 
\multirow{3}{*}{5} & 
\textbf{Compress-FT-LLaMA-U-Former-0} 
& \textbf{26.8} & 25.3 & \textbf{26.4} & \textbf{26.6} & \textbf{26.4} & 27.2\\
& \textbf{Compress-FT-LLaMA-U-Former-1} 
& 26.6 & 26.5 & 25.5 & 25.7 & 26.2 &26.8\\
& \textbf{Compress-FT-LLaMA-U-Former-2} 
& 22.7 & 23.7 & 22.8 & 23.5 & 23.1 & 22.7\\
\midrule 
\multirow{3}{*}{6} & 
\textbf{Compress-FT-LLaMA-Upsampling-1} 
& 26.6 & 26.6 & 25.6 & 25.7 & 26.2 & 26.8 \\
& \textbf{Compress-FT-LLaMA-Upsampling-2} 
& 26.6 & 26.5 & 25.5 & 25.7 & 26.2 & 26.8 \\
& \textbf{Compress-FT-LLaMA-Upsampling-4} 
&\textbf{26.8} & 25.3 & \textbf{26.4} & \textbf{26.6} & \textbf{26.4} & \textbf{27.3} \\
\midrule 
\multirow{2}{*}{7} & 
\textbf{Compress-FT-LLaMA} & 26.6 & \textbf{26.5} & 25.5 & 25.7 & 26.2&  26.8 \\
& \textbf{D-Compress-FT-LLaMA} & \textbf{26.8} & 26 & \textbf{26.8} & \textbf{26.5} & \textbf{26.6} & \textbf{26.9} \\
\midrule 
\multirow{4}{*}{8} & 
\textbf{Base-Qwen} & 46.62 & 65.52 & 51.2 & 63.69 & 56.07  & - \\
& \textbf{LoRA-Qwen} & \textbf{47.92} & 65.49 & 50.54 & 64.56 & 56.33 & -  \\
& \textbf{Cent-Qwen} & 47.32 & 65.58 & \textbf{51.43} & \textbf{64.95} &56.60 & - \\
& \textbf{Compress-Qwen} & 47.67  &  \textbf{66.01}  & 50.86 &  64.33 & \textbf{56.44}  &  - \\
& \textbf{Generalize-Compress-Qwen} & 47.03 & 65.71&  51.39 & 64.14 & 56.37 & - \\
\midrule 
\multirow{3}{*}{9} & 
\textbf{DP-LoRA-Qwen-$\epsilon$-0.25} & \textbf{26.2} & 23.24 & \textbf{26.8} & 23.82 & 25.22 & -\\
& \textbf{DP-LoRA-Qwen-$\epsilon$-2.0} & 24.99 & 22.94 & 26.56 & 25.61 & 25.21 & -\\
& \textbf{DP-LoRA-Qwen-$\epsilon$-8.0} & 25.25 &  \textbf{24.11} & 26.23 &\textbf{26.81} & \textbf{25.67} & - \\
\midrule 
\multirow{3}{*}{10} & 
\textbf{DP-Compress-Qwen-$\epsilon$-0.25} & 46.62	 & 64.8 & \textbf{52.64} & 63.56  & 56.37 & - \\
& \textbf{DP-Compress-Qwen-$\epsilon$-2.0} & \textbf{47.25}	 & \textbf{65.87} & 51.75 & \textbf{64.04} & \textbf{56.55} & - \\
& \textbf{DP-Compress-Qwen-$\epsilon$-8.0} &46.97	& 65.42	 & 51.56 & 63.98 & 56.31 & - \\
\bottomrule 
\end{tabular}
\end{table*}

\subsection{Experimental Settings}
\label{sec:exp_setup}

\noindent \textbf{Compared Methods.} We compare the performance of the LLM trained with our proposed method, \modelname~(\textbf{Compress-FT}), against several other models: the base model without fine-tuning (\textbf{Base}), the models trained with traditional FL using LoRA (\textbf{LoRA-FT}), and those trained with contrastive learning (\textbf{Cent}). Our implementation of FL for the LLMs is based on FedIT~\cite{zhang2023building,Shepherdgithub}.

\noindent \textbf{Foundation Models.}  
We evaluated our method on several 7B-parameter LLMs due to computational constraints:
- \textbf{LLaMA-7B}~\cite{touvron2023llama}: A foundational model trained on trillions of tokens from public data.
- \textbf{Alpaca-7B}~\cite{alpaca}: A fine-tuned variant of LLaMA-7B developed by Stanford, optimized for instruction following.
- \textbf{Qwen-7B}~\cite{bai2023qwen}: A 7B-parameter model from Alibaba Cloud's Qwen series, trained on diverse multilingual data.

\noindent \textbf{FL Settings.}
\textbf{Databricks-Dolly-15k~\cite{conover2023free}.}  
The Databricks-Dolly-15k dataset consists of eight categories: brainstorming, classification, closed question-answering (QA), creative writing, general QA, information extraction, open QA, and summarization. This dataset is divided into 100 segments using the Dirichlet allocation technique, with a parameter $\alpha$ set to 0.5. The scalar parameter $\alpha > 0$ determines how skewed or uniform the proportions of these segments are. A smaller value of $\alpha$ results in a highly skewed distribution, while a larger value leads to a more uniform distribution. By setting $\alpha = 0.5$, we achieve a non-IID data allocation for different clients. 
\textbf{C-Eval-dev~\cite{huang2024c}.} 
We utilize the C-Eval-dev set for federated fine-tuning. Each "dev" set for a specific subject includes five examples with explanations for a few-shot evaluation, totaling 260 training samples.
\textbf{MMLU-train~\cite{hendrycks2020measuring}.} 
The MMLU dataset covers 57 tasks, including elementary mathematics, US history, computer science, law, and more. We divide the first two datasets into $\mathcal{D}_1$ and $\mathcal{D}_2$ for stage one’s gradient data collection and stage two’s fine-tuning, respectively. Additionally, 600 samples are randomly selected from the MMLU dataset specifically for $\mathcal{D}_2$ in stage two. LLaMA-7B and Alpaca-7B each have 100 clients with a selection probability of 0.05, while Qwen-7B has three clients with a selection probability of 1. In the first scenario, we obtain a dropout FL situation. Furthermore, we use the test sets of the C-Eval and MMLU datasets for evaluation.

\noindent \textbf{Evaluation Metrics.}
We evaluated our models with two widely used LLM benchmarks. 
\textbf{C-Eval~\cite{huang2024c}} is a comprehensive Chinese evaluation system designed to assess advanced knowledge and reasoning skills of foundational models in a Chinese context. It includes 13,948 multiple-choice questions in 52 distinct fields, covering various educational stages. We use six conventional subject categories: STEM, social sciences, humanities, other, average, and Avg (hard). 
\textbf{MMLU~\cite{hendrycks2020measuring}}, standing for Measuring Massive Multitask Language Understanding, serves as a representative and well-recognized benchmark to evaluate the performance of LLMs. This benchmark includes approximately 16,000 multiple-choice questions in 57 academic disciplines, such as mathematics, philosophy, and medicine.

\subsection{Efficiency Enhancement (RQ1)}
\label{sec:cost}

To address RQ1, we performed experiments to evaluate the communication complexity, computational efficiency of \stageone, and the inference efficiency of using the AutoEncoder in \stagetwo. We evaluated the computational costs of the proposed methods by recording FLOPs, memory cost, and inference time for the AutoEncoder, verifying that our methods add tiny additional computational cost to FedLLM. The inference GPU time is calculated by measuring the inference time across 10,000 samples and then averaging the results.

\begin{table}[t!]
\centering
\caption{Computation Efficiency of \modelname.}
\label{tab:computational_efficiency}
\begin{tabular}{cccc} 
\toprule  
 & FLOPS & GPU  Memory &  Inference GPU Time \\
 \midrule 
 $\mathbf{Enc}$ & 0.81 G & 0.94 MB &  14.01 ms\\
 $\mathbf{Dec}$ & 1.80 G & 0.94 MB &  8.25 ms \\
 \textbf{LLM} &  48.23 G & 6,809.71MB & 33,601.13 ms \\
 \bottomrule
\end{tabular}
\end{table}

\subsection{Performance Comparisons (RQ2)}

We present the performance summary with C-Eval in Table~\ref{tab:C-eval Results}. Our methods (\textbf{Compress-FT}) show significant performance improvements compared to all baselines.
For the initial model LLaMA-7B, our approach (\textbf{Compress-FT}) improves the \textbf{Base} score as follows: from 21.6 to 26.6 in the STEM field, from 23.9 to 25.5 in the Humanities, from 23.3 to 25.7 in other fields, from 22.8 to 26.2 on average, and from 20.3 to 26.8 at the hard level (Avg(Hard)).
The performance of the \textbf{Compress-FT} approach is comparable to \textbf{LoRA-FT}, demonstrating improvements in total scores in all fields from STEM to Avg (hard), except in Social Sciences. In Social Sciences, \textbf{LoRA-FT} achieves a high score of 27.6, the highest among the methods compared. Notably, \textbf{Compress-FT} also outperforms centralized fine-tuning methods (\textbf{Cent}), suggesting that these fine-tuning methods may be more resilient to minor disturbances introduced by the compression process.
We observe similar performance improvements when using \textbf{Alpaca} and \textbf{Qwen} as the base models.

\subsection{Privacy-Preserving FedLLM (RQ3)} 
\label{sec:privacy}

To ensure differential privacy for the transmitted low-rank matrices, a small amount of Gaussian noise is added to the decompositions $\mathbf{A}$ and $\mathbf{B}$.
We compute the privacy budget using Gaussian Differential Privacy (GDP)~\cite{dong2022gaussian}, with detailed computations provided in Sect.~\ref{sec:append_privacy}. 
A smaller privacy budget necessitates the addition of more Gaussian noise to provide better privacy protection. 
The selected privacy budget $\epsilon \in \{ 0.25, 2.0, 8.0 \}$ represents tight, medium, and relaxed privacy budgets, respectively, as noted in~\cite{wu2025dpfedsub}.
In this situation, we design the AutoEncoder to minimize the influence of noise during the compression phase effectively. Applying varying intensities of noise to the transmitted data reveals that a small amount of noise before compression can enhance the effectiveness of fine-tuned LLM models, suggesting that minimal noise can improve model robustness.
In contrast, the same level of Gaussian noise is applied to the \textbf{LoRA-FT} scenarios without compression for comparison. Assessments show that this Gaussian noise negatively impacts model performance, with a tighter privacy budget leading to significant performance degradation in \textbf{LoRA-FT}.

Furthermore, we can place the decoder on the client side to enhance privacy preservation during downlink communication, as discussed in Sect.~\ref{sec:adv}. In this scenario, the server only aggregates the encrypted information through the encoder. The client then receives the aggregated encrypted gradients and applies the decoder. We present the model evaluations in the 7th Row of Table~\ref{tab:C-eval Results}. \textbf{D-Compress-FT-LLaMA}, which places the decoder on the client side, shows a slight improvement over the proposed methods compared to \textbf{Compress-FT-LLaMA}.

\subsection{Formal Privacy Guarantee (RQ3)}
\label{sec:append_privacy}
We formally analyze the differential privacy properties of our proposed methods. Since the low-rank gradients, denoted as \( \mathbf{A} \) and \( \mathbf{B} \), are symmetric in our privacy analysis, we will focus on the privacy analysis for \( \mathbf{A} \). The analysis for \( \mathbf{B} \) will be analogous.

The concept of differential privacy is introduced in a formal mathematical definition in~\cite{dwork2014algorithmic}. In this work, we focus on the client-level differential privacy, which means the smallest privacy unit we analyze the the per-client data. Thus, we follow the definition of differential privacy at the client level in~\cite{geyer2017differentially}.

\begin{definition}[Client-level Differential Privacy~\cite{geyer2017differentially}]
A random mechanism $\mathcal{M}$ satisfies client-level differential privacy if for any two neighboring datasets $\mathcal{D}$ and $\mathcal{D}'$ that differ by the addition or removal of a client, and for any subset output $S \subset \rm{Range} (\mathcal{M})$, the following inequality holds:
$$ Pr[\mathcal{D} \in S] \leq e^{\epsilon} Pr[\mathcal{D'} \in S] + \delta. 
$$
\end{definition}

Before we introduce the random mechanism for the publication of gradients, we first define the following operators.

\begin{definition}[Clipping]
\label{def:clip}
The clipping operator on a tensor $\mathbf{A}$ with constant value $C$ is defined by
$
\mathbf{Clip}(\mathbf{A}, C) = \mathbf{A} \min \left(1,{C}/{\| \mathbf{A} \|_2} \right). 
$
\end{definition}

\begin{definition}[Noising]
\label{def:noise}
The noising operator on a tensor $\mathbf{A}$ with Gaussian noise of magnitude $\sigma$ is defined by 
$
\mathbf{Noise}(\mathbf{A}, \sigma) =  \mathbf{A} + \mathbf{n}_A, \quad \mathbf{n}_A \sim \mathcal{N}(0, \sigma I).
$
\end{definition}

In our FL setting, the only information leakage occurs during gradient exchange between local client and global server. Thus, we operate the local gradients with several additional steps, including clipping ($\mathbf{Clip}$), noising ($\mathbf{Noise}$), and aggregation ($\mathbf{Aggre}$) for the publication of the differential private gradient at the client level between the encoding on the client and the decoding on the server. 
\begin{equation}
\label{eqn:random}
\mathcal{M} =  \mathbf{Aggre} \circ \mathbf{Noise}\circ \mathbf{Clip}.
\end{equation}
The overall gradient publication procedure can be defined as.
\begin{equation}
\label{eqn:random2}
\tilde{\mathcal{M}} = \mathbf{Dec} \circ \mathcal{M} \circ \mathbf{Enc}. 
\end{equation}

\begin{lemma} [Sensitivity of $\tilde{\mathcal{M}}$] Let $K$ be the number of selected clients per communication round. Suppose per-client low rank gradients $\mathbf{A}$ clipped with constant $C$. The sensitivity of aggregating local gradients is $ S = \frac{C}{K}$. 
\label{lem:sens_2}
\end{lemma}


Applying the sensitivity results of Lemma~\ref{lem:sens_2}, it is possible to calculate the loss of privacy for each iteration using Gaussian Differential Privacy (GDP)~\cite{bu2019deep}. Furthermore, as indicated by the Central Limit Theorem in~\cite{bu2019deep}, we can also derive the approximate value of \( G_{\mu} \) to assess the overall privacy loss.
\begin{theorem}
[Gaussian Differential Privact~\cite{bu2019deep}]
\label{thm:privacy_CLT}
Suppose \modelname~equipped with random gradient mechanism $\mathcal{M}$ defined in Eq.~\eqref{eqn:random} run with number of steps $T$ and Poisson sampling without replacement with probability $p = K/M$, which satisfies $p \sqrt{T} \rightarrow \nu$. 
By choosing $\sigma$ followed by the following equations 
\begin{eqnarray}
\delta(\epsilon) &=& \Phi(- \frac{\epsilon}{\mu} + \frac{\mu}{2}) - e^{\epsilon} \Phi(- \frac{\epsilon}{\mu} - \frac{\mu}{2}); \\
\mu & =& \nu \cdot \sqrt{ (e^{1/\sigma^2} - 1)}. 
\end{eqnarray}
Thus, \modelname~with $\mathcal{M}$ achieves client-level $(\epsilon, \delta)$-DP.
\end{theorem}

\subsection{Training Details of AutoEncoder (RQ4)}
\label{sec:details_auto}

In this section, we demonstrate the convergence behavior of \stageone. 
Due to the page limit, we provide the convergence curve in the full version~\cite{}.
In Figure~\ref{fig:auto_encoder_conv}, it is evident that compared to the 1D AutoEncoder, the 2D AutoEncoder equipped with CNN and U-Former blocks achieves a significantly lower endpoint in the training process, effectively capturing the essential information of the transmitted gradients. Moreover, among the architectures that incorporate the U-Former, the configuration with zero U-Former blocks exhibits the highest accuracy in fitting the transmitted data. 
The smaller log-loss achieved by AutoEncoders contributes to better model performances.
Moreover, with the U-Former design, the trained AutoEncoder shows a small gap between the train and test loss, which indicates a better generalization ability and avoids overfitting. 
Furthermore, by analyzing the distribution of the transmitted gradient values and the MSE loss detailed in Figure\ref{fig:auto_encoder_conv}, we can compute the signal-to-noise ratio (SNR) of the reconstructed gradients, which can assess the effects of noise reduction.
SNR is characterized as the proportion of signal power relative to noise power~\cite{johnson2006signal}.
The power of the $\ell_2$ norm for the low-rank decomposition of input gradients before compression is displayed in the first row of Table~\ref{tab:ratio_compress}. 
A higher SNR indicates better preservation of information during the compression process. According to Table~\ref{tab:ratio_compress}, the compressor developed using ResNet records the highest SNR, whereas the 1D CNN exhibits the least.

\subsection{Ablation Study (RQ4)}
\label{sec:ablation}

This section aims to demonstrate the impact of modifications in $\auto$ during \stageone~on the \modelname~performance, as detailed in Table~\ref{tab:C-eval Results}. 
Our experimental analysis shows that reducing the number of U-Former blocks improves the performance of fine-tuned models. In particular, the zero block U-Former configuration achieves the highest scores in all subjects, except ``Social Science". This finding aligns with the observation that the zero-block U-Former configuration also yields the smallest log-loss during Auto-Encoder training, see the second figure in Figure~\ref{fig:auto_encoder_conv}. 
Compared to the \textbf{Base} scenario, the zero-block U-Former configuration shows improvements ranging from 5.6 to 6.9. Relative to the \textbf{Cent} scenario, the top score increases by up to 3.8. When compared with the \textbf{LoRA-FT} scenario, improvements reach 2.4 in the Avg (hard) category. In conclusion, fewer U-Former building blocks lead to better performance in fine-tuned models.
 
Furthermore, adjustments were made to the number of up-sampling and down-sampling blocks within $\mathbf{Enc}$ and $\mathbf{Dec}$ while maintaining the U-Former blocks at 3. It was observed that reducing the number of up-sampling and down-sampling blocks to 4 (\textbf{Compress-FT-LLaMA-Upsampling-4}) led to better performance in the fine-tuned model compared to completely eliminating the U-Former blocks.
More details on AutoEncoder Architecture can be found in the Supplementary Material.

\subsection{Generalization to Unseen Datasets (RQ5)}
\label{sec:generalization} 
We further evaluate the generalizability of our proposed methods under two different conditions. In the first condition, the training set is divided into two parts: one to train the AutoEncoder in \stageone~and the other for fine-tuning FedLLM in \stagetwo. The data sets are divided into a 3:7 ratio, with 30\% used to train the AutoEncoder and the remaining 70\% to fine-tune the LLM.
In the second condition, we use two different but related datasets for \stageone~and \stagetwo. Specifically, the AutoEncoder is pre-trained on the C-Eval dataset, while the LLM is fine-tuned using 600 randomly sampled examples from the MMLU training dataset~\cite{hendrycks2020measuring}.
To ensure a fair evaluation, both settings are tested on MMLU. The results are presented in the seventh section of Table~\ref{tab:C-eval Results}. The \textbf{Compress-Qwen} represents pre-training and fine-tuning using different subsets of the same dataset, while the \textbf{Generalize-Compress-Qwen} represents pre-training with C-Eval and fine-tuning with dolly. The generalized approach shows an improvement over \textbf{Cent-Qwen}.

\section{Conclusion}

In summary, to address the communication efficiency and privacy guarantee issues in federated fine-tuning of LLMs, we propose a novel gradient compression method utilizing pre-trained AutoEncoders. The encoder is kept on the local client, while the decoder is adopted on the server. Through compressed federated fine-tuning, our methods achieve superior performance in representative evaluations.
Furthermore, a detailed analysis encompassing the signal-to-noise ratio, compression ratio, and robustness in the noised gradients further verifies the effectiveness of our proposed compression method. 
One limitation of our work is the additional training and inference cost that the AutoEncoder brings. Although the additional cost is not heavy (e.g., only a few minutes for training in this paper), we will explore more efficient architecture and training strategy in the future.

\section*{Acknowledgments}
This work was supported by the \textbf{Key R\&D Program of Zhejiang} (Grant No.~2024C01036).

\bibliographystyle{plain}
\bibliography{ecai25}

\appendix

\begin{figure*}[h]
\centering 
     \begin{subfigure}[b]{0.95\textwidth}
         \centering
         \includegraphics[width=\textwidth]{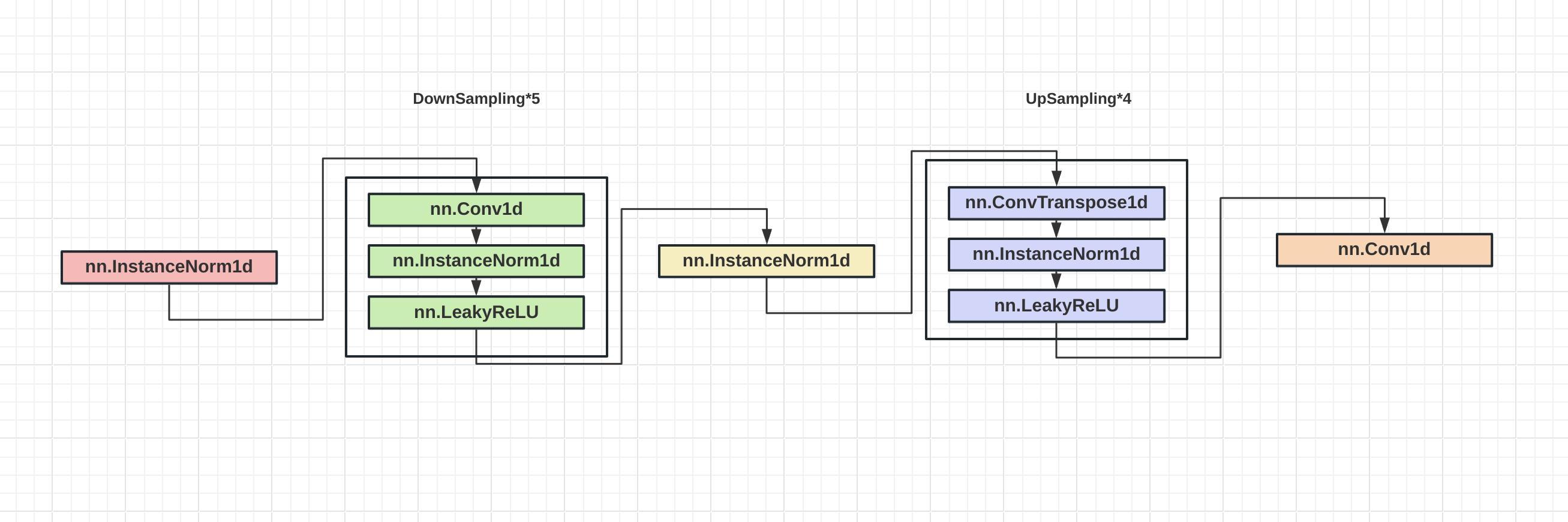}
         \vspace{0.1cm}
         \caption{Architecture of AutoEncoder with 1D CNN~\cite{abrahamyan2021learned}}
         \label{fig:archi_ae_cnn}
     \end{subfigure}
    \vspace{0.25cm}
    \vfill 
     \begin{subfigure}[b]{0.95\textwidth}
         \centering
         \includegraphics[width=\textwidth]{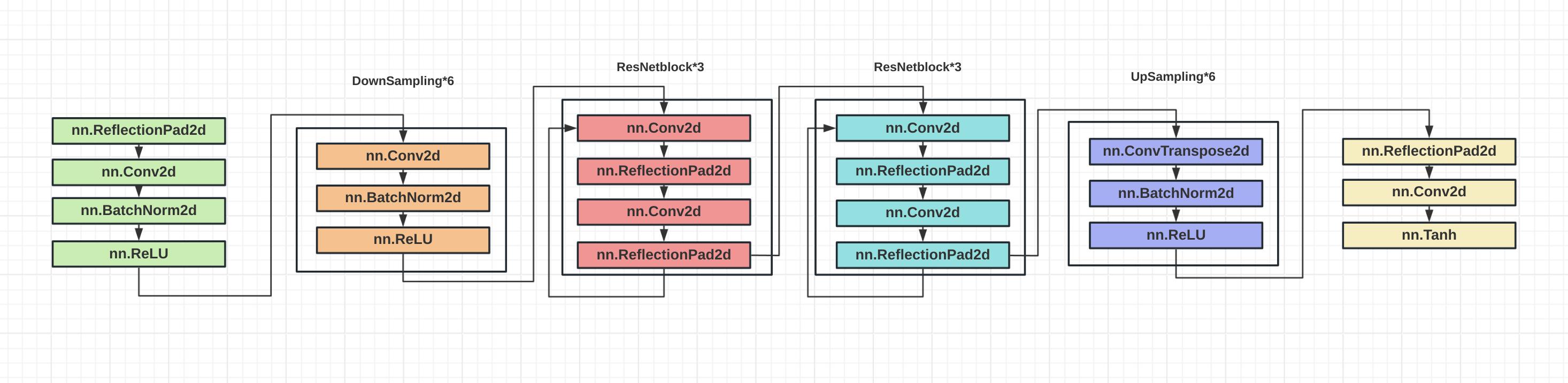}
         \caption{Architecture of AutoEncoder with ResNet~\cite{he2016deep}}
         \label{fig:archi_ae_resnet}
     \end{subfigure}
     \vspace{0.25cm}
     \vfill 
     \begin{subfigure}[b]{0.95\textwidth}
         \centering
         \includegraphics[width=\textwidth]{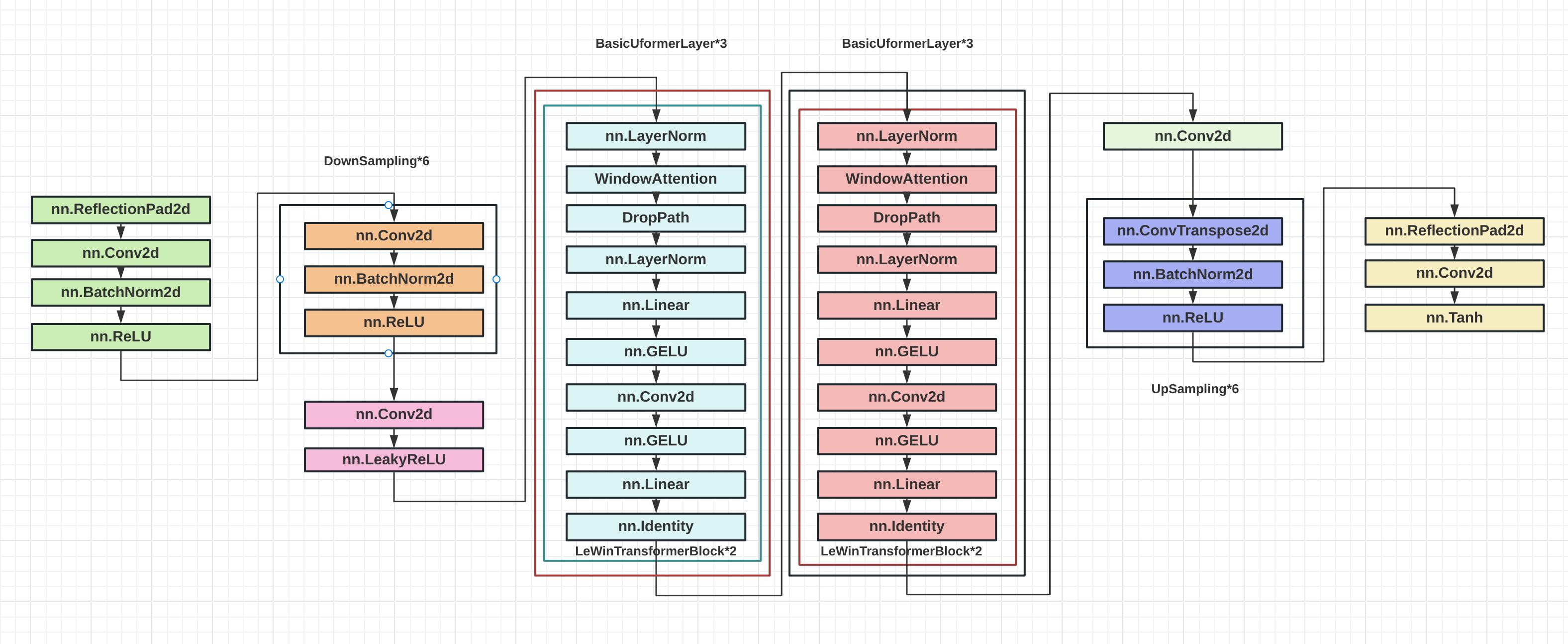}
         \caption{Architecture of AutoEncoder with Uformer~\cite{wang2022uformer}}
         \label{fig:archi_ae_uformer}
     \end{subfigure}
     \vspace{0.25cm}
     \vfill 
\caption{Architectures of AutoEncoder with different building blocks. From top to bottom are 1D CNN~\cite{abrahamyan2021learned}, ResNet~\cite{he2016deep}, and Uformer~\cite{wang2022uformer} respectively.}
\label{fig:archi_autoencoder}
\end{figure*}

\section{Experimental Details}
\label{sec:append_details}
Here we present more experimental details, including computing power in Sect.~\ref{sec:append_comput}, training details of \stageone ~in Sect.~\ref{sec:append_tgap}, and training details of \stagetwo~in Sect.~\ref{sec:append_faf}.

\subsection{Computing Power}
\label{sec:append_comput}

\noindent \textbf{Hardware Specifications.}
To support GPU-accelerated computations, one of the nodes was outfitted with an NVIDIA RTX 4090 GPU, each possessing 24 GB of memory. This setup increases our ability to perform parallel processing on tasks that require high computational power, such as the pre-training of AutoEncoder in \stageone~ and federated fine-tuning in \stagetwo.~\\
\noindent  \textbf{Software Environment.}
Our computational infrastructure was developed using Ubuntu Linux 22.04. Additionally, to improve the performance of certain machine learning components in our simulations, we utilized Pytorch 2.2.1 coupled with CUDA 12.4, which facilitated the smooth incorporation of GPU processing into our processes.

\subsection{Training Details of \stageone}
\label{sec:append_tgap}
During the initial phase (\stageone) of the AutoEncoder training, the loss of reconstruction is defined by the mean squared error (MSE), which represents the difference $\ell_2$ between the original and reconstructed data. The Adam optimizer~\cite{kingma2014adam} is utilized, set at a learning rate of $2 \times 10^{-4}$. A comprehensive outline of the AutoEncoder's architecture is provided in the following section, referenced in Sect.~\ref{sec:append_model}.

\subsection{Training Details of \stagetwo}
\label{sec:append_faf}
In this section, we describe the dataset partitioning approach, federated training hyperparameters, and aggregation techniques for the second phase \stagetwo~ of federated fine-tuning. During the data setup phase, the \textbf{Databricks-dolly-15k} dataset is segmented among 10 clients with no disparity in data quantity. Similarly, the \textbf{C-eval-dev} dataset is allocated among 3 clients with zero difference in data quantity. Furthermore, the hyperparameters used in the federated fine-tuning are detailed in Table~\ref{tab:append_hyper}.

\begin{table*}[t]
\centering
\caption{Hyperparameters in \stagetwo}
\label{tab:append_hyper}
\resizebox{1.0\linewidth}{!}{
\begin{tabular}{c   c c c c c c}   
\toprule
 & \textbf{ChatGLM} + \textbf{C-Eval}  &  \textbf{LLaMA} + \textbf{C-Eval} &  \textbf{Alpaca} + \textbf{dolly-15k}  & \textbf{LLaMA} + \textbf{dolly-15k} & \textbf{Qwen} + \textbf{MMLU}   \\ \midrule
\# communication rounds & 5 & 5  & 20 & 20 & 8  \\
client selection fraction & 1.0 & 1.0 & 0.05 & 0.05 & 1.0 \\
local batch size & 24 & 24 & 32 & 32 & 8  \\
local micro batch size & 8 & 8 & 16 & 16 & 4  \\
local learning rate & $3 \cdot 10^{-4}$ & $2.0 \cdot 10^{-4}$ &  $1.5 \cdot 10^{-4}$ & $1.5 \cdot 10^{-4}$ & $3.5 \cdot 10^{-4}$ \\
low rank parameters & 16 & 16 & 8 & 8 & 8  \\
local number epochs  & 3  & 3  & 1 & 1 & 2\\
\bottomrule
\end{tabular}}
\end{table*}

\section{Architecture of AutoEncoder}
\label{sec:append_model}

In this section, we explore various designs of the AutoEncoder ($\auto$) that were pretrained during \stageone~and detail their implementations in \stagetwo.

\subsection{AutoEncoder with 1-D CNN}
Initially, we introduce the AutoEncoder design which uses the fundamental elements of 1D convolutional neural networks (CNN)~\cite{abrahamyan2021learned} as shown in Table~\ref{tab:archi-1d}. This configuration mirrors the structure of the 1D CNN networks discussed in \cite{abrahamyan2021learned}. The architecture comprises five 1d convolutional layers (conv1d) combined with ReLU activation layers to compress the input gradients. Subsequently, four 1d deconvolutional layers (deconv1d) along with ReLU and an additional conv1d are incorporated to ensure that the output gradients maintain the same form as the input gradients. The entire process is segmented into two phases, the encoder $\mathbf{Enc}$ and the decoder $\mathbf{Dec}$. The encoder is made up of the initial five conv1d layers, whereas the decoder includes the subsequent five deconv1d layers and one more conv1d layer. During \stageone, the encoder and decoder are simultaneously trained with data derived from temporal gradients. In \stagetwo, the encoder is deployed on the client side and the decoder on the server side, with the encoder tasked with compressing the gradients for transmission and the decoder tasked with restoring the compressed gradients to their original state. The detailed architecture of the 1D-CNN Auto-Encoder is outlined in Table~\ref{tab:archi-1d} and the training methodology is depicted in Figure~\ref{fig:archi_ae_cnn}.

\subsection{Analysis of Compressed Data}

\begin{figure*}[h]
     \begin{subfigure}[b]{0.475\textwidth}
        \centering 
         \includegraphics[width=\textwidth]{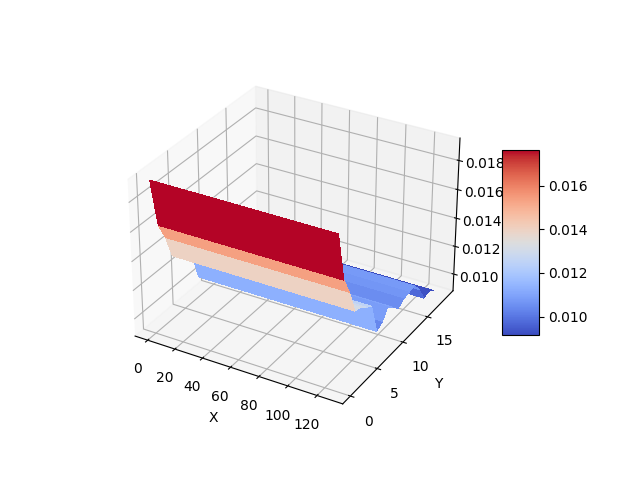}
     \end{subfigure}
     \begin{subfigure}[b]{0.475\textwidth}
        \centering 
         \includegraphics[width=\textwidth]{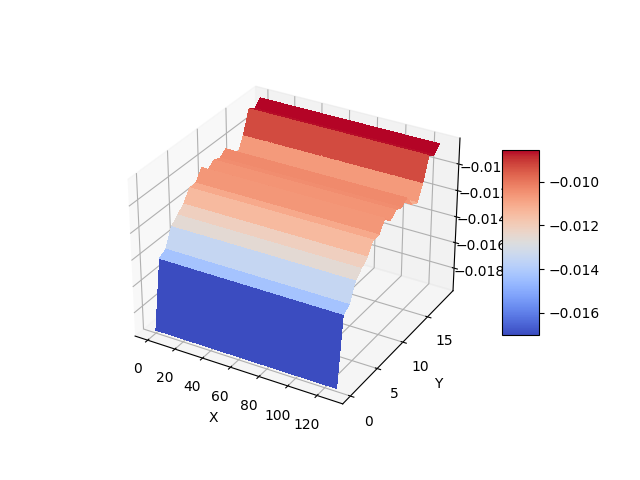}
     \end{subfigure}
\caption{Distribution of maximum values (left) and minimum values (right) of matrices $\mathbf{A}$'s and $\mathbf{B}$'s from the iteration 1 to iteration 20. Y axis represents the training epoch, X axis denotes the index of the matrices $\mathbf{A}$ and $\mathbf{B}$ to compress ($2 \times 4 \times 32=128$ dimension).}
\label{fig:max_min_dist}
\end{figure*}

In this section, we show the experimental analysis of compressed data. 
As mentioned in the main context, Sects.3.2 and 3.3, we compress the gradients processed with LoRA, as $\mathbf{A}$ and $\mathbf{B}$.
Using the \textbf{LLaMA-7B} model, the shapes of the matrices $\mathbf{A}^{\intercal}$ and $\mathbf{B}$ are $8 \times 4096$. 
We concatenate the matrices $\mathbf{A}^{\intercal}$ and $\mathbf{B}$ for the projection matrices $\mathbf{Q}, \mathbf{K}, \mathbf{V}, \mathbf{O}$ for 32 Transformers blocks in the first dimension.  Thus, the input shape of the AutoEncoder becomes $(8 \times 2 \times 4 \times 32) \times 4096 =  2048 \times 4096$.
Figure~\ref{fig:max_min_dist} illustrates the distribution of maximum and minimum values in all concatenated matrices $\mathbf{A}$ and $\mathbf{B}$ during TGAP. During training, the maximum values (left) vary from 0.0018 to 0.0010, while the minimum values (right) are symmetrically distributed from -0.0018 to -0.0010. Furthermore, from iterations 1 to 20, there is a noticeable convergence trend in both the maximum and minimum values of the transmitted parameters. 
Ultimately, the maximum and minimum values of $\mathbf{A}$ and $\mathbf{B}$ converge to the order of $10^{-3}$, indicating an overall distribution of $\mathbf{A}$ and $\mathbf{B}$ approaching zero as iterations increase. This zero gradient value signifies convergence, and the regular evolution of $\mathbf{A}$ and $\mathbf{B}$ allows the success of training the AutoEncoder.

\subsection{Auto-Encoders with ResNet}
\label{sec:archi_resnet}
Presented here is the 2D architecture utilizing ResNet~\cite{he2016deep}. The downsampling layers are composed of six 2d convolutional layers (conv2d) paired with ReLU activation layers, designed to refine the shape of the input gradient. Each ResNet block includes two conv2d layers, focused on extracting essential features from the compressed gradients while maintaining the original shape. The upsampling process involves six 2d deconvolutional layers (deconv2d) along with ReLU activation layers, tasked with restoring the original gradient from the compressed versions. The AutoEncoder $\auto$ is segmented into two components: the encoder $\mathbf{Enc}$ and the decoder $\mathbf{Dec}$. The encoder consists of the downsampling layers and three ResNet blocks, whereas the decoder comprises three ResNet blocks and the upsampling layers. During \stageone, both the encoder and the decoder are concurrently trained using data derived from temporal gradients. In \stagetwo, the encoder is placed on the client side and the decoder on the server side, with the encoder's objective being to compress the gradients for transmission and the decoder's to revert the compressed gradients back to their original form. The comprehensive architecture of the ResNet Auto-Encoder is detailed in Table~\ref{tab:archi-2d} and the training process is illustrated in Figure~\ref{fig:archi_ae_resnet}.

\begin{table*}[t]
\centering
\caption{AutoEncoder with 1-D CNN}
\label{tab:archi-1d}
\begin{tabular}{c c c c c c}  
\toprule
 & Layer & In Channels & Out Channels  & Kernel Size & Stride  \\ \midrule
\multirow{5}{*}{Encoder} 
& conv1d & 1 & 64 & 3 & 2 \\ 
 & conv1d & 64 & 128 & 3 & 2 \\ 
& conv1d & 128 & 256 & 3 & 2 \\  
& conv1d & 256 & 64 & 3 & 2 \\  
& conv1d & 64 & 4 & 1 & 1 \\ 
\midrule
\midrule 
\multirow{5}{*}{Decoder} 
& deconv1d & 4 & 64 & 3 & 2 \\ 
 & deconv1d & 64 & 128 & 3 & 2 \\ 
& deconv1d & 128 & 256 & 3 & 2 \\  
& deconv1d & 256 & 128 & 3 & 2 \\  
& deconv1d & 128 & 64 & 3 & 2 \\  
& conv1d & 64 & 1 & 1 & 1 \\
\bottomrule
\end{tabular}
\end{table*}

\begin{table*}[t]
\centering
\caption{Auto-Encoders with ResNet}
\label{tab:archi-2d}
\begin{tabular}{c c c c c c c}  
\toprule
 & Layer & In Channels & Out Channels  & Kernel Size & Stride  \\ \midrule
\multirow{8}{*}{Encoder}  
& \multirow{7}{*}{DownSampling Layers} 
& conv2d & 1 & 1 & (3,3) & (1,1) \\ 
& & conv2d & 1 & 2 & (3,3) & (2,2) \\ 
& & conv2d & 2 & 4 & (3,3) & (2,2) \\ 
& & conv2d & 4 & 8 & (3,3) & (2,2) \\ 
& & conv2d & 8 & 16 & (3,3) & (2,2) \\ 
& & conv2d & 16 & 32 & (3,3) & (2,2) \\  
& & conv2d & 32 & 64 & (3,3) & (2,2) \\  
&  \multicolumn{6}{c}{ResNet Block $\times 3$} \\
\midrule
\midrule
\multirow{8}{*}{Decoder}
&  \multicolumn{6}{c}{ResNet Block $\times 3$} \\
& 
\multirow{7}{*}{UpSampling Layers} 
& deconv1 & 4 & 32 & (3,3) & (2,2) \\ 
& & deconv2d & 64 & 32 & (3,3) & (2,2) \\ 
& & deconv2d & 32 & 16 & (3,3) & (2,2) \\  
& & deconv2d & 16 & 8 & (3,3) & (2,2) \\  
& & deconv2d & 8 & 4 & (3,3) & (2,2) \\ 
& & deconv2d & 4 & 2 & (3,3) & (2,2) \\  
& & deconv2d & 2 & 1 & (3,3) & (2,2) \\ 
& & conv2d & 1 & 1 & (7,7) & (1,1) \\
\bottomrule
\end{tabular}
\end{table*}

\subsection{AutoEncoder with Uformer}
We continue to develop the AutoEncoder with Uformer as suggested in~\cite{wang2022uformer}, showcasing its structure in Table~\ref{tab:archi-uformer} and its training process in Figure~\ref{fig:archi_ae_uformer}. In this version of AutoEncoder, we retain the upsampling and downsampling layers used in the AutoEncoder with ResNet, but replace the core building blocks with BasicUformer, each containing two LeWinTransformerBlock. The encoder is equipped with the DownSampling layers and three BasicUformer blocks, while the decoder incorporates three BasicUformer blocks and the UpSampling layers. Both the pre-training in \stageone~and the fine-tuning in \stagetwo~are conducted as outlined in Sect.~\ref{sec:archi_resnet}.

\begin{table*}[t]
\centering
\caption{AutoEncoder with Uformer}
\label{tab:archi-uformer}
\begin{tabular}{c c c c c c c}  
\toprule
 & Layer & In Channels & Out Channels  & Kernel Size & Stride  \\ \midrule
\multirow{8}{*}{Encoder}  
& \multirow{7}{*}{DownSampling Layers} 
& conv2d & 1 & 1 & (3,3) & (1,1) \\ 
& & conv2d & 1 & 2 & (3,3) & (2,2) \\ 
& & conv2d & 2 & 4 & (3,3) & (2,2) \\ 
& & conv2d & 4 & 8 & (3,3) & (2,2) \\ 
& & conv2d & 8 & 16 & (3,3) & (2,2) \\ 
& & conv2d & 16 & 32 & (3,3) & (2,2) \\   
& & conv2d & 32 & 64 & (3,3) & (2,2) \\
&  \multicolumn{6}{c}{Uformer Block $\times 3$} \\
\midrule
\midrule
\multirow{8}{*}{Decoder} 
&  \multicolumn{6}{c}{Uformer Block $\times 3$} \\
& \multirow{7}{*}{UpSampling Layers} 
& deconv1 & 4 & 32 & (3,3) & (2,2) \\ 
& & deconv2d & 64 & 32 & (3,3) & (2,2) \\ 
& & deconv2d & 32 & 16 & (3,3) & (2,2) \\  
& & deconv2d & 16 & 8 & (3,3) & (2,2) \\  
& & deconv2d & 8 & 4 & (3,3) & (2,2) \\ 
& & deconv2d & 4 & 2 & (3,3) & (2,2) \\  
& & deconv2d & 2 & 1 & (3,3) & (2,2) \\ 
& & conv2d & 1 & 1 & (7,7) & (1,1) \\
\bottomrule
\end{tabular}
\end{table*}

\section{More Evaluation results}
In this section, we present a comparative analysis of our approaches \textbf{Compress-FT}, \textbf{LoRA-FT}, \textbf{Cent}, and \textbf{Base} across various benchmarks such as C-Eval~\cite{huang2024c}, Matrix Entropy~\cite{wei2024large} and Qwen API~\cite{bai2023qwen}.

\subsection{More Evaluation with C-Eval}
\label{sec:append_ceval}

\subsubsection{Statistical Significance on C-Eval Results}
\label{sec:append_stats}
\begin{figure*}[t!]
     \begin{subfigure}[b]{0.45\textwidth}
         \centering
         \includegraphics[width=\textwidth]{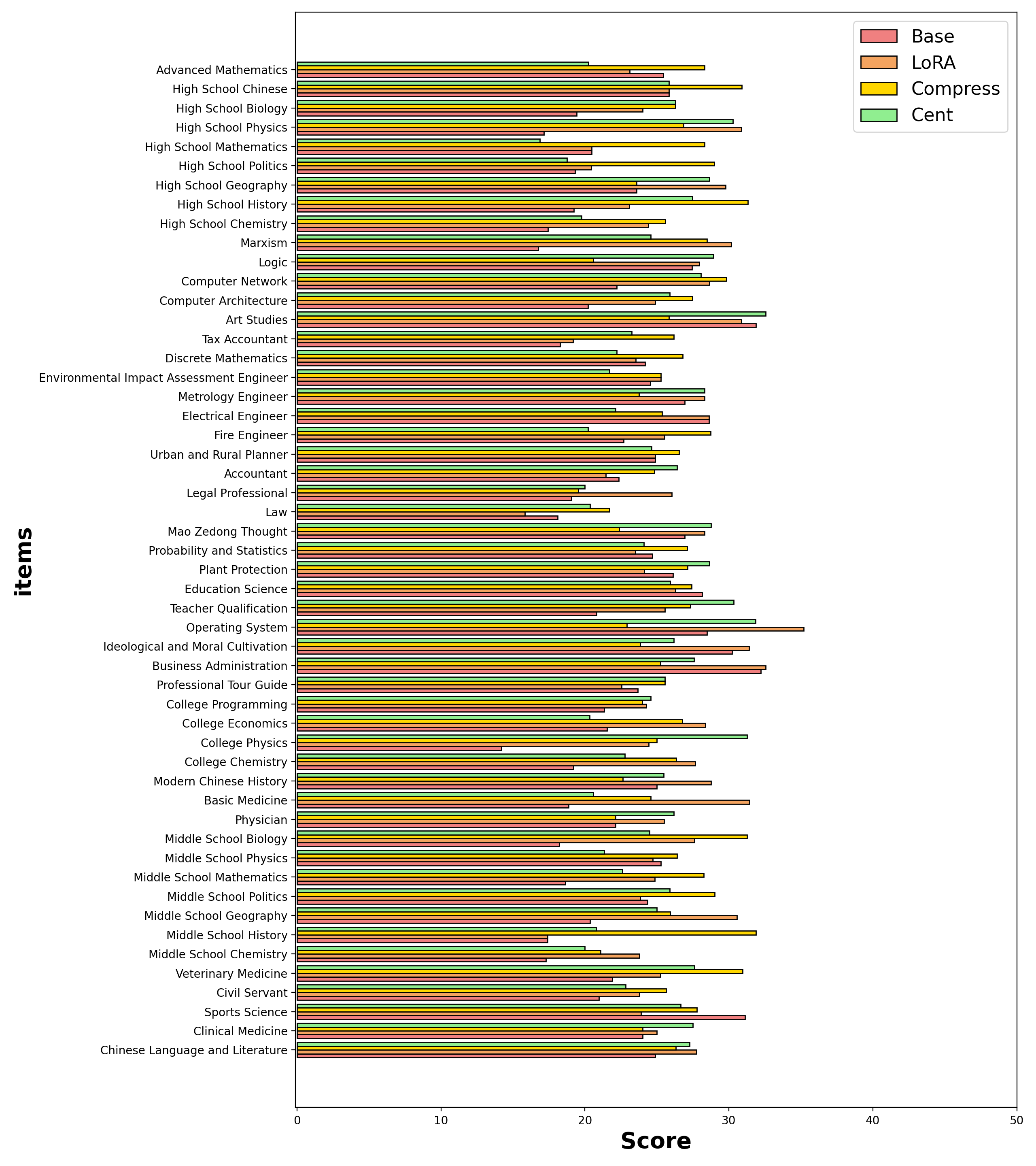}
     \end{subfigure}
     \begin{subfigure}[b]{0.45\textwidth}
         \centering
         \includegraphics[width=\textwidth]{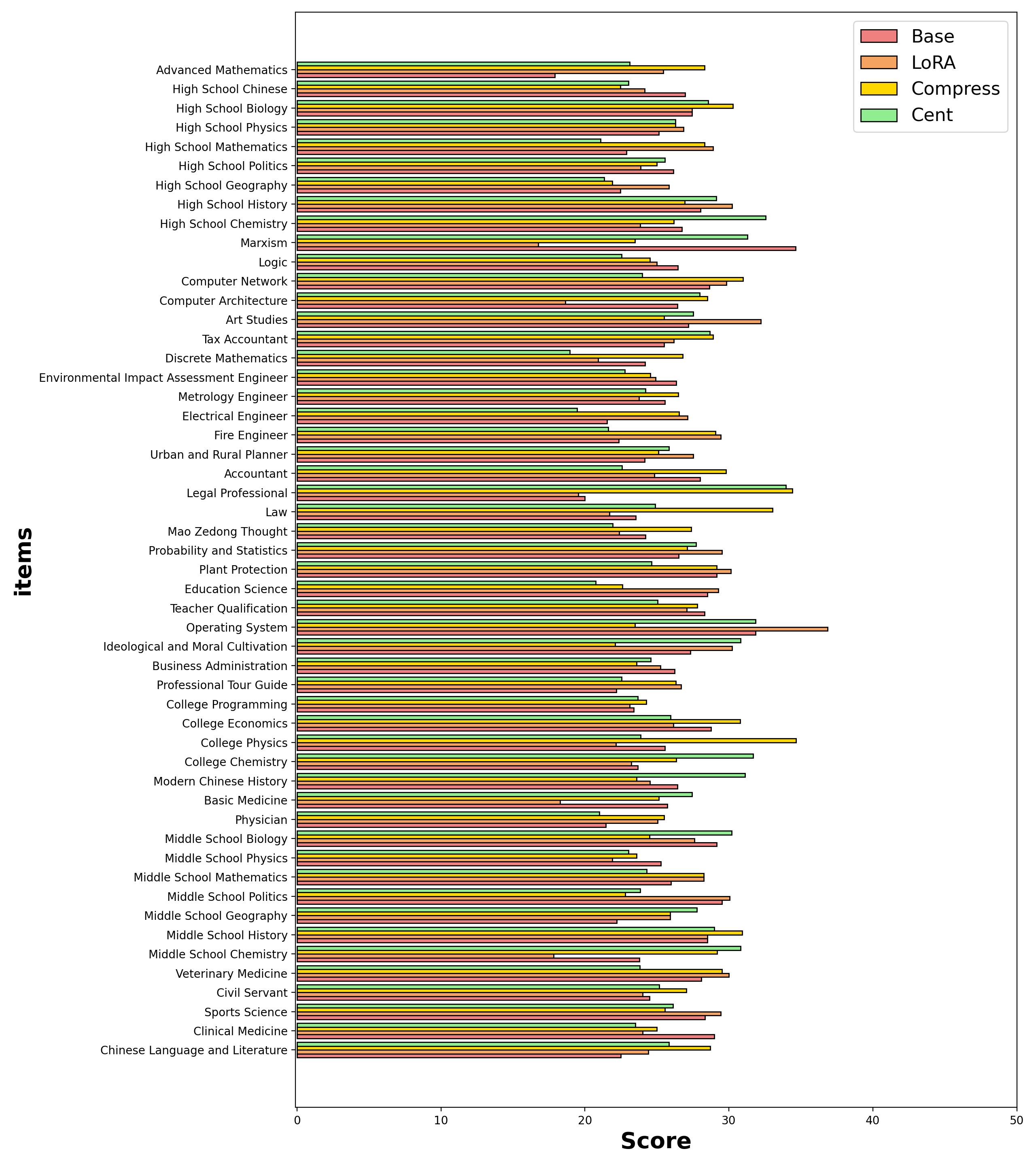}
     \end{subfigure}
     \vfill
     \begin{subfigure}[b]{0.45\textwidth}
         \centering
         \includegraphics[width=\textwidth]{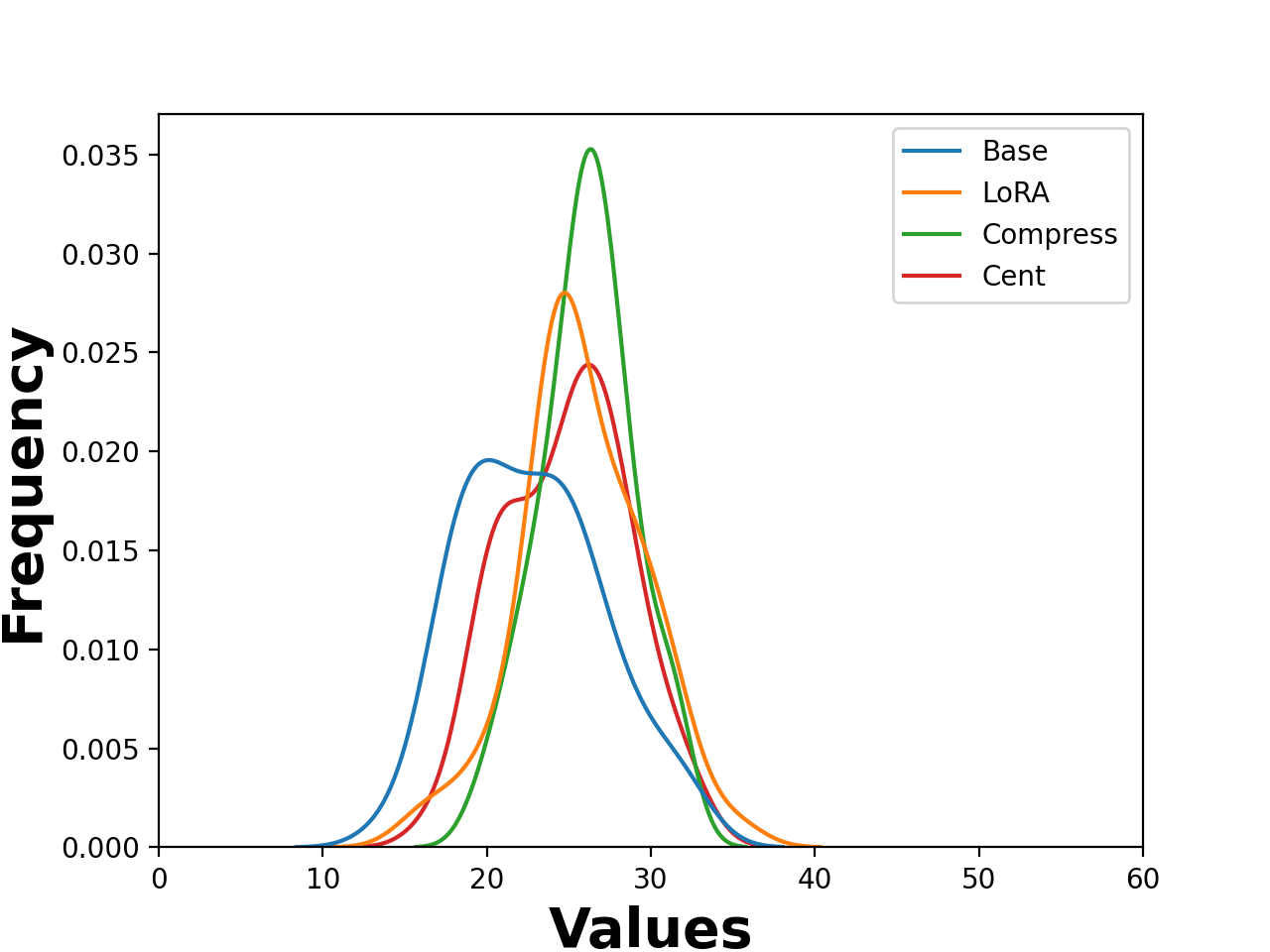}
     \end{subfigure}
     \begin{subfigure}[b]{0.45\textwidth}
         \centering
         \includegraphics[width=\textwidth]{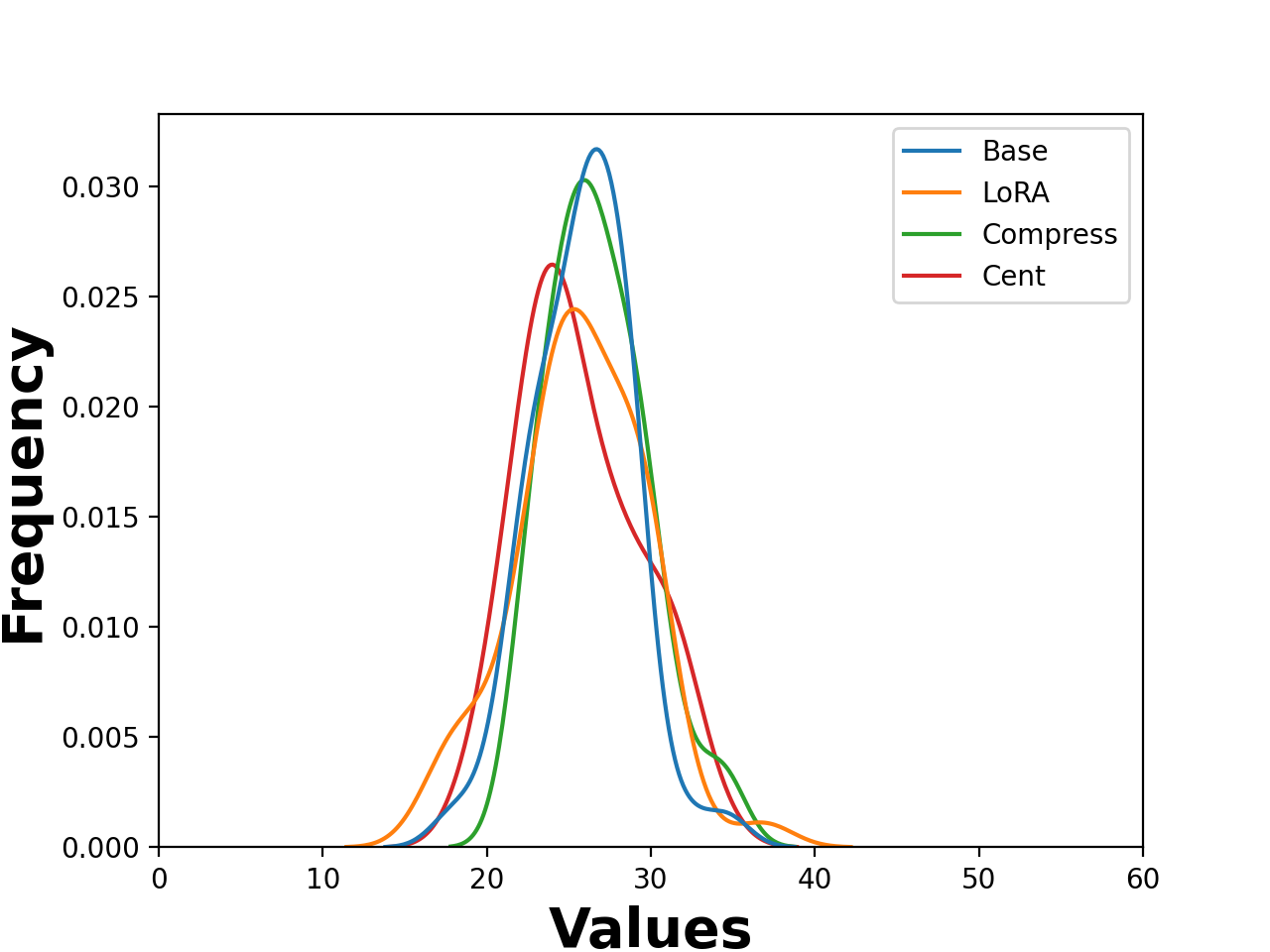}
     \end{subfigure}
\caption{Comprehensive evaluation results in frequency (Top) and distribution (Down) based on C-Eval with foundation model \textbf{LLaMA} (Left) and \textbf{Alpaca} (Right).}
\label{fig:full_ceval}
\end{figure*}

In this section, we show the statistical significance of our methods compared to existing models and base models. 
Here we detail the exhaustive results of C-Eval~\cite{huang2024c} as shown in Figure~\ref{fig:full_ceval}. A total of 52 examinations were conducted. We provide a detailed comparison of our method \textbf{Compress} (yellow) against the foundational models \textbf{Base} (red), the LoRA fine-tuned models \textbf{LoRA} (orange), and the centralized fine-tuned models \textbf{Cent} (green). On the left in Figure~\ref{fig:full_ceval}, the performance of the model is shown alongside the foundational model \textbf{LLaMA} is shown, while to the right, the performance with the foundational model \textbf{Alpaca}.
Furthermore, we perform a statistical analysis on the extensive evaluation data presented in Table~\ref{tab:full_ceval_llama}, involving the foundational models \textbf{LLaMA} and \textbf{Alpaca}. In the initial section of Table~\ref{tab:full_ceval_llama}, it is observed that our approach \textbf{Compress-FT} records the highest scores in the categories of `Mean', `Min', `25\% percentile', and `50\% percentile'. Furthermore, this method also shows the smallest standard deviation in scores, as indicated in the 'Std' column. Conversely, the \textbf{LoRA-FT} model stands out by securing the highest scores in the `75\% percentile' and `Max' categories.
In the second part of Table~\ref{tab:full_ceval_llama}, with the foundation model \textbf{Alpaca}, the \textbf{Compress-FT} methods achieves the highest score in all including `Mean', `25 \%', `50 \%', `75 \%', and `Max'. And our methods have similar performance with \textbf{Base} model in Std. value. 
Given that each evaluation uses a 4-choice format, a score of 25.0 represents a random selection. We calculate the score count exceeding 25.0 and display these counts in the final column of Table~\ref{tab:full_ceval_llama}.
The method \textbf{Compress-FT} records the highest frequency of scores above 25.0, demonstrating the effectiveness of our federated fine-tuning approach.
Moreover, we graphically represent these scores using a frequency and value approach in Figure~\ref{fig:full_ceval}, further confirming the statistical significance of our techniques.

\begin{table*}[t!]
\centering
\caption{Statistics of comprehensive C-Eval with foundation model \textbf{LLaMA}.}
\label{tab:full_ceval_llama}
\begin{tabular}{ccccccccc} 
\toprule
Methods & 	 Mean & Std. & Min & 25\%	& 50\% & 75\% & Max	&  number of scores $\geq 25.0$ \\
\midrule
\textbf{Cent-LLaMA} & 24.9 & 3.6 & 16.8 & 22.0 & 25.5 & 27.5 & 32.5 & 27\\
\textbf{Compress-FT-LLaMA} (ours) & \textbf{26.2} & \textbf{2.8} & \textbf{19.5} & \textbf{24.4} & \textbf{26.3} & 27.9 & 31.9 & \textbf{36} \\
\textbf{LoRA-FT-LLaMA} &  25.7 & 3.7 & 15.8 & 23.8 & 25.3 & \textbf{28.3} & \textbf{35.1} & 27 \\
\textbf{Base-LLaMA} & 22.7 & 4.3 & 14.2 & 19.2 & 22.3 & 25.3 & 32.2 & 14 \\
\midrule
\textbf{Cent-Alpaca} & 25.7 & 3.6 & 18.9 &	23.0 &	24.9 &	28.1 & 33.9 &	26\\
\textbf{Compress-FT-Alpaca} (ours) &  \textbf{26.8} & \textbf{3.0} & \textbf{21.9} & \textbf{24.5} & \textbf{26.4} & \textbf{28.8} & \textbf{34.7} & \textbf{36} \\
\textbf{LoRA-FT-Alpaca} &  25.7 & 3.9 & 16.7 & 23.8 & 25.6 & 28.6 & 36.8 & 29 \\
\textbf{Base-Alpaca} &  25.8 & \textbf{3.0} & 17.9 &	23.7 & 26.1 &	28.0 & 34.6 & 34 \\
\bottomrule
\end{tabular}
\end{table*}


\subsubsection{A Reduction in the ResNet Block}

Further results from C-Eval are presented in Table~\ref{tab:ceval_uformer} and Table~\ref{tab:ceval_resnet}.
Due to the fact that we observe an increasing performance with a decrease in the number of former U blocks in Table~\ref{tab:ceval_uformer}, we carry out more experiments with a reduction of the number of ResNet blocks while observing a different conclusion. 
Decreasing the number of ResNet blocks in the Auto-Encoder is associated with lowered model performance. The AutoEncoder setup with three ResNet blocks demonstrates the best results in the C-Eval, followed by the setup with one ResNet block. Consequently, in AutoEncoder systems that utilize ResNet structures, an increase in the number of ResNet blocks leads to improved model tuning. 

\begin{table*}[t]
\centering
\caption{A Comparison of C-Eval with different number of Uformer blocks.}
\label{tab:ceval_uformer}
\begin{tabular}{ccccccc} 
\toprule
Methods & Stem & Social Sciences & Humanities & Others & Average & Avg(hard) \\
\midrule
\textbf{Compress-FT-LLaMA-Uformer-0} (ours) 	
& \textbf{26.8} & 25.3 & \textbf{26.4} & \textbf{26.6} & \textbf{26.4} & \textbf{27.2}\\
\textbf{Compress-FT-LLaMA-Uformer-1} (ours) 
& 26.6 & 26.5 & 25.5 & 25.7 & 26.2 &26.8\\
\textbf{Compress-FT-LLaMA-Uformer-2} (ours) 
& 22.7 & 23.7 & 22.8 & 23.5 & 23.1 & 22.7\\
\textbf{Compress-FT-LLaMA-Uformer-3} (ours) 
& 21.8 & 24.2 & 23.9 & 23.8 & 23.1 & 20.1 \\
\midrule 
\textbf{Cent-LLaMA} 
& 24.5 & 25.6 & 25.5 & 24.4 & 24.9 & 23.4 \\ 
\textbf{LoRA-FT-LLaMA} 
& 25.9 & \textbf{27.6} & 25.2 & 24.5 & 25.8 & 24.8 \\
\textbf{Base-LLaMA} 
& 21.6 & 23.4 & 23.9 & 23.3 & 22.8 & 20.3 \\ 
\bottomrule
\end{tabular}
\end{table*}

\begin{table*}[t]
\centering
\caption{A Comparison of C-Eval with different number of ResNet blocks.}
\label{tab:ceval_resnet}
\begin{tabular}{ccccccc} 
\toprule
Methods & Stem & Social Sciences & Humanities & Others & Average & Avg(hard) \\
\midrule 
\textbf{Compress-FT-LLaMA-ResNet-0} (ours) 
&21.8 & 23.6 & 23.7 & 23.3 & 22.9 & 20.3 \\
\textbf{Compress-FT-LLaMA-ResNet-1} (ours) 
& 23.8 & 24.5 & 23.9 & 23.7 & 23.9 & 22.5\\ 
\textbf{Compress-FT-LLaMA-ResNet-2} (ours) 
& 21.6 & 23.3 & 23.9 & 23.3 & 22.8 & 20.2 \\
\textbf{Compress-FT-LLaMA-ResNet-3} (ours) 
& \textbf{26.6} & 26.5 & \textbf{25.5} & \textbf{25.7} & \textbf{26.2} & \textbf{26.8} \\
\midrule 
\textbf{Cent-LLaMA} 
& 24.5 & 25.6 & 25.5 & 24.4 & 24.9 & 23.4 \\ 
\textbf{LoRA-FT-LLaMA} 
& 25.9 & \textbf{27.6} & 25.2 & 24.5 & 25.8 & 24.8 \\
\textbf{Base-LLaMA} 
& 21.6 & 23.4 & 23.9 & 23.3 & 22.8 & 20.3 \\ 
\bottomrule
\end{tabular}
\end{table*}

\subsubsection{Temporal Behavior of \stagetwo}

Here we present the results of C-Eval in different epochs to observe how the model is used in \stagetwo. 
We selected four experimental groups: 
\textbf{Compress-LLaMA-UpSampling-4}, 
\textbf{Compress-LLaMA-Uformer-1}, 
\textbf{Cent-Alpaca}, 
and \textbf{Compress-Alpaca}, 
to monitor the behavior of the model during fine-tuning epochs. 
Each experimental group was tested in three randomly selected epochs out of a total of 20 epochs. The results show that the basic \textbf{LLaMA} model achieves early optimal convergence. Specifically, \textbf{Compress-LLaMA-Uformer-1} and \textbf{Compress-LLaMA-UpSampling-4} attained their highest performance by Epoch 10. Following this peak, continued fine-tuning tended to reduce overall model effectiveness, yet it improved performance in certain specific areas such as Avg(Hard), demonstrating a balance between different specialized areas. 
In the set of experiments \textbf{Compress-LLaMA-ResNet-3}, the model achieves the best peroformance around Epoch 13 and keeps on the best performance until Epoch 19. 
In contrast, the foundational model \textbf{Alpaca} showed a delayed convergence to its best form during compression fine-tuning. For instance, in the \textbf{Cent-Alpaca} experiment set, the optimal model emerged around Epoch 15, while in \textbf{Compress-Alpaca}, the peak model appeared in Epoch 19, enhancing scores by up to 2.5 on average over the model tuned in Epoch 10. In particular, the model from Epoch 17 scored the highest at 28 in the Avg (hard) category. This phenomenon can be attributed to a minor loss of information in the initial phases of fine-tuning due to compression, necessitating more epochs to reach the optimal model. However, despite the delayed convergence, the compressed model ultimately surpasses the performance of the centralized fine-tuning approach with the foundational model \textbf{Alpaca}. 

\begin{table*}[t]
\centering
\caption{Convervence behavior under C-Eval.}
\label{tab:ceval_conv}
\begin{tabular}{ccccccc} 
\toprule
Methods & Stem & Social Sciences & Humanities & Others & Average & Avg(hard) \\
\midrule 
\textbf{Compress-LLaMA-UpSampling-4-Epoch-19} & 26.6 & 25 & 26.2 & 26 & 26 & \textbf{27.9} \\
\textbf{Compress-LLaMA-UpSampling-4-Epoch-13} & 26.2 & \textbf{25.3}	 & \textbf{26.4} & \textbf{26.6} & 26.1 & 25.6 \\
\textbf{Compress-LLaMA-UpSampling-4-Epoch-10} & \textbf{26.8} & 25.3 & \textbf{26.4} & \textbf{26.6} & \textbf{26.4} & 27.3 \\
\midrule 
\textbf{Compress-LLaMA-Uformer-1-Epoch-19}	& 24.9	 & 24.9 & 	24.2	& 24.4	& 24.6 & 25.8 \\
\textbf{Compress-LLaMA-Uformer-1-Epoch-17}	& 24.9 & 24.6 & 24.6 & 24.6 & 24.7 & 25.7 \\
\textbf{Compress-LLaMA-Uformer-1-Epoch-10}	& \textbf{26.6} & \textbf{26.5} & \textbf{25.5} & \textbf{25.7} & \textbf{26.2} & \textbf{26.8} \\
\midrule 
\textbf{Compress-LLaMA-ResNet-3-Epoch-19} & 26.6 & 26.5 & 25.5 & 25.7 & 26.2 & 26.8 \\
\textbf{Compress-LLaMA-ResNet-3-Epoch-13} & 26.6 & 26.5 & 25.5 & 25.7 & 26.2 & 26.8 \\
\textbf{Compress-LLaMA-ResNet-3-Epoch-12}  & 21.7 & 24.6 & 23.7 & 24.2 & 23.2 & 20.4 \\
\midrule 
\textbf{Cent-Alpaca-Epoch-17} & 24.8 & 24.9 & 25.6 & 25.3 & 25.1 & 24.3 \\
\textbf{Cent-Alpaca-Epoch-15} & 25.4 & \textbf{25.4} & 26.9 & 25.4 & \textbf{25.7} & 25 \\
\textbf{Cent-Alpaca-Epoch-13} & \textbf{25.9} & 24.8 & \textbf{27.3} & 24.5 & \textbf{25.7} & \textbf{25.7} \\
\midrule 
\textbf{Compress-Alpaca-Epoch-19} & \textbf{27.6} & 24.9 & \textbf{27.4} & \textbf{26.9} & \textbf{26.9} & 27.6 \\
\textbf{Compress-Alpaca-Epoch-17} & 27.5 & 25.1 & 27.1 & 26.8 & 26.8 & \textbf{28} \\
\textbf{Compress-Alpaca-Epoch-10} & 25.3 & \textbf{27} & 25.5 & 25.5 & 25.7 & 23.7 \\
\bottomrule
\end{tabular}
\end{table*}

\begin{figure*}[htbp]
     \begin{subfigure}[b]{0.475\textwidth}
         \centering
         \includegraphics[width=\textwidth]{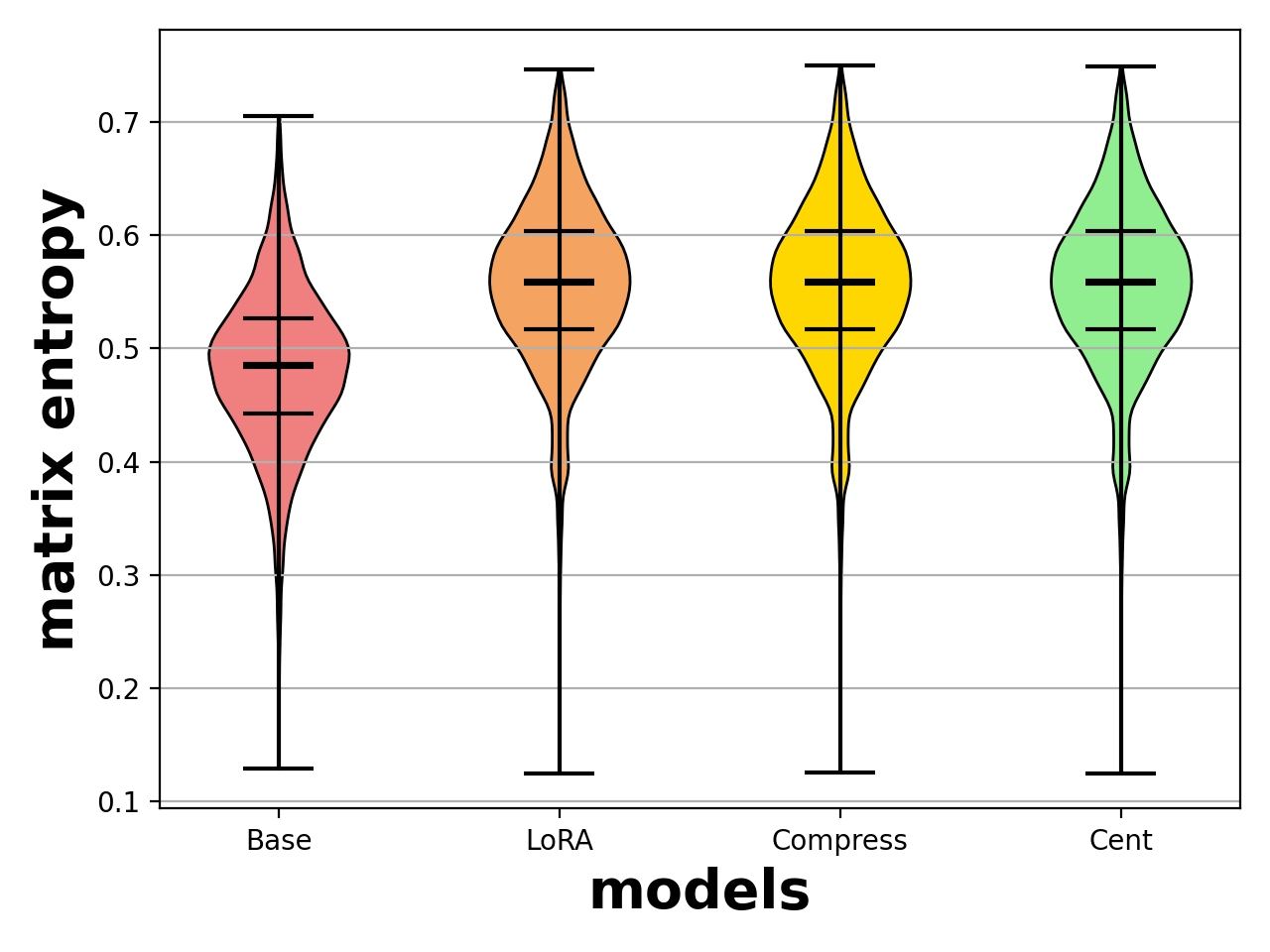}
     \end{subfigure}
     \begin{subfigure}[b]{0.475\textwidth}
         \centering
         \includegraphics[width=\textwidth]{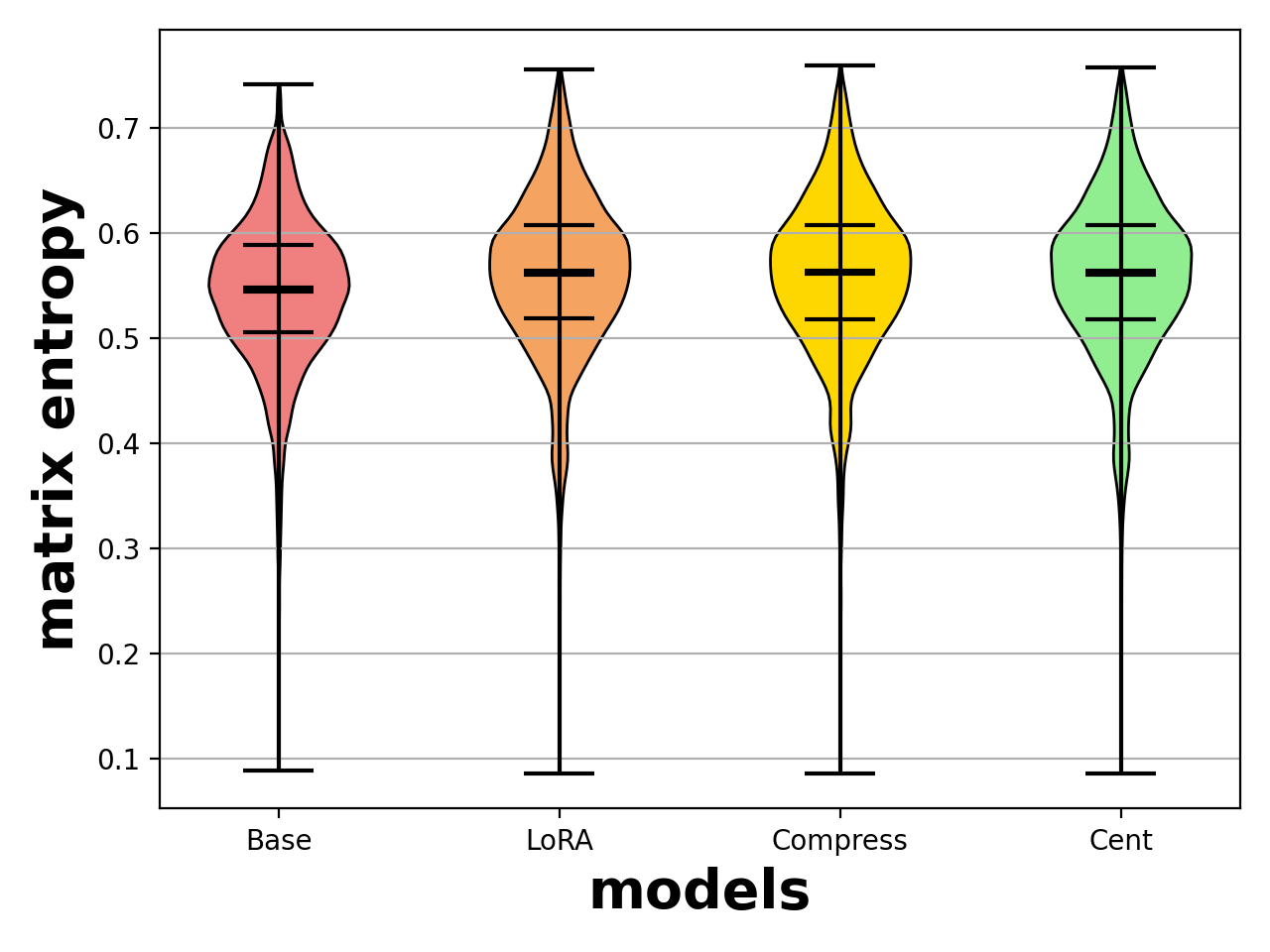}
     \end{subfigure}
\caption{Matrix Entropy on LLaMA-7B and Alpaca-7B foundation models and fine-tuned models.}
\label{fig:matrix_entropy}
\end{figure*}

\subsection{Evaluation by Matrix Entropy}
\label{sec:append_matrix_entropy}
In this section, we evaluate the matrix entropy for the LLM models introduced in~\cite{wei2024large}. The author introduces matrix entropy as a metric for LLMs based on information theory and provides a clear explanation of quantum information theory~\cite{bennett1998quantum,wilde2013quantum}. Imagine considering each token as a state within a quantum system. In this analogy, matrix entropy corresponds to a density matrix for quantum states, indicative of the mean number of qubits needed for state encoding.
The matrix entropy for matrix $\mathbf K$ is defined as 
$$
H(\mathbf K) = - \rm{tr} (\mathbf K \log \mathbf K). 
$$
Define $\Sigma_{S_i}$ as the covariance matrix formed by the representations of sentences, which are the hidden states of an LLM. Algorithms a, b, and c are designed to calculate the matrix entropy of an LLM across particular datasets. 
Algorithm a is computed by
$$
H_a (\mathcal{D}) = \frac{\sum_{i=1}^n H (\mathbf \Sigma_{S_i})}{n \log d}. 
$$
Algorithm b is computed by
$$
H_b (\mathcal{D}) = \frac{\exp(\sum_{i=1}^n H (\mathbf{\Sigma}_{S_i}) / n)}{d}. 
$$
And Algorithm c is computed by
$$
H_c (\mathcal{D}) = \frac{\exp(\sum_{i=1}^n H (\mathbf{\Sigma}_{S_i}) )}{nd}. 
$$
From an information theory viewpoint~\cite{schumacher1995quantum}, matrix entropy quantifies the 'intrinsic' information within the data. Higher values indicate a high uncertainty of the LLM model. According to the findings in~\cite{wei2024large}, there is a trend of reduction in the fine-tuning models. In our analysis, we note that our model \textbf{Compress-FT} records matrix entropy values comparable to \textbf{Cent} and \textbf{LoRA-FT}, but shows increased values on \textbf{Base} for the foundational models \textbf{LLaMA} and \textbf{Alpaca}, aligned with the results reported in~\cite{wei2024large}.

\begin{table*}[htbp]
\centering
\caption{LLM Matrix Entropy Evaluation Results}
\label{tab:matrix_entropy}
\begin{tabular}{cccc} 
\toprule
Methods & Algorithm a & Algorithm b & Algorithm c \\
\midrule
\textbf{Cent-LLaMA} & 0.557 & 0.025 & 0.029 \\
\textbf{Compress-FT-LLaMA} (ours) & 0.557 & 0.025 & 0.029 \\
\textbf{LoRA-FT-LLaMA} & 0.557 & 0.025 & 0.029 \\
\textbf{Base-LLaMA} & 0.483 & 0.013 & 0.016 \\
\midrule 
\textbf{Cent-Alpaca} &  0.560 &  0.025  &  0.031 \\
\textbf{Compress-FT-Alpaca} (ours) &  0.560 &  0.025  &  0.031 \\
\textbf{LoRA-FT-Alpaca} & 0.560   &  0.025  &  0.031 \\
\textbf{Base-Alpaca} & 0.544 & 0.022 & 0.026 \\
\bottomrule
\end{tabular}
\end{table*}

\begin{figure*}[t]
     \begin{subfigure}[b]{0.245\textwidth}
         \centering
         \includegraphics[width=\textwidth]{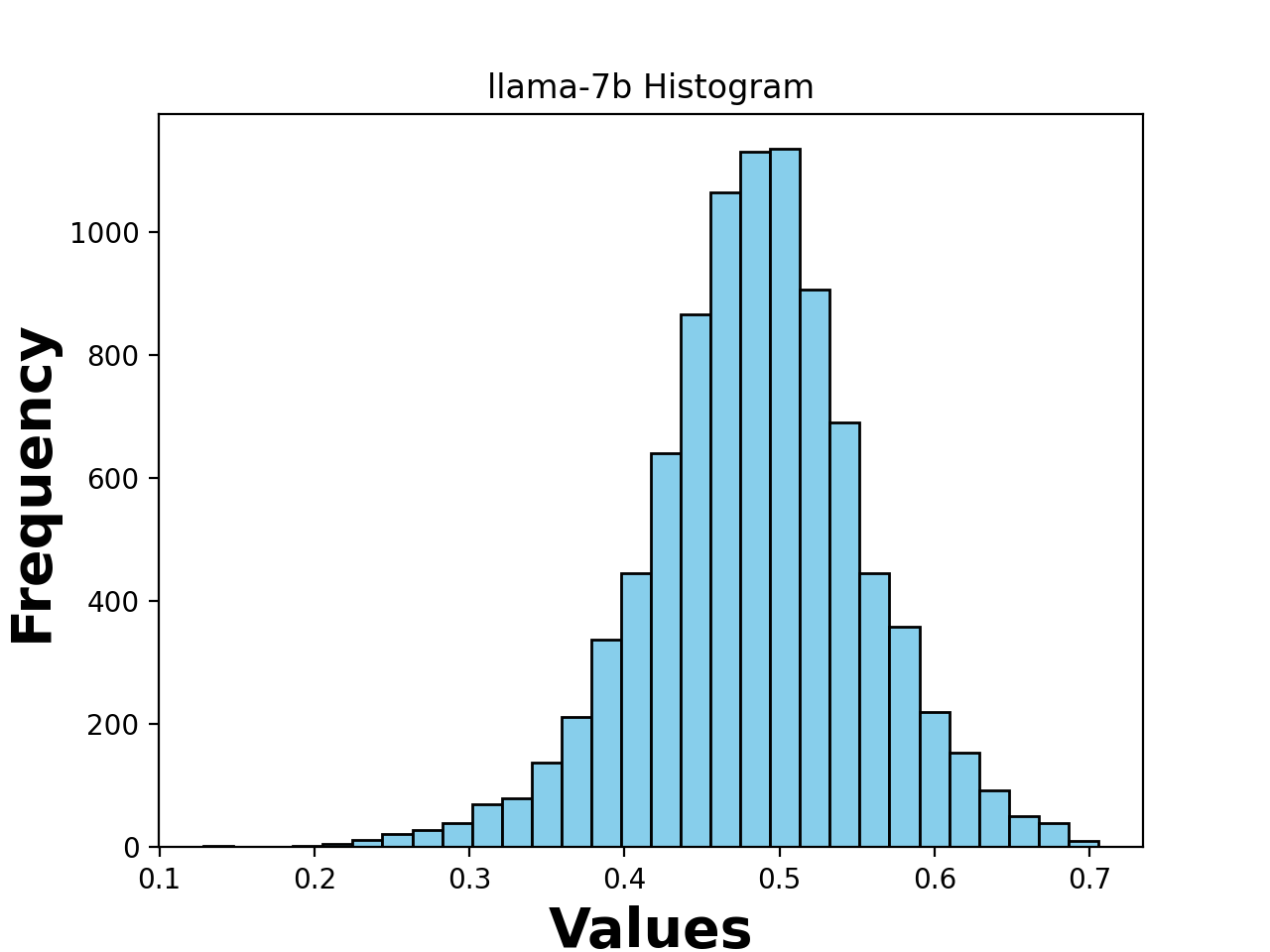}
     \end{subfigure}
     \begin{subfigure}[b]{0.245\textwidth}
         \centering
         \includegraphics[width=\textwidth]{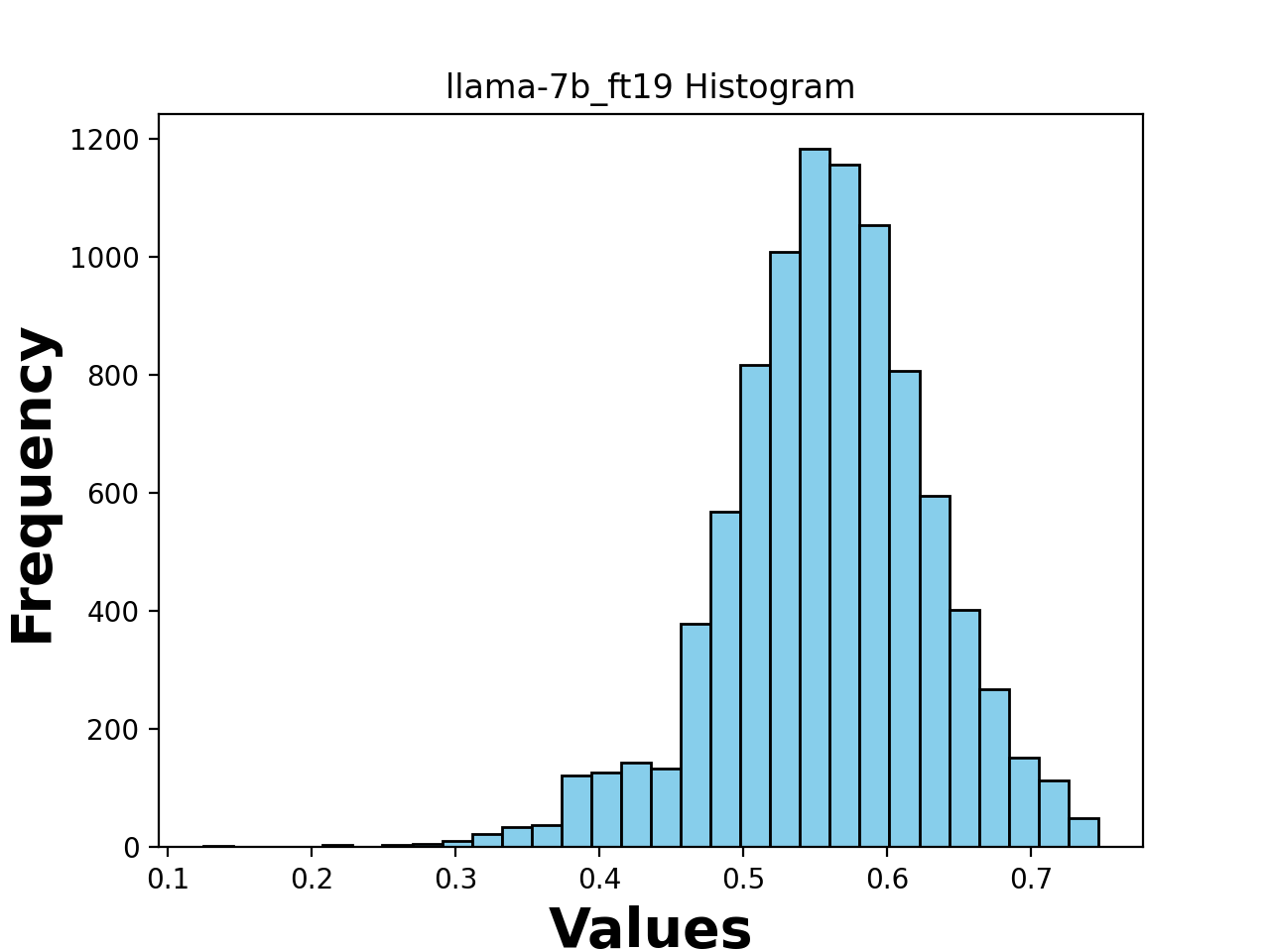}
     \end{subfigure}
     \begin{subfigure}[b]{0.245\textwidth}
         \centering
         \includegraphics[width=\textwidth]{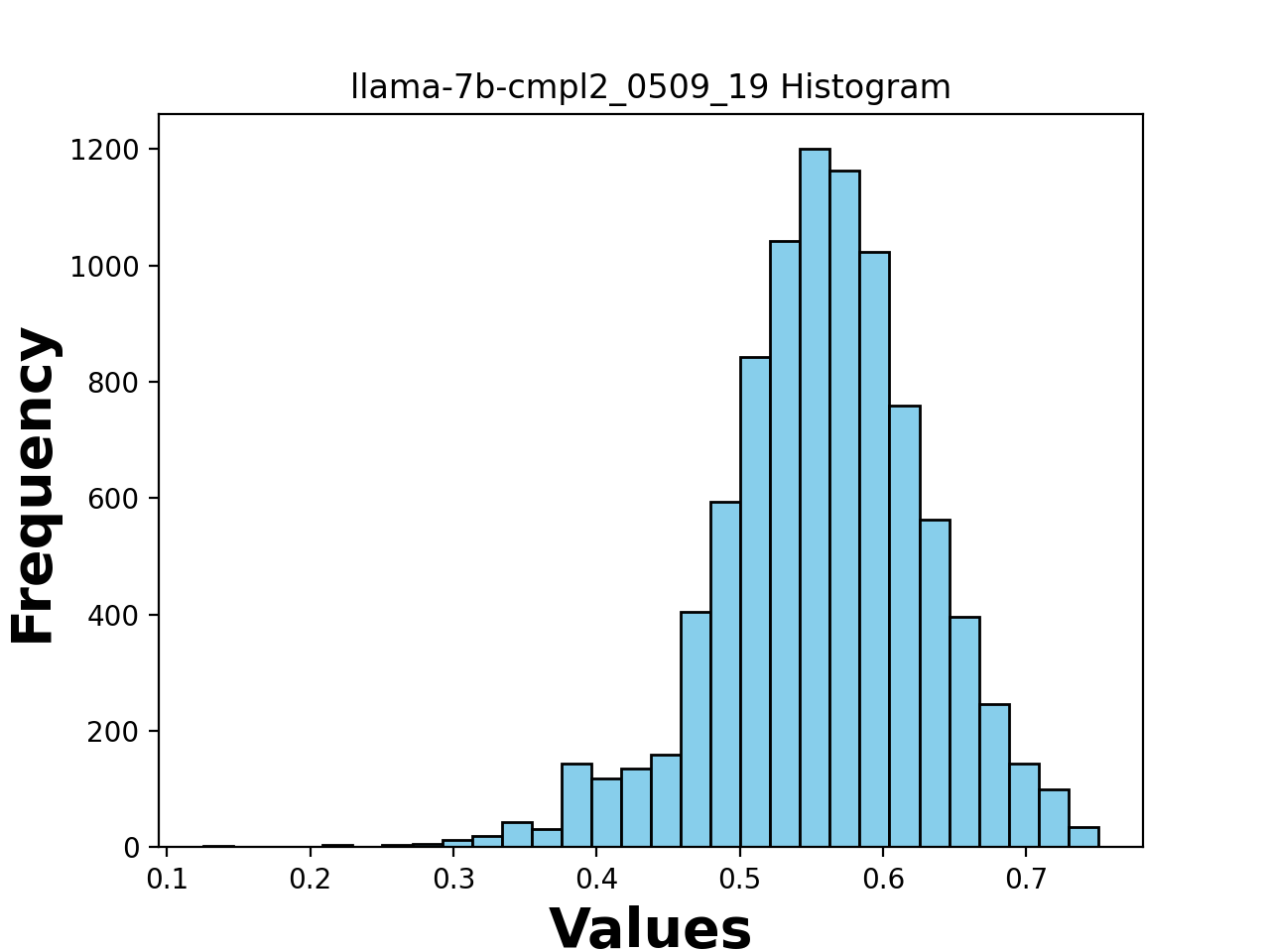}
     \end{subfigure}
     \begin{subfigure}[b]{0.245\textwidth}
         \centering
         \includegraphics[width=\textwidth]{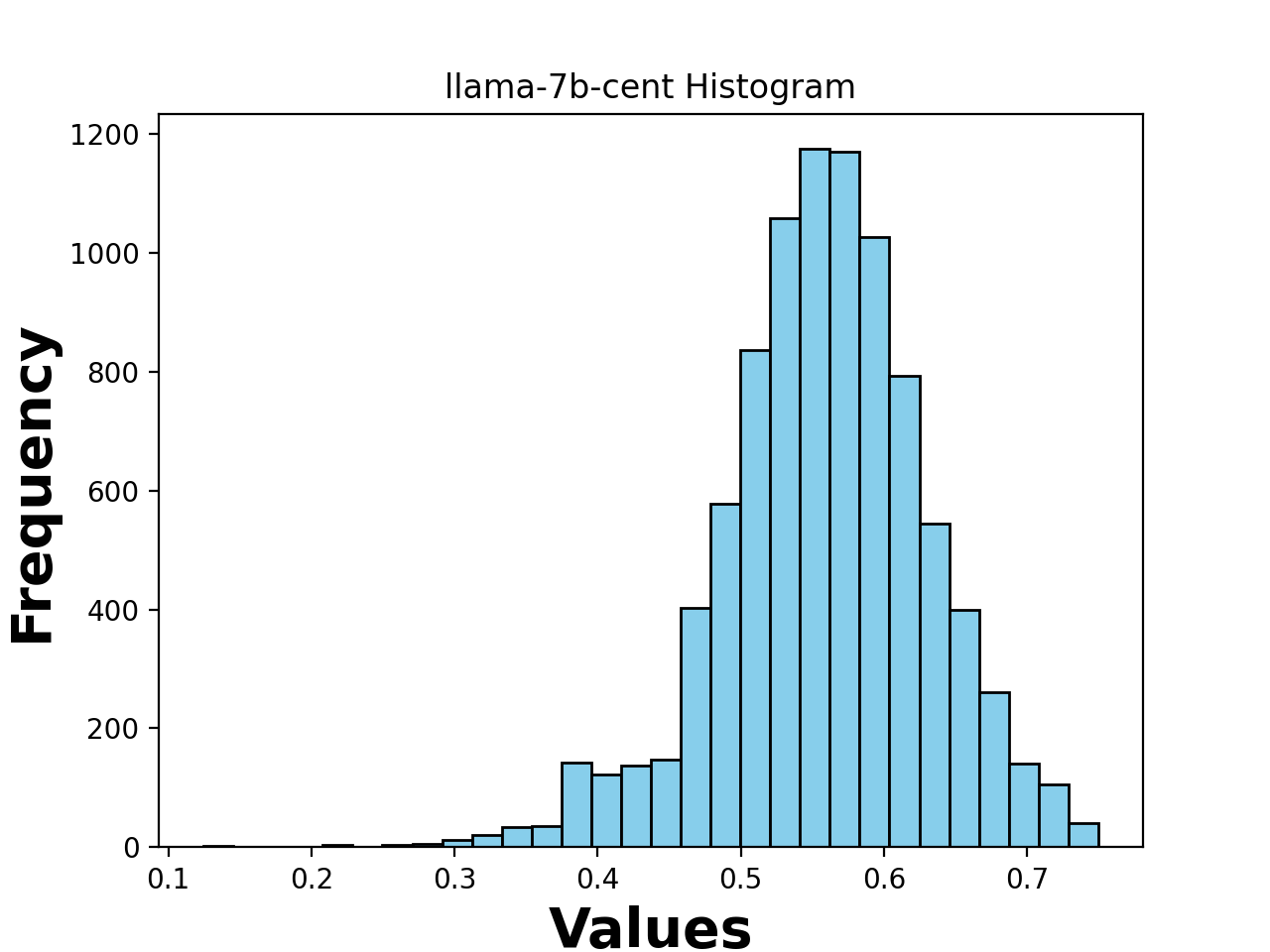}
     \end{subfigure}
     \begin{subfigure}[b]{0.245\textwidth}
         \centering
         \includegraphics[width=\textwidth]{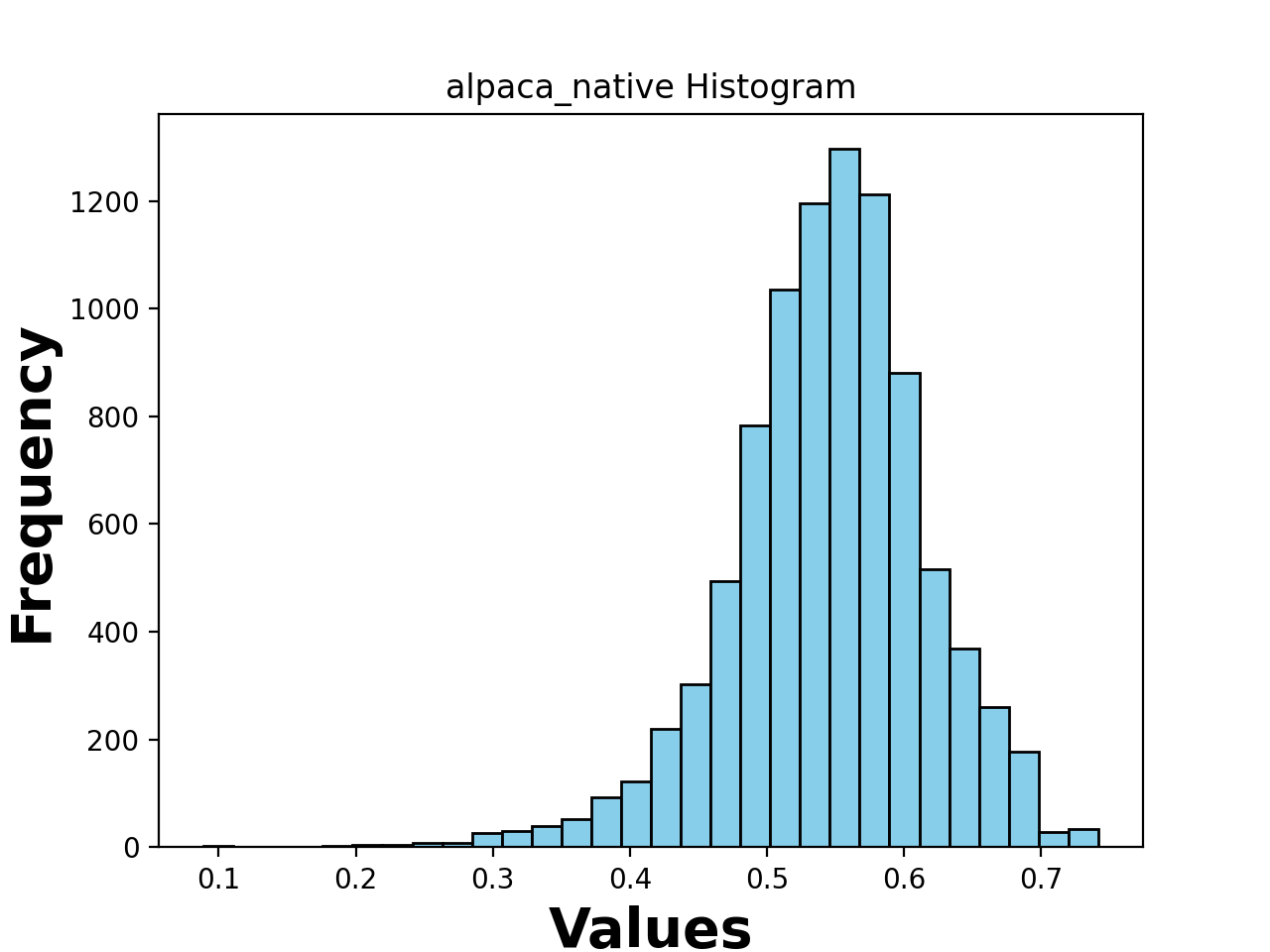}
     \end{subfigure}
     \begin{subfigure}[b]{0.245\textwidth}
         \centering
         \includegraphics[width=\textwidth]{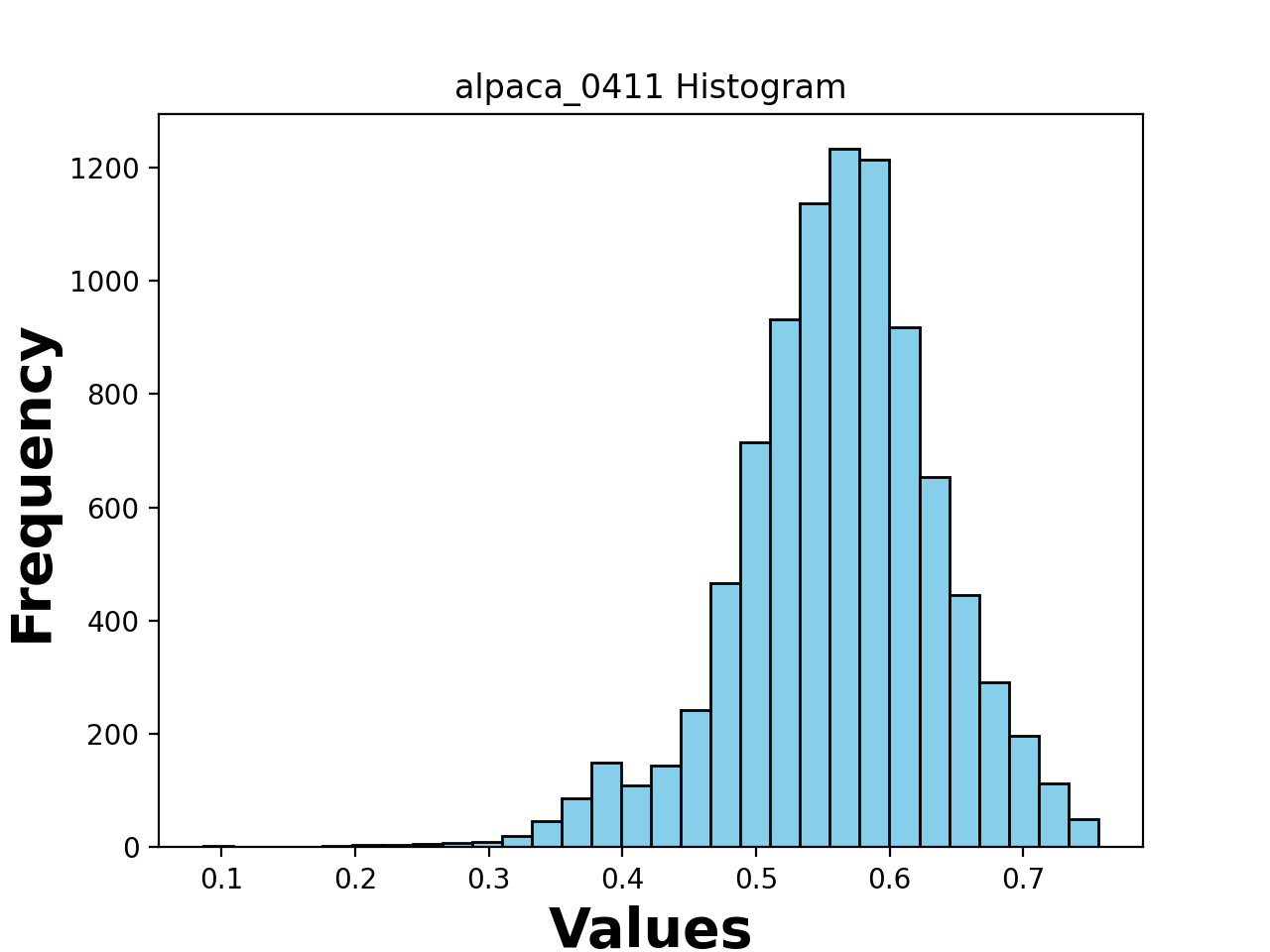}
     \end{subfigure}
     \begin{subfigure}[b]{0.245\textwidth}
         \centering
         \includegraphics[width=\textwidth]{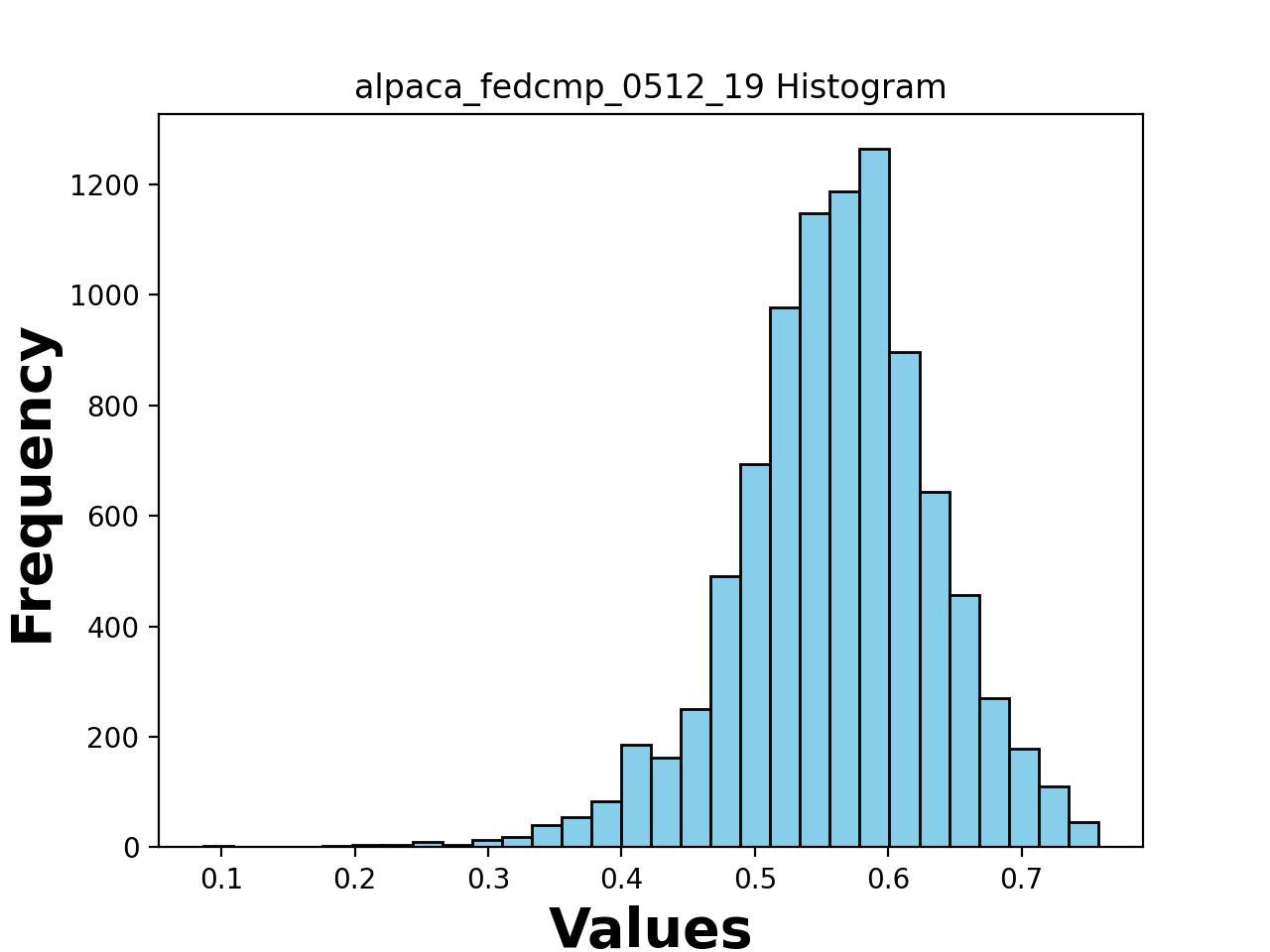}
     \end{subfigure}
     \begin{subfigure}[b]{0.245\textwidth}
         \centering
         \includegraphics[width=\textwidth]{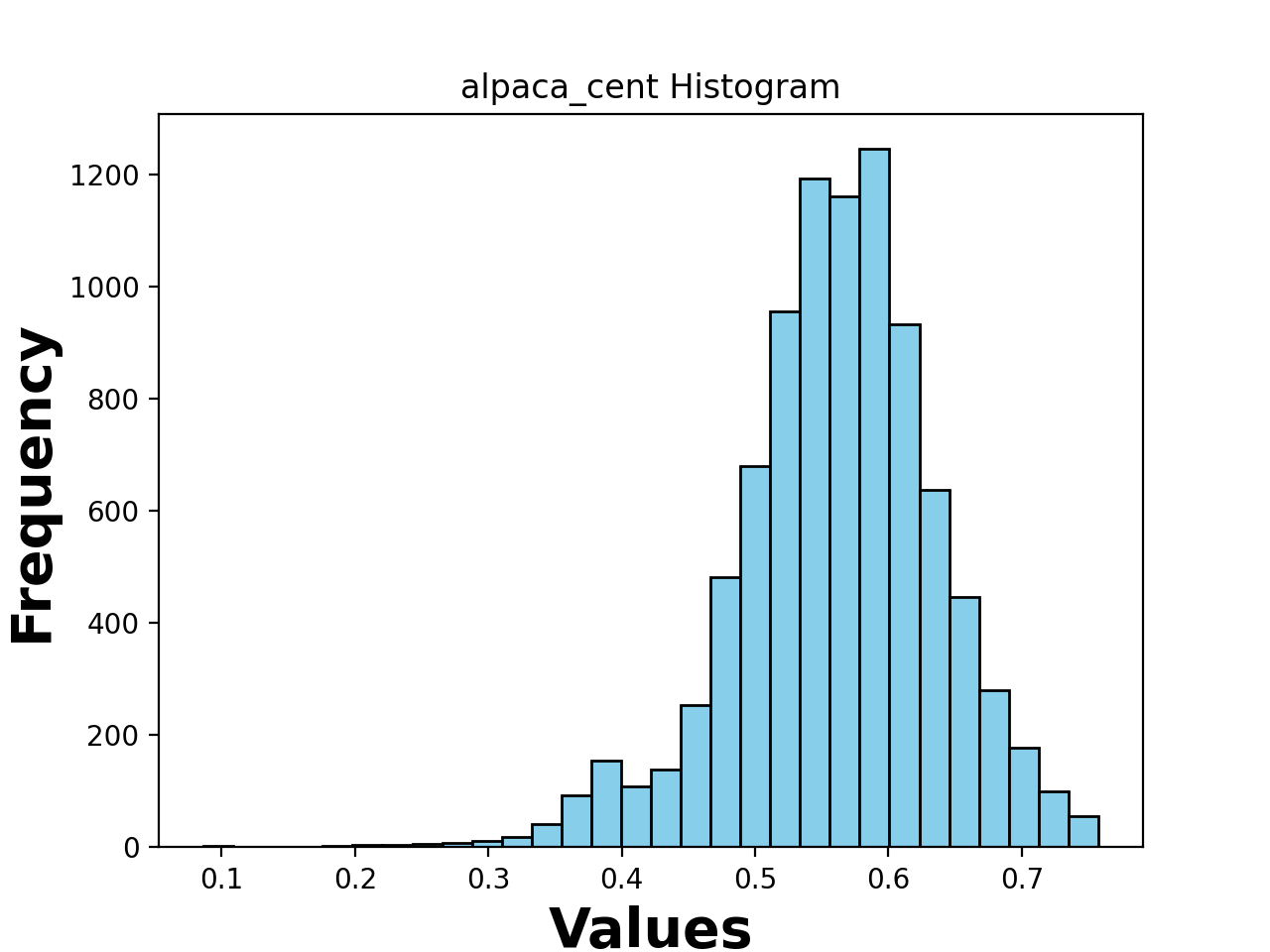}
     \end{subfigure}
\caption{Matrix Entropy Distribution across Datasets on \textbf{LLaMA-7B} and \textbf{Alpaca-7B} for both foundation and fine-tuned models. The upper row displays the matrix entropy distribution for \textbf{LLaMA-7B} associated models, whereas the bottom row details the same for \textbf{Alpaca-7B} foundation model. The distributions are sequentially arranged from left to right as \textbf{Base}, \textbf{LoRA-FT}, \textbf{Compress-FT}, and \textbf{Cent}. }
\vspace{0.25cm}
\label{fig:matrix_entropy_dist}
\end{figure*}

Furthermore, we display the matrix entropy distributions for both the foundation and the fine-tuned models using the \textbf{Databricks-dolly15k} dataset~\cite{conover2023free} as shown in Figure~\ref{fig:matrix_entropy_dist}. Most of the matrix entropy values cluster between $0.55$ and $0.60$, exhibiting a shape similar to that of a Gaussian distribution. 

\subsection{Model Evaluation by QWen API}
\label{sec:append_qwen}

In this section, we discuss the evaluations conducted using the QWen LLM API~\cite{bai2023qwen}. The complete process for employing the Qwen model for evaluation involves the following phases. Initially, we used foundational models or fine-tuned models to produce responses for \textbf{CNN-DailyMail News Text Summarization}~\cite{chen2017examination}. Subsequently, answer sheets are created from the outcomes of the foundational and fine-tuned models, along with standardized answers, which are then fed into the Qwen API. Following this, the Qwen API assesses the answer sheets prepared in the previous step and delivers the assessment outcomes. The ultimate step involves calculating the statistical analyses of these assessment outcomes. The evaluation by Qwen encompasses four aspects: 'Relevance, Fluency, Coherence, and Consistency.'
We show the results of the evaluation based on the Qwen API in Table~\ref{tab:qwen_results}. \textbf{Gold-Reference} denotes the scores assigned to the standard responses. \textbf{Relevance} is about selecting significant content from the source, ensuring that the summary incorporates essential information from the original document. \textbf{Fluency} concerns the quality of each sentence, ensuring that no sentence in the summary suffers from formatting issues, synchronization errors, or grammatical mistakes. \textbf{Coherence} examines the overall quality of all sentences, assessing if the summary is structured and logical. \textbf{Consistency} checks the factual agreement between the summary and the original sources, which means that a factually consistent summary should only include information depicted in the source document.

Our ongoing analysis of the Qwen scores, as shown in Table~\ref{tab:qwen_results}, begins by noting that the \textbf{Gold-Reference} achieves the highest scores from the four perspectives, a result of the tailored evaluation methodology in Qwen. Moreover, our method \textbf{Compress-FT} consistently delivers stable results in all evaluation dimensions, outperforming \textbf{LoRA-FT} but not matching the performance of \textbf{Gold-Reference} or \textbf{Cent}. 
Additionally, our approach \textbf{Compress-FT} demonstrates a decrease in ``Standard Deviation'' relative to \textbf{LoRA-FT}, indicating improved stability. 
In summary, the evaluation of the Qwen API indicates that the fine-tuned models surpass the foundational models in all aspects of the evaluation. Furthermore, our strategy \textbf{Compress-FT} not only advances the performance beyond \textbf{LoRA-FT} but also achieves parity with \textbf{Cent}.

\begin{table*}[htbp]
\centering
\caption{Qwen API Evaluation Results}
\label{tab:qwen_results}
\begin{tabular}{ccc} 
\toprule
Methods & Average & Standard Deviation \\
\midrule
\textbf{Relevance-Gold-Reference} & 4.032 & 0.405 \\
\textbf{Relevance-Cent-LLaMA} & 3.505 & 0.562\\
\textbf{Relevance-Compress-FT-LLaMA} (ours) & 3.462 & 0.684 \\
\textbf{Relevance-LoRA-FT-LLaMA} & 3.395 & 0.850\\
\midrule 
\textbf{Fluency-Gold-Reference} & 4.055 & 0.228 \\
\textbf{Fluency-Cent-LLaMA} & 4.022 & 0.209\\
\textbf{Fluency-Compress-FT-LLaMA} (ours) & 3.835 & 0.560 \\
\textbf{Fluency-LoRA-FT-LLaMA} & 3.769 & 0.576\\
\midrule 
\textbf{Coherence-Gold-Reference} & 3.835 & 0.400 \\
\textbf{Coherence-Cent-LLaMA} & 3.077 & 0.425\\
\textbf{Coherence-Compress-FT-LLaMA} (ours) & 2.989 & 0.584\\
\textbf{Coherence-LoRA-FT-LLaMA} & 2.934 & 0.708\\
\midrule 
\textbf{Consistency-Gold-Reference} & 4.055 & 0.343\\
\textbf{Consistency-Cent-LLaMA} & 3.527 & 0.561\\
\textbf{Consistency-Compress-FT-LLaMA} (ours) & 3.396	& 0.709\\
\textbf{Consistency-LoRA-FT-LLaMA} & 3.308 & 0.821\\
\bottomrule
\end{tabular}
\end{table*}

\section{More Experimental Analysis}

\subsection{Gradients Low-rank Decomposition Distributions}
\begin{figure*}[htbp]
     \begin{subfigure}[b]{0.24\textwidth}
         \centering
         \includegraphics[width=\textwidth]{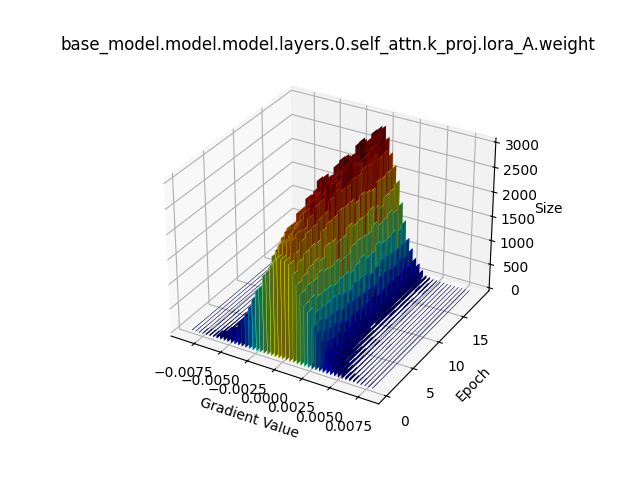}
     \end{subfigure}
     \begin{subfigure}[b]{0.24\textwidth}
         \centering
         \includegraphics[width=\textwidth]{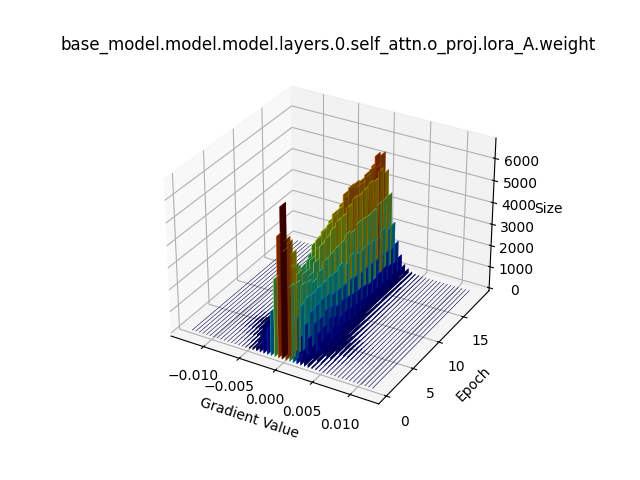}
     \end{subfigure}
     \begin{subfigure}[b]{0.24\textwidth}
         \centering
         \includegraphics[width=\textwidth]{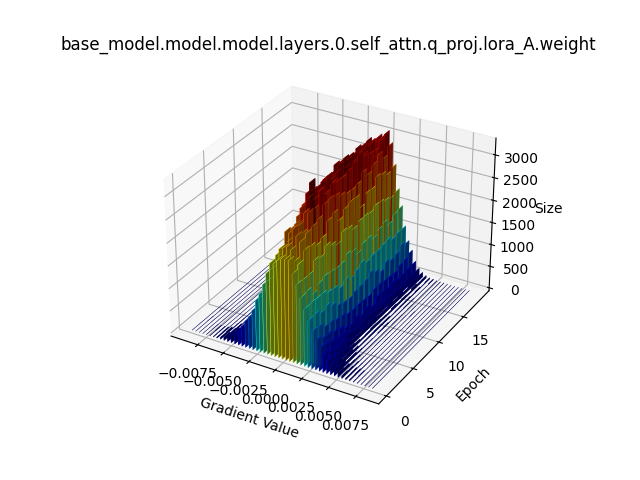}
     \end{subfigure}
     \begin{subfigure}[b]{0.24\textwidth}
         \centering
         \includegraphics[width=\textwidth]{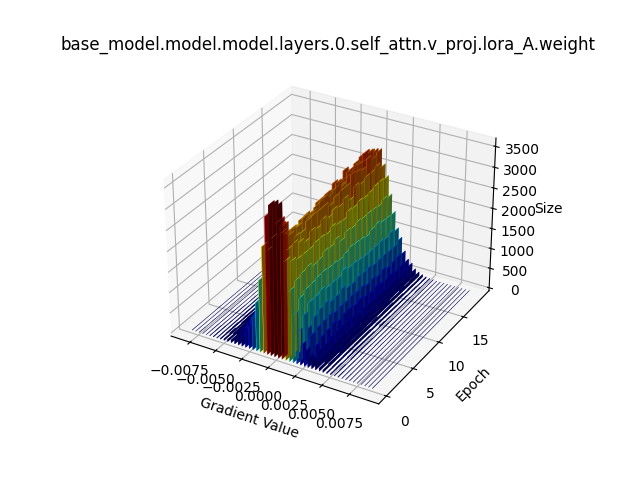}
     \end{subfigure}
     \begin{subfigure}[b]{0.24\textwidth}
         \centering
         \includegraphics[width=\textwidth]{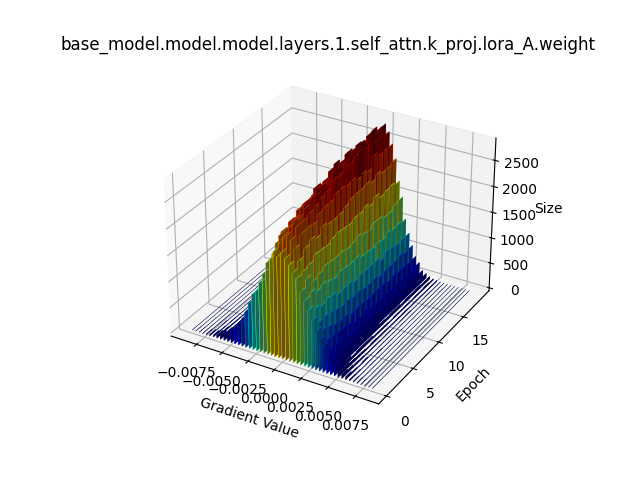}
     \end{subfigure}
     \begin{subfigure}[b]{0.24\textwidth}
         \centering
         \includegraphics[width=\textwidth]{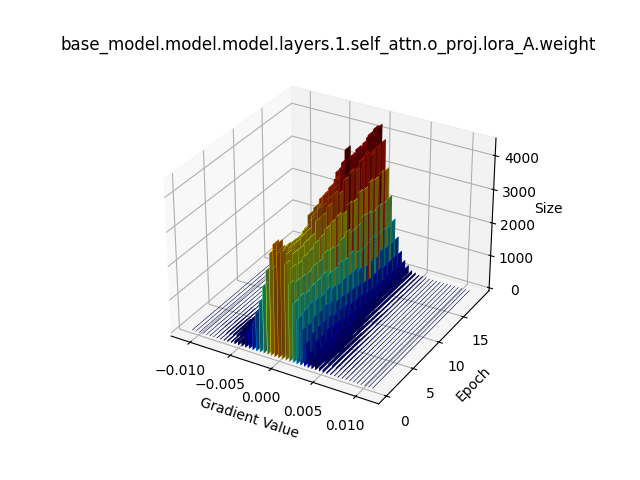}
     \end{subfigure}
     \begin{subfigure}[b]{0.24\textwidth}
         \centering
         \includegraphics[width=\textwidth]{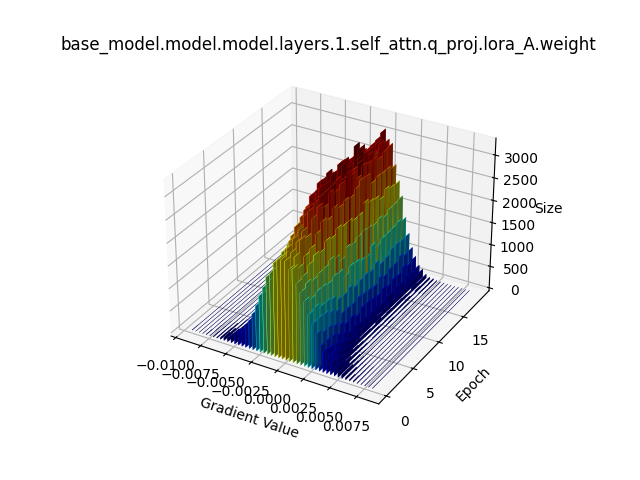}
     \end{subfigure}
     \begin{subfigure}[b]{0.24\textwidth}
         \centering
         \includegraphics[width=\textwidth]{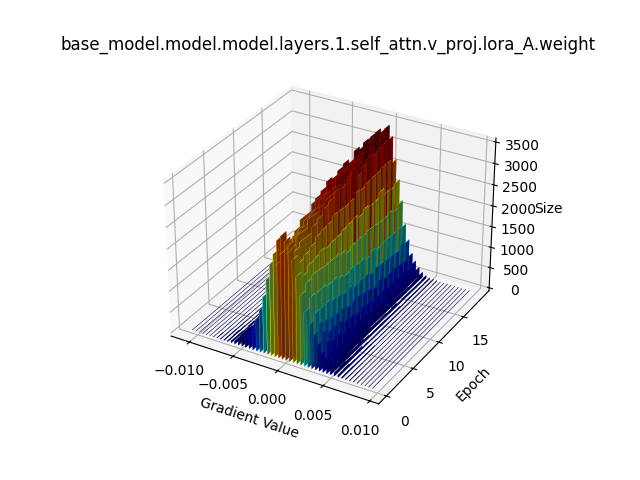}
     \end{subfigure}
     \begin{subfigure}[b]{0.24\textwidth}
         \centering
         \includegraphics[width=\textwidth]{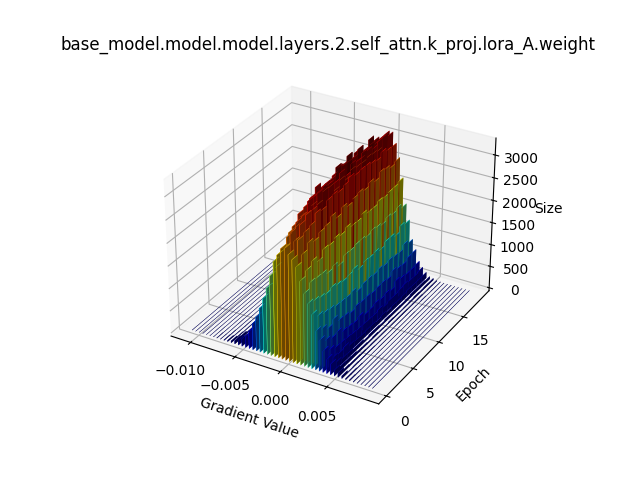}
     \end{subfigure}
     \begin{subfigure}[b]{0.24\textwidth}
         \centering
         \includegraphics[width=\textwidth]{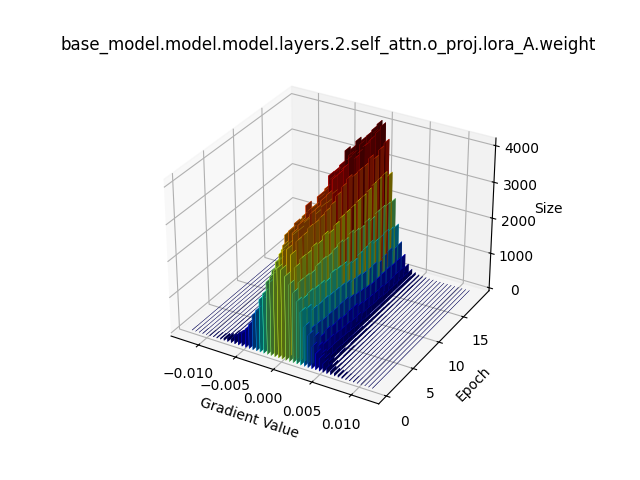}
     \end{subfigure}
     \begin{subfigure}[b]{0.24\textwidth}
         \centering
         \includegraphics[width=\textwidth]{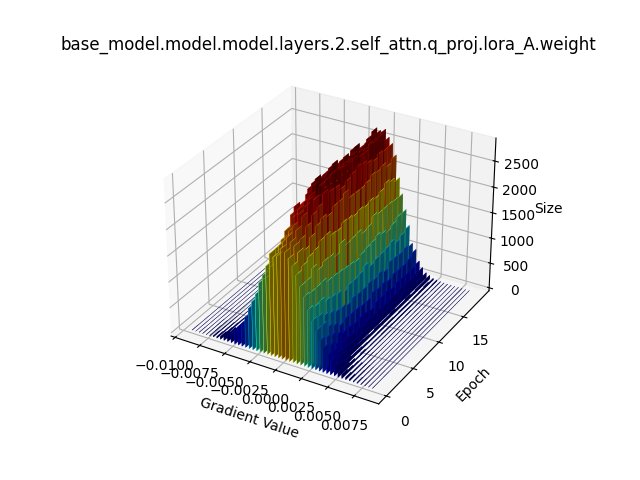}
     \end{subfigure}
     \begin{subfigure}[b]{0.24\textwidth}
         \centering
         \includegraphics[width=\textwidth]{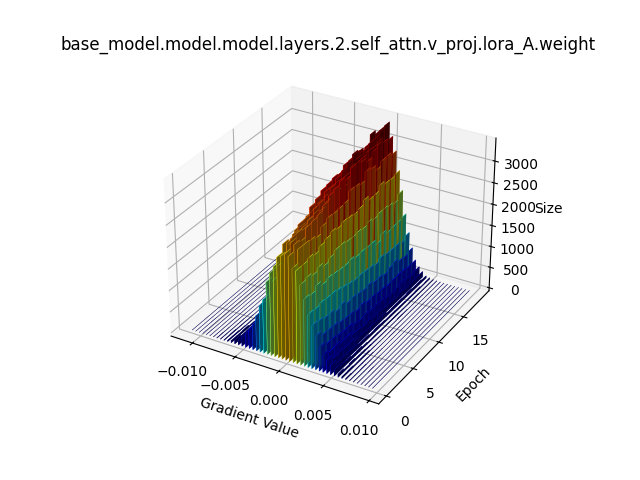}
     \end{subfigure}
     \begin{subfigure}[b]{0.24\textwidth}
         \centering
         \includegraphics[width=\textwidth]{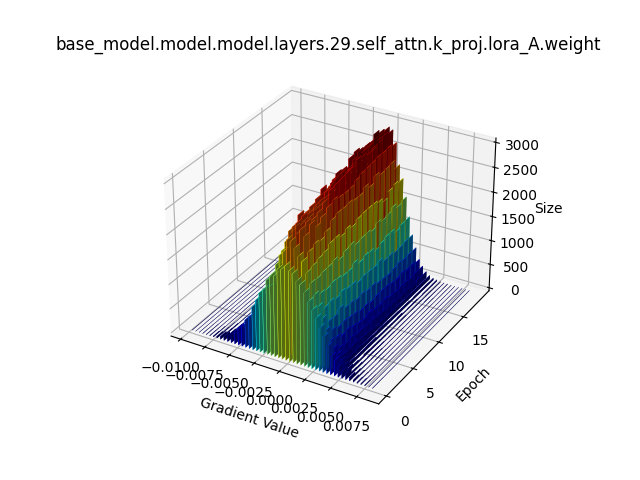}
     \end{subfigure}
     \begin{subfigure}[b]{0.24\textwidth}
         \centering
         \includegraphics[width=\textwidth]{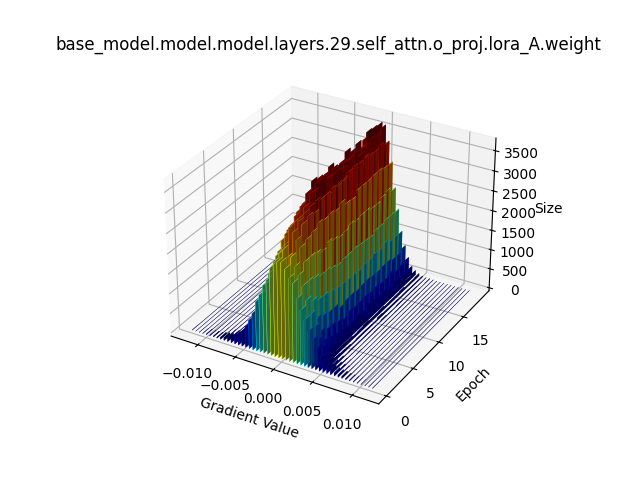}
     \end{subfigure}
     \begin{subfigure}[b]{0.24\textwidth}
         \centering
         \includegraphics[width=\textwidth]{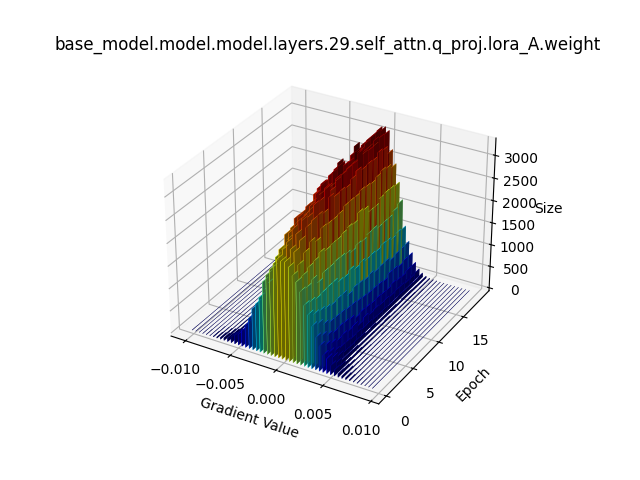}
     \end{subfigure}
     \begin{subfigure}[b]{0.24\textwidth}
         \centering
         \includegraphics[width=\textwidth]{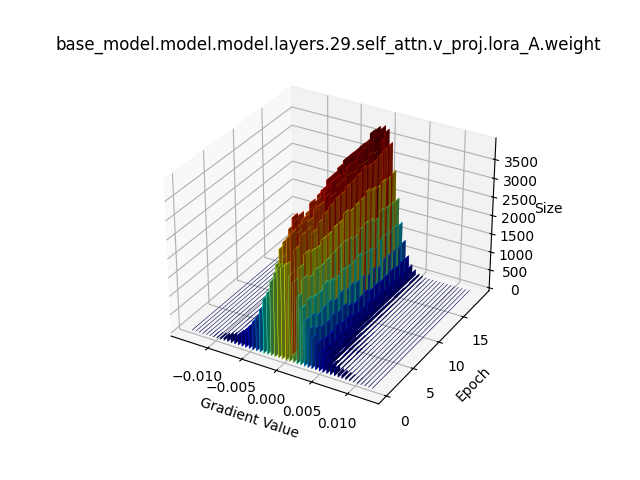}
     \end{subfigure}
     \begin{subfigure}[b]{0.24\textwidth}
         \centering
         \includegraphics[width=\textwidth]{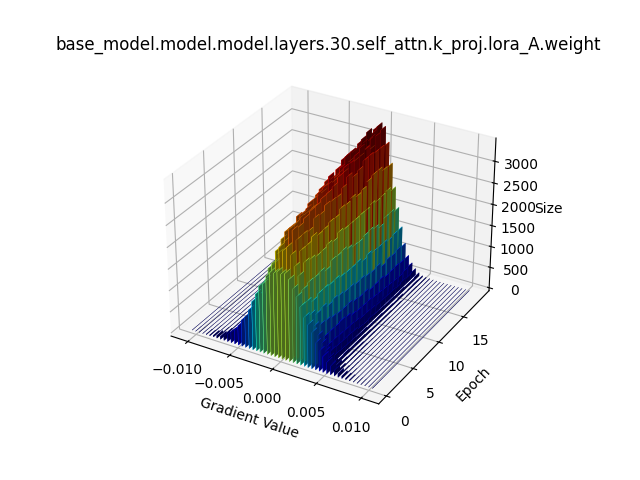}
     \end{subfigure}
     \begin{subfigure}[b]{0.24\textwidth}
         \centering
         \includegraphics[width=\textwidth]{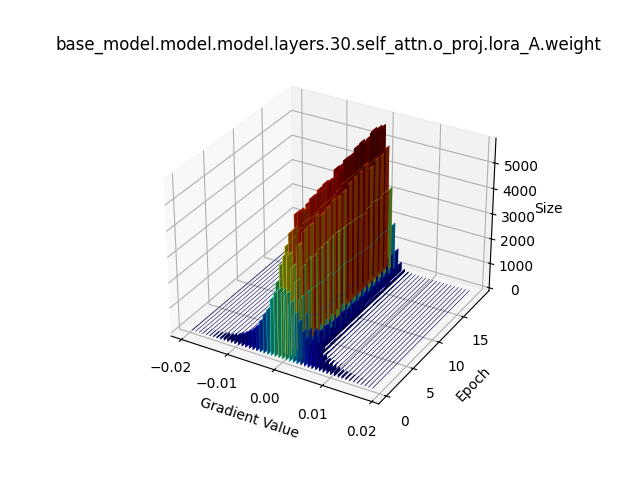}
     \end{subfigure}
     \begin{subfigure}[b]{0.24\textwidth}
         \centering
         \includegraphics[width=\textwidth]{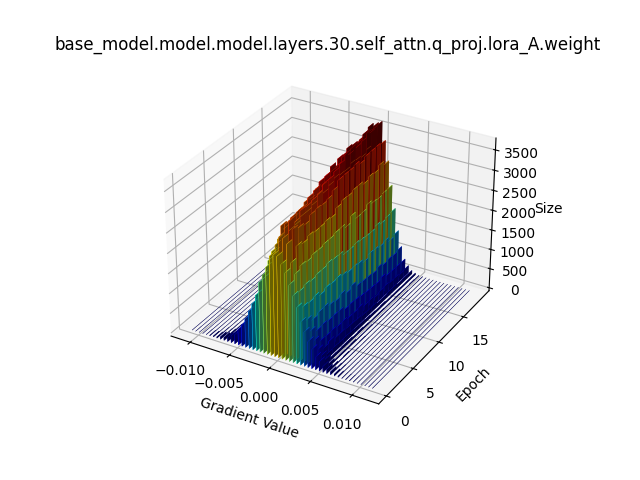}
     \end{subfigure}
     \begin{subfigure}[b]{0.24\textwidth}
         \centering
         \includegraphics[width=\textwidth]{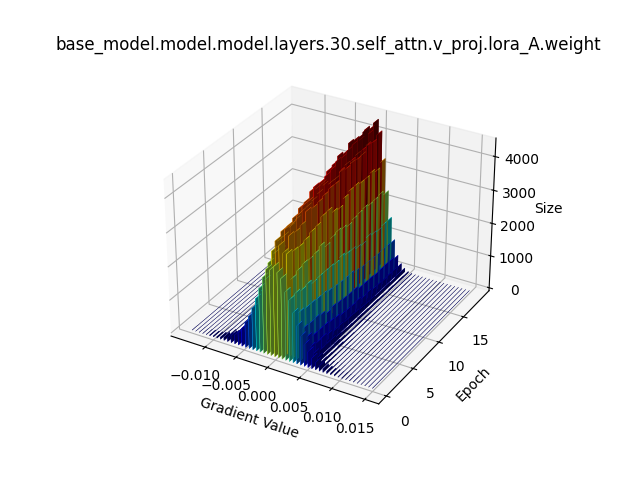}
     \end{subfigure}
     \begin{subfigure}[b]{0.24\textwidth}
         \centering
         \includegraphics[width=\textwidth]{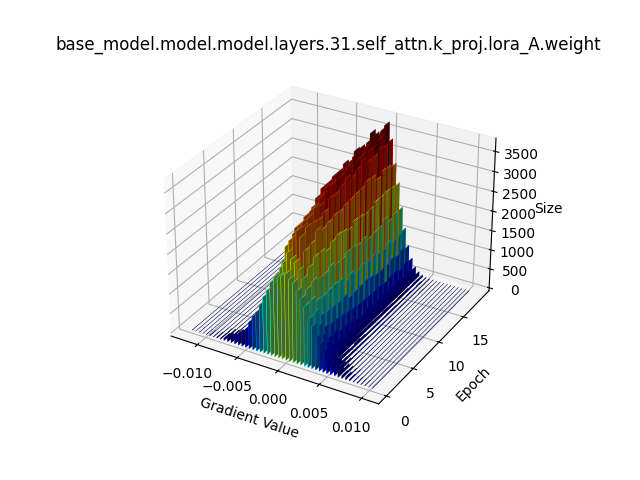}
     \end{subfigure}
     \begin{subfigure}[b]{0.24\textwidth}
         \centering
         \includegraphics[width=\textwidth]{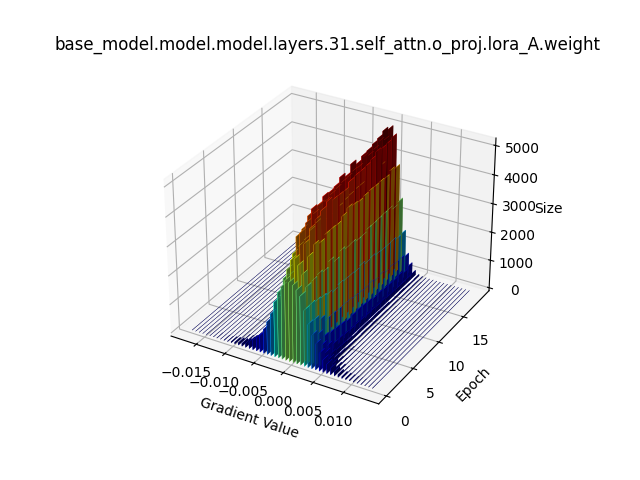}
     \end{subfigure}
     \begin{subfigure}[b]{0.24\textwidth}
         \centering
         \includegraphics[width=\textwidth]{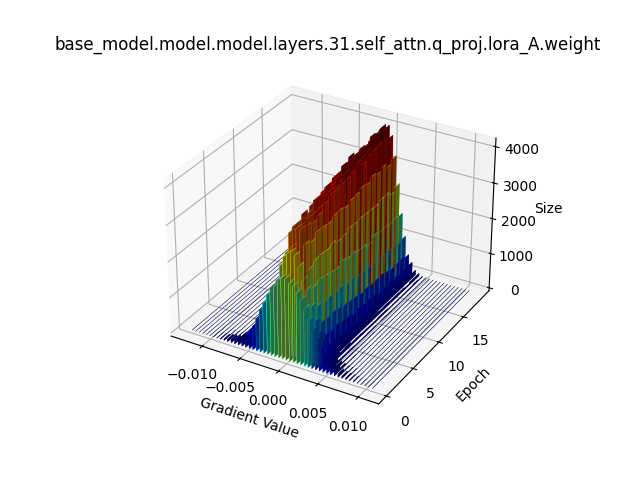}
     \end{subfigure}
     \begin{subfigure}[b]{0.24\textwidth}
         \centering
         \includegraphics[width=\textwidth]{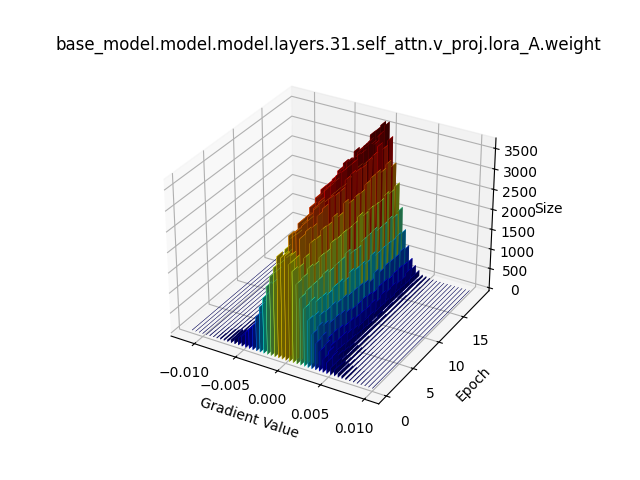}
     \end{subfigure}
\caption{The behavior of low-rank decomposition matrices $\mathbf{A}$ across various layers during iterative training sessions. The x-axis represents the range of values for the matrices. The y-axis indicates the iteration steps and the z-axis shows the frequency of the matrix values. From the upper row to the bottom row, the distribution of $\mathbf{A}$ in layers $\{0,1,2,29,30,31\}$ is presented. Horizontally, from left to right, the distribution of $\mathbf{A}$ for the attention matrices $\mathbf{K}, \mathbf{Q}, \mathbf{V}, \mathbf{O}$ is displayed, respectively.}
\label{fig:A_dist}
\end{figure*}

\begin{figure*}[htbp]
     \begin{subfigure}[b]{0.25\textwidth}
         \centering
         \includegraphics[width=\textwidth]{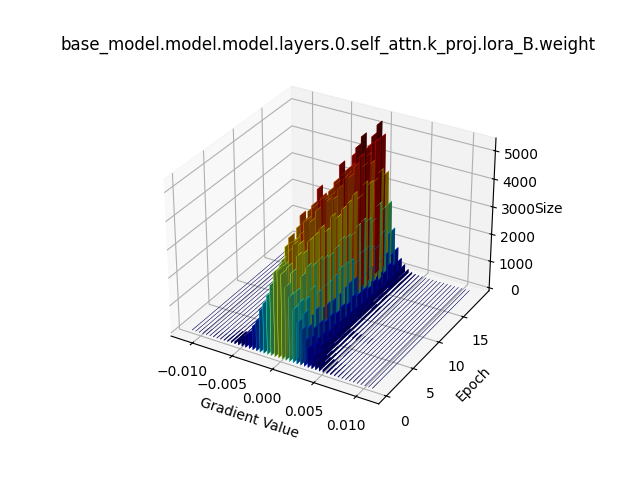}
     \end{subfigure}
     \begin{subfigure}[b]{0.245\textwidth}
         \centering
         \includegraphics[width=\textwidth]{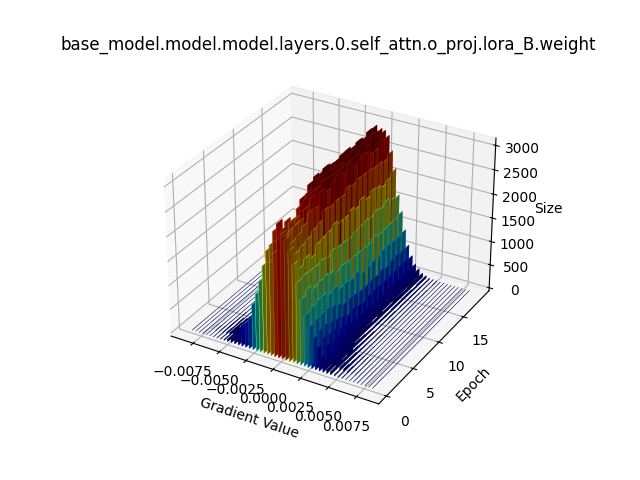}
     \end{subfigure}
     \begin{subfigure}[b]{0.245\textwidth}
         \centering
         \includegraphics[width=\textwidth]{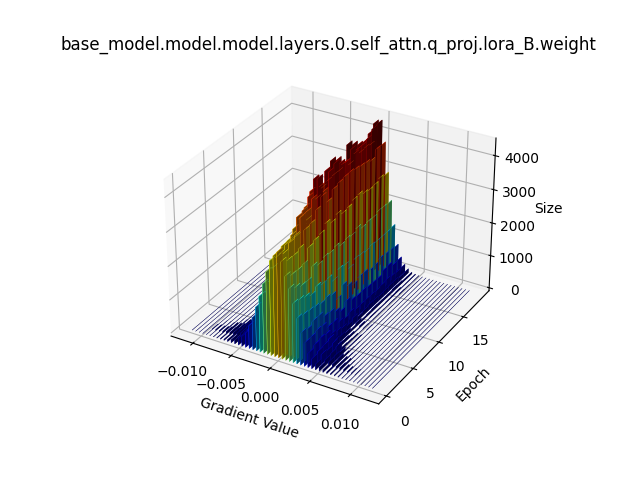}
     \end{subfigure}
     \begin{subfigure}[b]{0.245\textwidth}
         \centering
         \includegraphics[width=\textwidth]{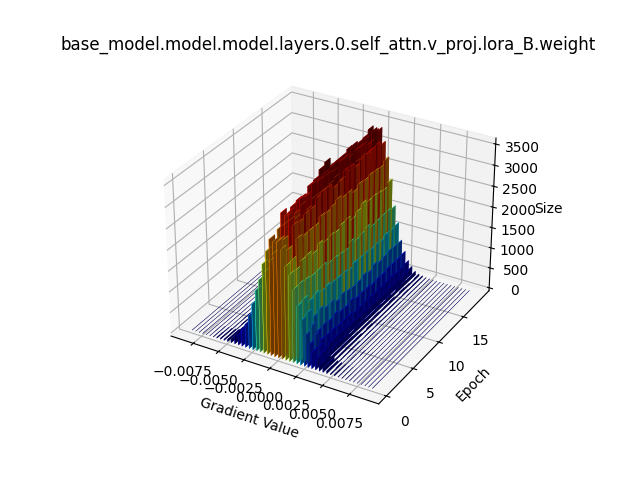}
     \end{subfigure}
     \begin{subfigure}[b]{0.245\textwidth}
         \centering
         \includegraphics[width=\textwidth]{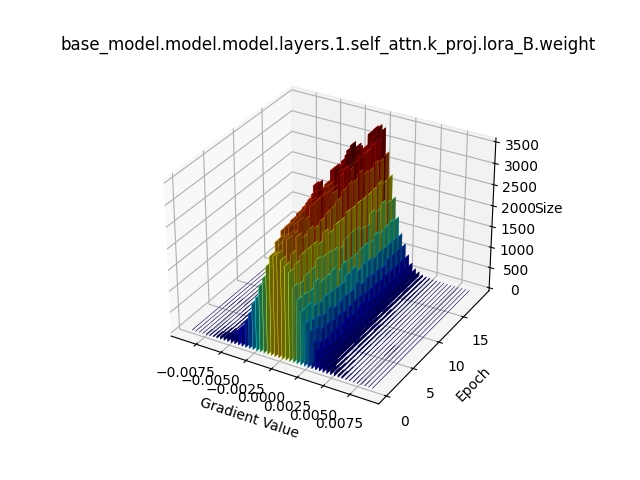}
     \end{subfigure}
     \begin{subfigure}[b]{0.245\textwidth}
         \centering
         \includegraphics[width=\textwidth]{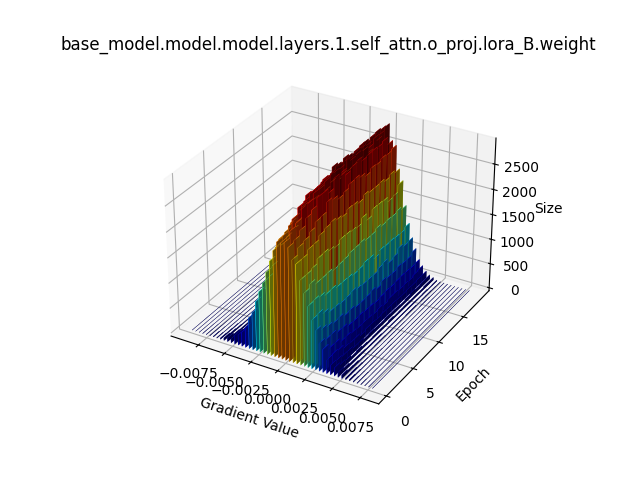}
     \end{subfigure}
     \begin{subfigure}[b]{0.245\textwidth}
         \centering
         \includegraphics[width=\textwidth]{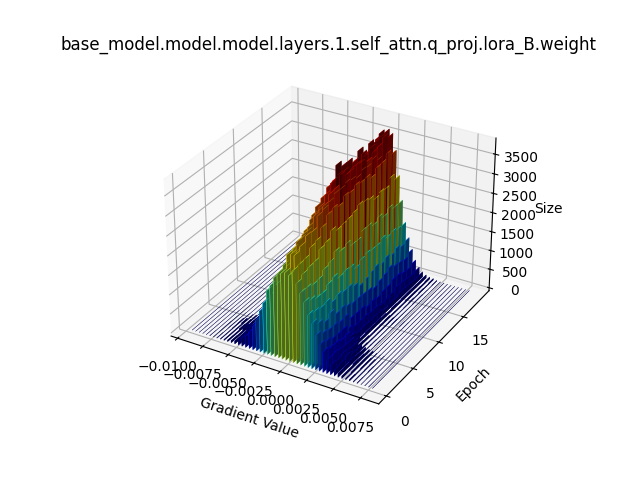}
     \end{subfigure}
     \begin{subfigure}[b]{0.245\textwidth}
         \centering
         \includegraphics[width=\textwidth]{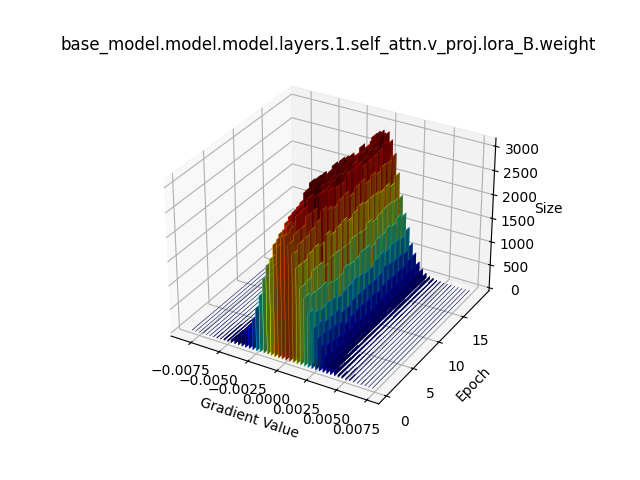}
     \end{subfigure}
     \begin{subfigure}[b]{0.245\textwidth}
         \centering
         \includegraphics[width=\textwidth]{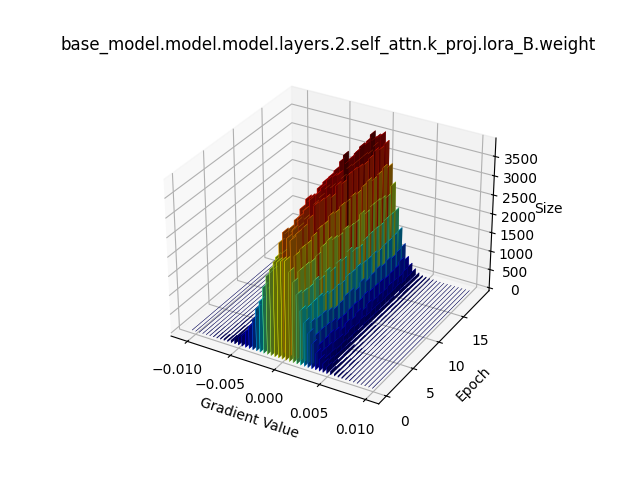}
     \end{subfigure}
     \begin{subfigure}[b]{0.245\textwidth}
         \centering
         \includegraphics[width=\textwidth]{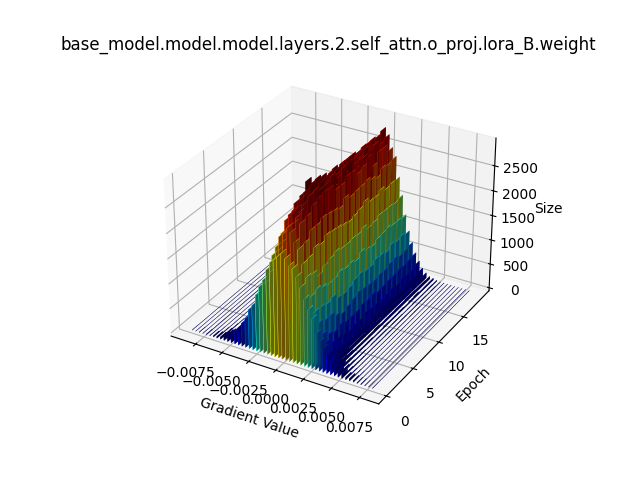}
     \end{subfigure}
     \begin{subfigure}[b]{0.245\textwidth}
         \centering
         \includegraphics[width=\textwidth]{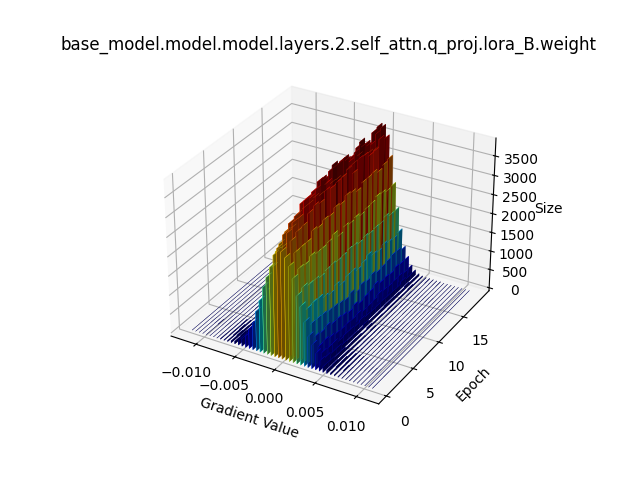}
     \end{subfigure}
     \begin{subfigure}[b]{0.245\textwidth}
         \centering
         \includegraphics[width=\textwidth]{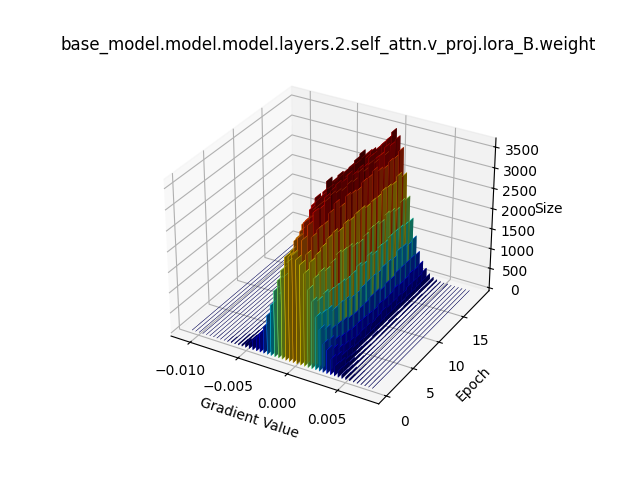}
     \end{subfigure}
     \begin{subfigure}[b]{0.245\textwidth}
         \centering
         \includegraphics[width=\textwidth]{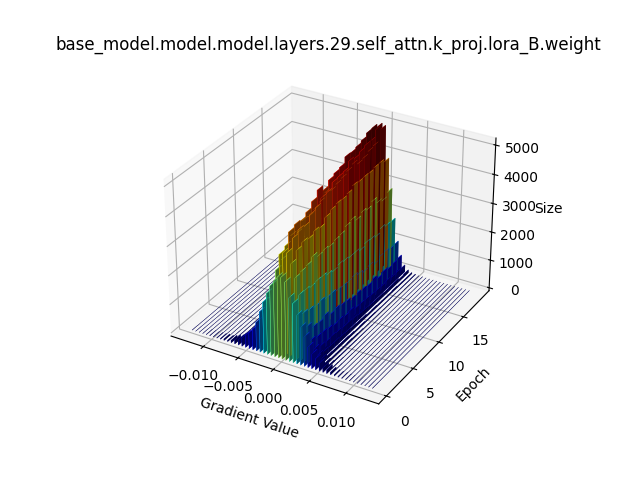}
     \end{subfigure}
     \begin{subfigure}[b]{0.245\textwidth}
         \centering
         \includegraphics[width=\textwidth]{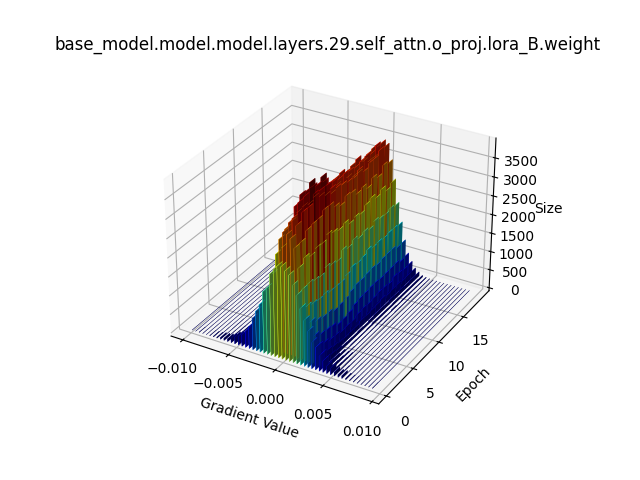}
     \end{subfigure}
     \begin{subfigure}[b]{0.245\textwidth}
         \centering
         \includegraphics[width=\textwidth]{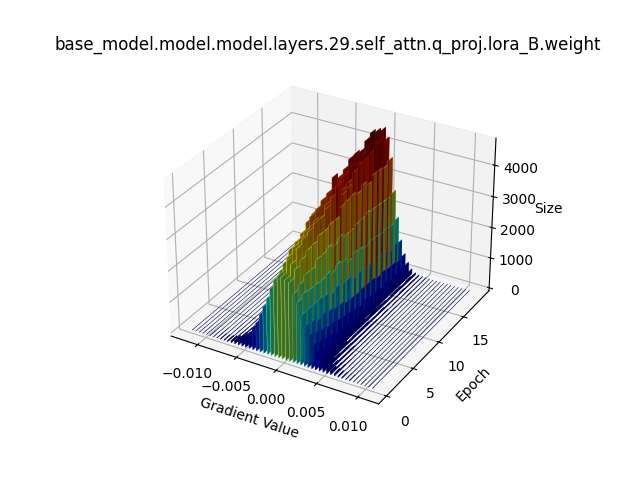}
     \end{subfigure}
     \begin{subfigure}[b]{0.245\textwidth}
         \centering
         \includegraphics[width=\textwidth]{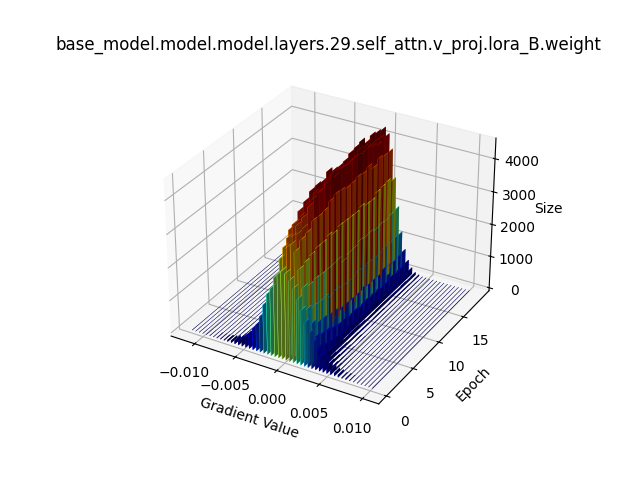}
     \end{subfigure}
     \begin{subfigure}[b]{0.245\textwidth}
         \centering
         \includegraphics[width=\textwidth]{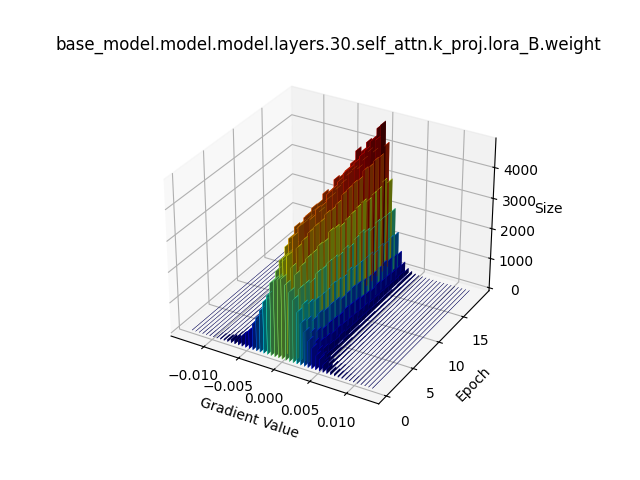}
     \end{subfigure}
     \begin{subfigure}[b]{0.245\textwidth}
         \centering
         \includegraphics[width=\textwidth]{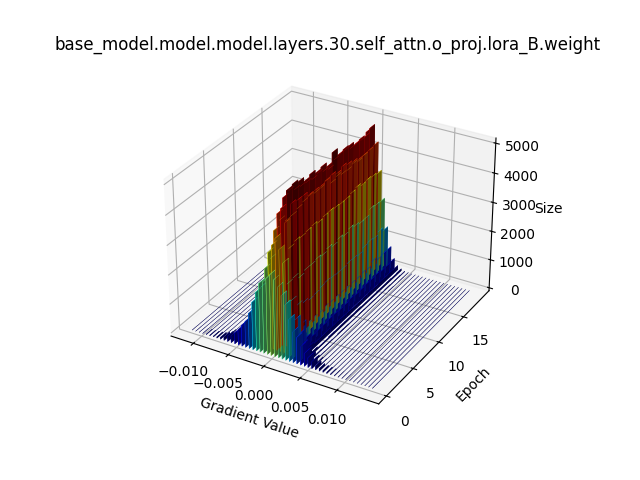}
     \end{subfigure}
     \begin{subfigure}[b]{0.245\textwidth}
         \centering
         \includegraphics[width=\textwidth]{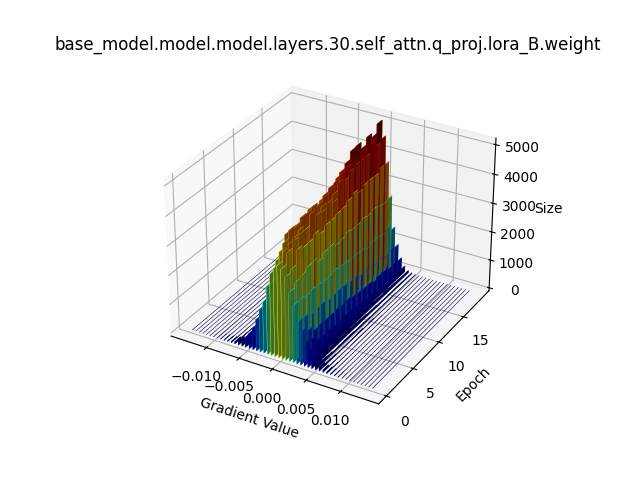}
     \end{subfigure}
     \begin{subfigure}[b]{0.245\textwidth}
         \centering
         \includegraphics[width=\textwidth]{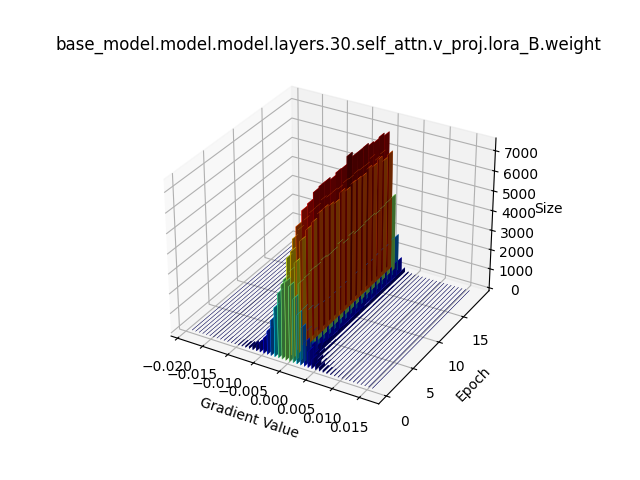}
     \end{subfigure}
     \begin{subfigure}[b]{0.245\textwidth}
         \centering
         \includegraphics[width=\textwidth]{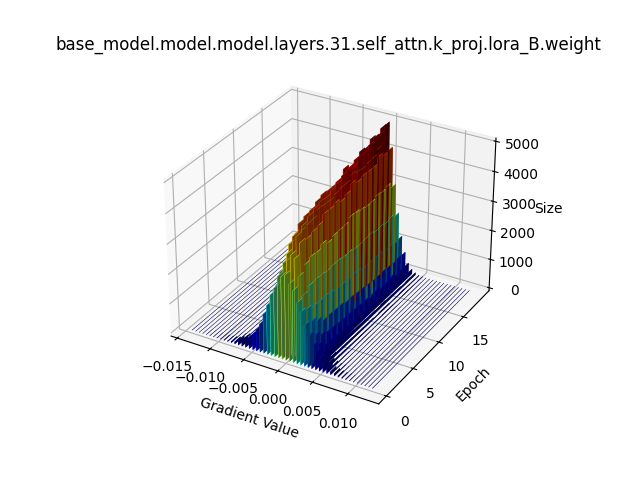}
     \end{subfigure}
     \begin{subfigure}[b]{0.245\textwidth}
         \centering
         \includegraphics[width=\textwidth]{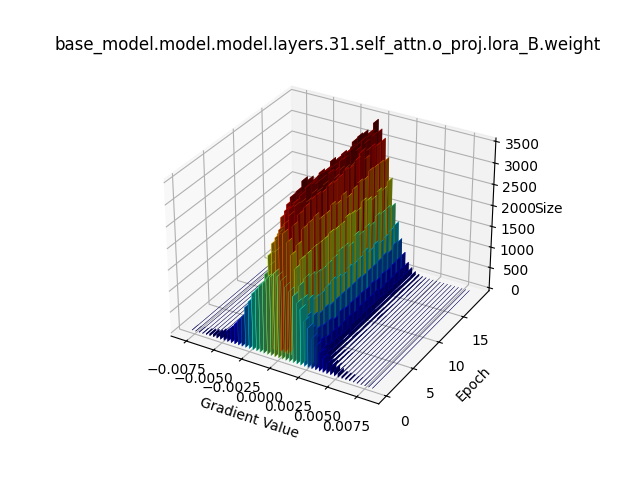}
     \end{subfigure}
     \begin{subfigure}[b]{0.245\textwidth}
         \centering
         \includegraphics[width=\textwidth]{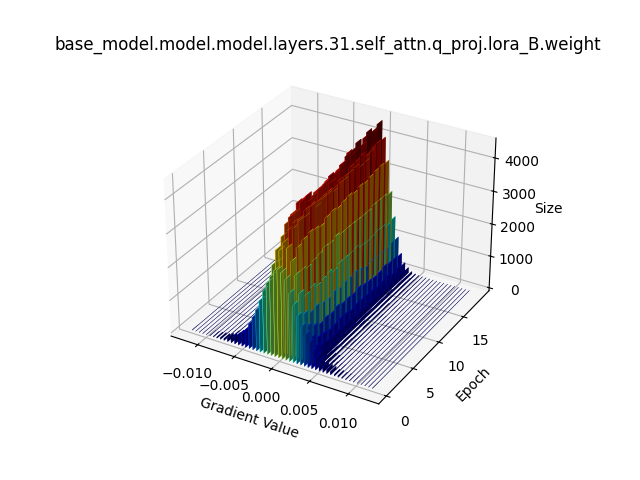}
     \end{subfigure}
     \begin{subfigure}[b]{0.245\textwidth}
         \centering
         \includegraphics[width=\textwidth]{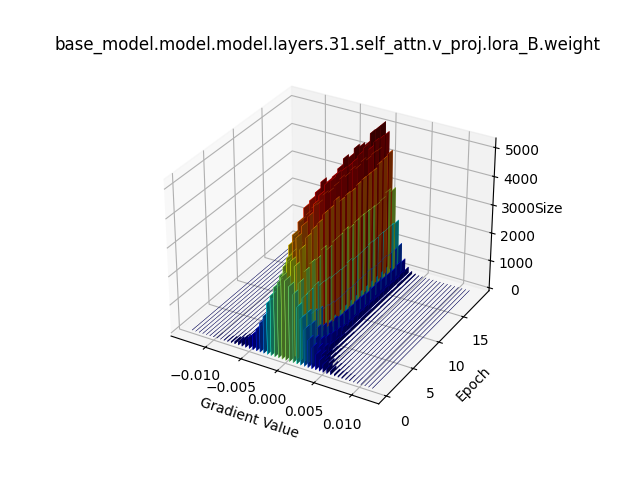}
     \end{subfigure}
\caption{The distribution patterns of low-rank decomposition matrices $\mathbf{B}$ across various layers during iterative training are depicted. The x-axis represents the range of values for the matrices. The y-axis indicates the iteration steps and the z-axis shows the frequency of the values of the matrices. From the top row to the bottom row, the distribution of $\mathbf{B}$ is presented for layers $\{0,1,2,29,30,31\}$. Horizontally, from left to right, the distributions of $\mathbf{B}$ for the attention matrices $\mathbf{K}, \mathbf{Q}, \mathbf{V}, \mathbf{O}$ are displayed, respectively.}
\label{fig:B_dist}
\end{figure*}

In this section, we introduce the parameter distributions for the low-rank matrices $\mathbf{A}$ and $\mathbf{B}$ in each layer within the Transformer blocks, which serve as input data for training \stageone. 
For the \textbf{LlaMA-7B} foundation models, there are 32 layers of the Transformer block; each block consists of four attention matrices $\mathbf{K}, \mathbf{Q}, \mathbf{V}$ and $\mathbf{O}$. 
Each attention matrix has the shape $4096$. 
Let $\mathbf{G}_K, \mathbf{G}_Q, \mathbf{G}_V, \mathbf{G}_O$ be the gradients corresponding to $\mathbf{K}, \mathbf{Q}, \mathbf{V}$ and $\mathbf{O}$. 
In \textbf{LoRA} training, we decompose the transformer gradients $\mathbf{G}_K, \mathbf{G}_Q, \mathbf{G}_V, \mathbf{G}_O$ into the product of two low-rank matrices
\begin{align*}
\mathbf{G}_{K} &= \mathbf{B}_{K} \mathbf{A}_{K}; \\
\mathbf{G}_{Q} &= \mathbf{B}_{Q} \mathbf{A}_{Q}; \\
\mathbf{G}_{V} &= \mathbf{B}_{V} \mathbf{A}_{V}; \\
\mathbf{G}_{O} &= \mathbf{B}_{O} \mathbf{A}_{O};
\end{align*}
where $\mathbf{A}$ is in the shape of $8 \times 4096$ and $\mathbf{B}$ is in the shape of $4096 \times 8$ if we choose the low-rank parameters as $8$. 
Consequently, the total count for each of the matrices $\mathbf{A}$ and $\mathbf{B}$ is 128. These matrices are collected over 20 training epochs and are used as input for training an AutoEncoder. The primary objective is to enable the AutoEncoder to identify common features in temporal gradients, which are crucial for federated fine-tuning. This section presents visualizations of the inputs $\mathbf{A}$ and $\mathbf{B}$ in Figure~\ref{fig:A_dist} and Figure~\ref{fig:B_dist}, respectively. Observations from Figure~\ref{fig:A_dist} indicate that the distribution of most matrix values resembles a Gaussian distribution, predominantly centered around 0, with an x-axis range from $-0.01$ to $0.01$. Furthermore, in most layers, as iterations increase, the matrix values tend to cluster more tightly around $0$, resulting in a higher peak and a narrower x-axis range. A similar pattern is observed in the $\mathbf{B}$ matrices shown in Figure~\ref{fig:B_dist}. The consistency across different layers and training epochs of $A$ and $B$ underscores the efficacy of the trained AutoEncoder.

\begin{table*}
\caption{A Comparison of C-Eval with Noise in Gradients.}
\label{tab:ceval_noise_full}
\resizebox{1.0\linewidth}{!}{
\begin{tabular}{ccccccc} 
\toprule
Methods & Stem & Social Sciences & Humanities & Others & Average & Avg(hard) \\
\midrule
\textbf{Noised-Compress-FT-LLaMA $\sigma = 5 \cdot 10^{-1}$}	
& 24.8 & 24.8 & 24.2 & 24.5 & 24.6 & 25.7 \\
\textbf{Noised-Compress-FT-LLaMA $\sigma = 5 \cdot 10^{-2}$} 
& 24.5 & 24.9 & 24.1 & 24.4 & 24.5 & 26.1 \\
\textbf{Noised-Compress-FT-LLaMA $\sigma = 5 \cdot  10^{-3}$} 
& 24.1 & 26.5 & 24.0 & 24.2 & 24.6 & 25.3 \\
\textbf{Noised-Compress-FT-LLaMA $\sigma = 5 \cdot 10^{-4}$} 
& \textbf{26.8} & 25.3 & \textbf{26.4} & \textbf{26.6} & \textbf{26.4} & \textbf{27.2} \\
\textbf{Noised-Compress-FT-LLaMA $\sigma = 5 \cdot 10^{-5}$} 
& 26.6 & 26.0 & 26.0  & 25.7 &26.3 & 26.8 \\
\midrule
\textbf{Noised-LoRA-FT-LLaMA $\sigma = 5 \cdot 10^{-2}$*} 
& 24.4 & 24.5 & 23.5 & 24.5 & 24.3 & 24.6 \\
\textbf{Noised-LoRA-FT-LLaMA $\sigma = 5 \cdot 10^{-3}$} 
& 21.8 & 23.3 & 23.6 & 23.3 & 22.8 & 20.2 \\
\textbf{Noised-LoRA-FT-LLaMA $\sigma = 5 \cdot 10^{-4}$} 
& 23.6 & 25.1 & 25.1 & 24.3 & 24.4 & 21.6\\
\textbf{Noised-LoRA-FT-LLaMA $\sigma = 5 \cdot 10^{-5}$} 
& 24.3 & 25.6 & 24 & 23.6 & 24.3 & 24.1\\
\midrule 
\textbf{Cent-LLaMA} 
& 24.5 & 25.6 & 25.5 & 24.4 & 24.9 & 23.4 \\ 
\textbf{Compress-FT-LLaMA} (ours) 
& 26.6 & 26.5 & 25.5 & 25.7 & 26.2 &  26.8 \\
\textbf{LoRA-FT-LLaMA} 
& 25.9 & \textbf{27.6} & 25.2 & 24.5 & 25.8 & 24.8 \\
\textbf{Base-LLaMA} 
& 21.6 & 23.4 & 23.9 & 23.3 & 22.8 & 20.3 \\ 
\bottomrule
\end{tabular}}
\end{table*}

\begin{table*}[t]
  \centering 
  \caption{Denoising Effect and SNR Analysis for \textbf{Noised-Compress}.}
  \label{tab:snr_noised}
  \begin{tabular}{cccccc}
    \toprule
 $\sigma$ &$\|[\mathbf{A}, \mathbf{B}]\|^2_2$ 
 & \textbf{LoRA-FT} MSE 
 & \textbf{LoRA-FT} SNR
 & \textbf{Compress-FT} MSE 
 & \textbf{Compress-FT} SNR\\
    \midrule
$5 \cdot 10^{-5}$ & 14.29 & $2.50 \cdot 10^{-9}$ & $5.72 \cdot 10^{9}$ & $4.59 \cdot 10^{-7}$ & $3.11 \cdot 10^{7}$  \\
$5 \cdot 10^{-4}$ & 14.29 & $2.50 \cdot 10^{-7}$ & $5.72 \cdot 10^{7}$ & $5.12 \cdot 10^{-7}$ & $2.79 \cdot 10^{7}$  \\
$5 \cdot 10^{-3}$ & 14.29 & $2.50 \cdot 10^{-5}$ & $5.72 \cdot 10^{5}$& $5.25 \cdot 10^{-7}$ & $2.72 \cdot 10^{7}$   \\
$5 \cdot 10^{-2}$ & 14.29 & $2.50 \cdot 10^{-3}$ & $5.72 \cdot 10^{3}$& $5.24 \cdot 10^{-7}$ & $2.73 \cdot 10^{7} $\\
$5 \cdot 10^{-1}$ & 14.29 & $2.50 \cdot 10^{-1}$ & $5.72 \cdot 10^{1}$& $5.21 \cdot 10^{-7}$ & $2.74 \cdot 10^{7}$ \\ 
  \bottomrule
\end{tabular}
\end{table*}

\begin{figure*}[t!]
     \begin{subfigure}[b]{0.50\textwidth}
         \centering
         \includegraphics[width=\textwidth]{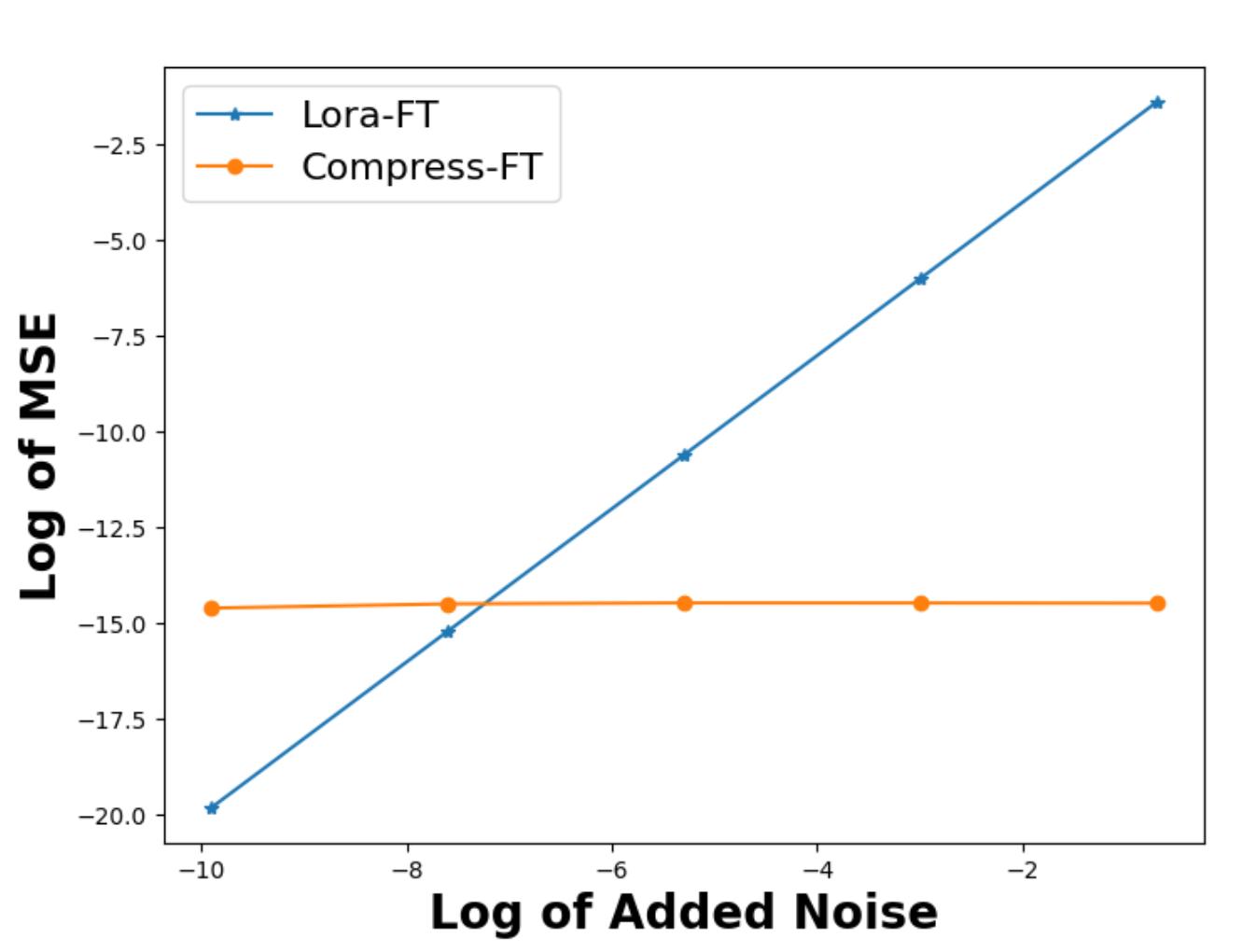}
     \end{subfigure}
     \begin{subfigure}[b]{0.50\textwidth}
         \centering
         \includegraphics[width=\textwidth]{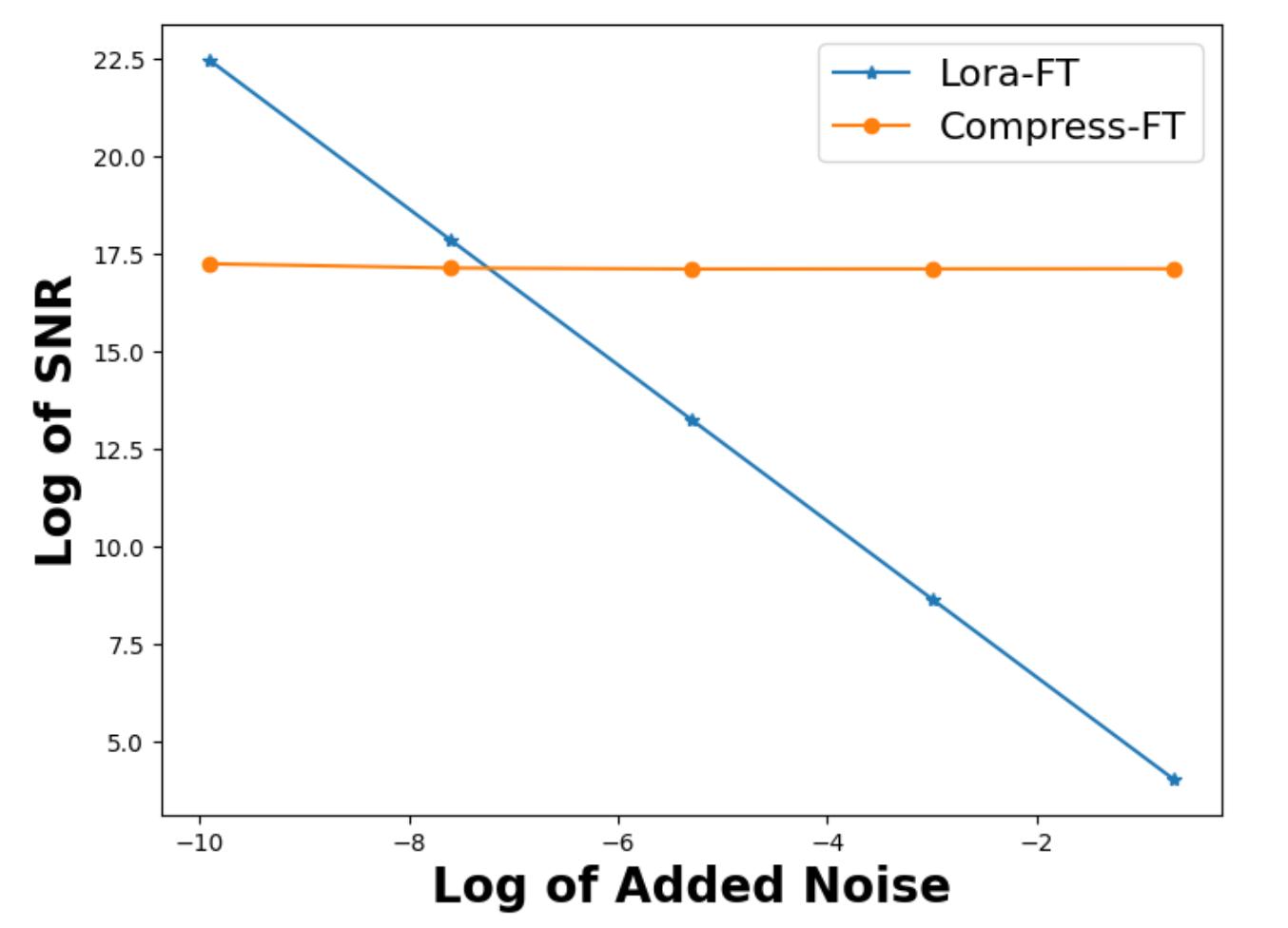}
     \end{subfigure}
\caption{Logarithmic MSE (on the left) and Logarithmic SNR (on the right) for \textbf{LoRA-FT} (in blue) and \textbf{Compress-FT} (in orange). }
\label{fig:noised_snr}
\end{figure*}

\subsection{Noising Effects of \textbf{Compress-FT} and \textbf{LoRA-FT}}
\label{sec:append_noised}

In this section, we demonstrate the denoising effect of the AutoEncoder pre-trained in \stageone. 
In \stagetwo~, in order to provide privacy guarantee, we add Gaussian noise on low-rank matrices $\mathbf{A}$'s and $\mathbf{B}$'s as a preprocess for the compression procedure. 
According to the differential privacy definition, Gaussian noise is added to the low-rank decomposition with respect to the sensitivity of the gradients. 
We first clip the local gradients to make the sensitivity equal 1.
Then we add random noise with magnitude $\sigma$ calculated by Gaussian Differential Privacy (GDP)~\cite{dong2022gaussian}.
\begin{align*}
\tilde{\mathbf{A}} &= \mathbf{A} + \mathbf{n}_A, \quad \mathbf{n}_A \sim \mathcal{N}(0, \sigma I), \\
\tilde{\mathbf{B}} &= \mathbf{B} + \mathbf{n}_B, \quad \mathbf{n}_B \sim \mathcal{N}(0, \sigma I);
\end{align*}
where $\sigma$ is the magnitude of added noise on low rank matrices and $I$ is the identity matrix representing the isotrophic noise distribution and $\mathbf{n}_A$ and $\mathbf{n}_B$ are the random matrices with the same shape as matrices $\mathbf{A}$ and $\mathbf{B}$. 
Once the low-rank matrices with added noise are prepared on the client side, they are transmitted to the server using \textbf{LoRA-FT}. In contrast, for \textbf{Compress-FT}, the noised matrices are first compressed using an AutoEncoder before being sent to the server. 
We tested the influence of different noise magnitudes $\sigma$ on the fine-tuned model performance with $\sigma$ ranging from $5 \cdot 10^{-5}$ to $5 \cdot 10^{-1}$ for \textbf{Compress-FT} and \textbf{LoRA-FT} respectively in Table~\ref{tab:ceval_noise_full}.
In the first part, we present the model performance with noise on \textbf{Compress-FT}.
In the second part, we show the model performance with noise on \textbf{LoRA-FT}. 
In the last part, we show the model performance without noise for comparison. 
In the case of \textbf{LoRA-FT} with a noise level of $5 \cdot 10^{-2}$, NaN values were encountered during the fine-tuning process in the fifth epoch, leading us to evaluate the model performance in Epoch 4. For the same noise level in \textbf{LoRA-FT}, NaN values appeared in the first epoch, preventing any performance evaluation of this model. All other models were tested in Epoch 19. According to Table~\ref{tab:ceval_noise_full}, noise significantly impacts the performance of \textbf{LoRA-FT}, correlating higher noise levels with poorer model outcomes. Conversely, \textbf{Compress-FT} shows minimal impact from the added Gaussian noise. Remarkably, a lower noise level of $\sigma = 5 \cdot 10^{-4}$ in \textbf{Compress-FT} not only mitigates the negative effects but also enhances performance compared to the noise-free scenario. This improvement is attributed to the AutoEncoder's capacity to manage abnormal gradients and the slight noise helping to avoid local minima, thus enhancing overall model effectiveness. Ultimately, our experiments demonstrate the effectiveness of incrementally increasing noise levels on low-rank matrices, underscoring the advantages of our approaches.

\subsection{Denoising Effects of AutoEncoder}
\label{sec:append_snr}

We now examine the denoising capabilities of the trained AutoEncoder to demonstrate the advantages of our \textbf{Compress-FT} method over \textbf{LoRA-FT} in noisy experiments, as discussed in Section~\ref{sec:append_noised}. 
$[\mathbf{A}, \mathbf{B}]$ are the original low-rank gradients. 
$[\tilde{\mathbf{A}}, \tilde{\mathbf{B}}]$ are the perturbed low-rank gradients with Gaussian noise of magnitude $\sigma = 5 \cdot [10^{-1}, 10^{-2}, 10^{-3}, 10^{-4}, 10^{-5}]$. 
$[\bar{\mathbf{A}}, \bar{\mathbf{B}}] = \auto([\tilde{\mathbf{A}}, \tilde{\mathbf{B}}])$ are the output of AutoEncoder with the input of perturbed low-rank gradients. 
Adding Gaussian noise proportional to the $\ell_2$ gradient norm is a common strategy to guarantee differential privacy in federated training~\cite{wei2020federated}. 
We increase the magnitude of added Gaussian noise from $5 \cdot 10^{-5}$ to $5 \cdot 10^{-1}$ to collect the SNR computed by \textbf{LoRA-FT} and \textbf{Compress-FT}. 
The primary rationale is that processing the noisy gradients $\tilde{\mathbf{A}}$ and $\tilde{\mathbf{B}}$ allows the removal of Gaussian noise and the restoration of the original, noise-free local updates $\mathbf{A}$ and $\mathbf{B}$. Table~\ref{tab:snr_noised} presents the denoising results of the proposed AutoEncoder equipped with three ResNet blocks and includes the calculation of the SNR.
For both MSE loss and SNR, \textbf{LoRA} shows an linear change with the magnitude of Gaussian noise $\sigma$, while our method \textbf{Compress-FT} remains independent of the magnitude of added Gaussian noise, which affirms the strong denoising effect of the proposed AutoEncoder.
We also show the MSE loss of \textbf{LoRA-FT} and \textbf{Compress-FT} in Figure~\ref{fig:noised_snr}.

\section{Surface Visualization of Low-rank Gradients}

We introduce the matrix visualizations of $\mathbf{B} \mathbf{A}$ during the federated fine-tuning of the \textbf{LLaMA-7B} model in conjunction with the \textbf{Databricks-olly-15k}. At the midpoint of the fine-tuning phase, we observe a slight change in the updated model parameters, roughly $10^{-3}$. The reconstruction loss was estimated to be around $10^{-7}$. The significant difference in the magnitudes of these two measurements highlights the success of our integrated compression methods, \stageone~and \stagetwo.
\begin{figure*}[htbp]
     \begin{subfigure}[b]{0.25\textwidth}
         \centering
         \includegraphics[width=\textwidth]{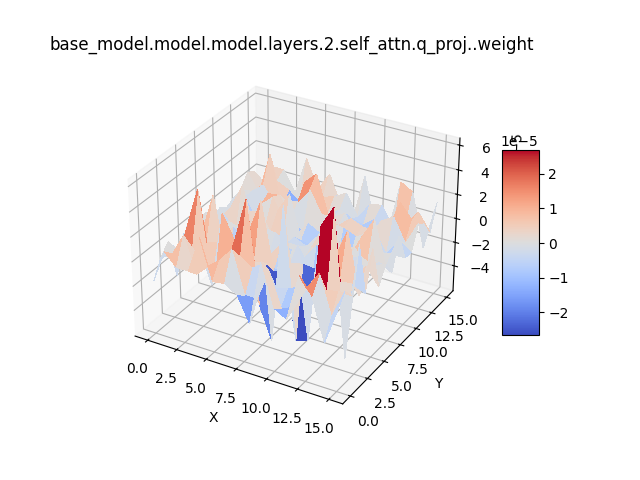}
     \end{subfigure}
     \begin{subfigure}[b]{0.245\textwidth}
         \centering
         \includegraphics[width=\textwidth]{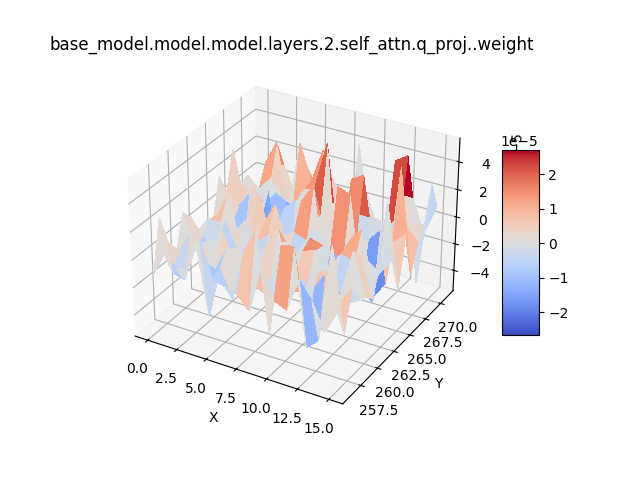}
     \end{subfigure}
     \begin{subfigure}[b]{0.245\textwidth}
         \centering
         \includegraphics[width=\textwidth]{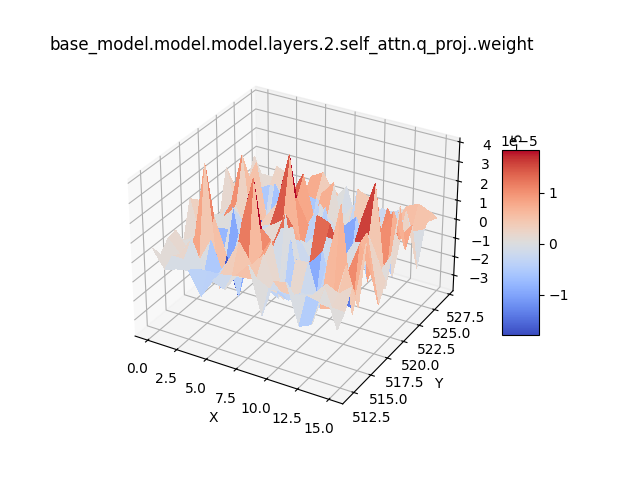}
     \end{subfigure}
     \begin{subfigure}[b]{0.245\textwidth}
         \centering
         \includegraphics[width=\textwidth]{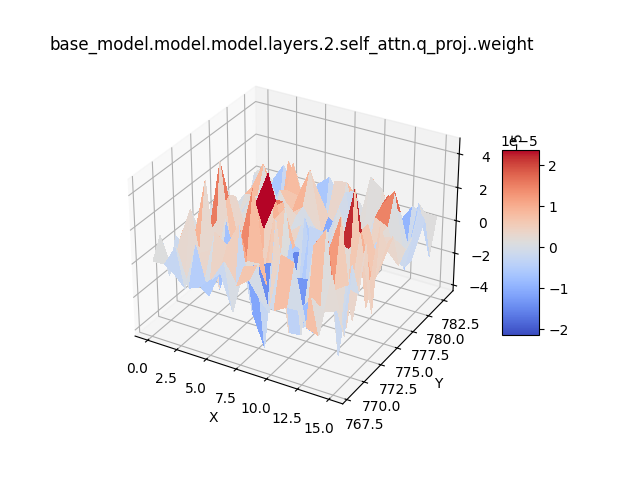}
     \end{subfigure}
     \begin{subfigure}[b]{0.25\textwidth}
         \centering
         \includegraphics[width=\textwidth]{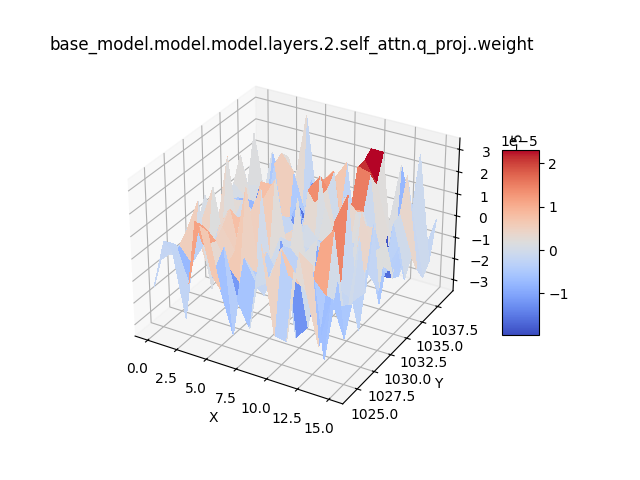}
     \end{subfigure}
     \begin{subfigure}[b]{0.245\textwidth}
         \centering
         \includegraphics[width=\textwidth]{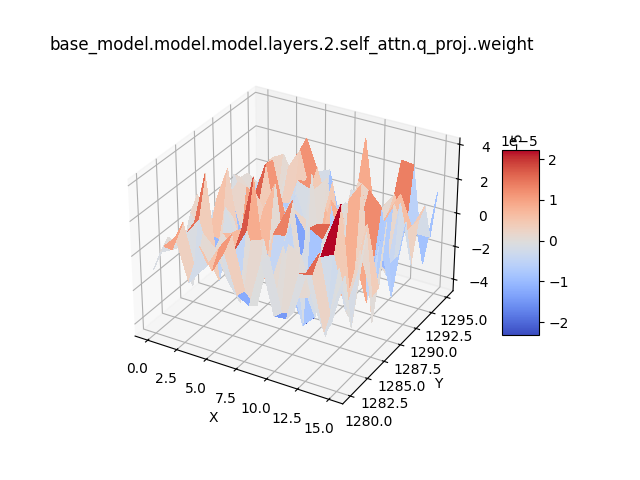}
     \end{subfigure}
     \begin{subfigure}[b]{0.245\textwidth}
         \centering
         \includegraphics[width=\textwidth]{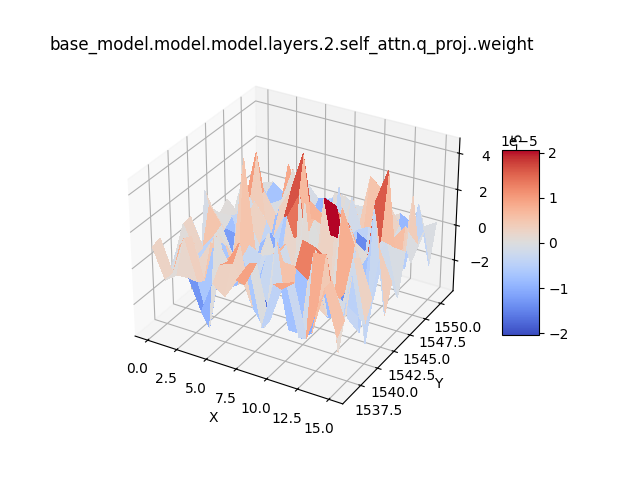}
     \end{subfigure}
     \begin{subfigure}[b]{0.245\textwidth}
         \centering
         \includegraphics[width=\textwidth]{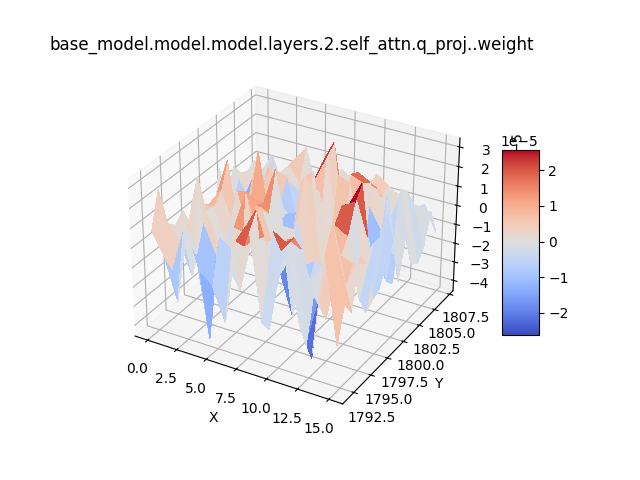}
     \end{subfigure}
     \begin{subfigure}[b]{0.25\textwidth}
         \centering
         \includegraphics[width=\textwidth]{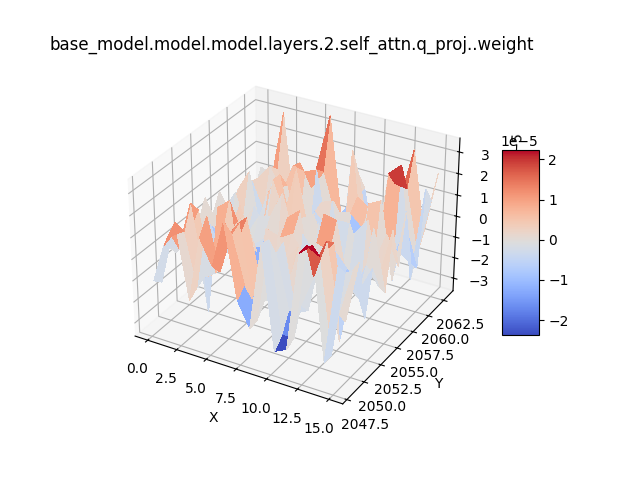}
     \end{subfigure}
     \begin{subfigure}[b]{0.245\textwidth}
         \centering
         \includegraphics[width=\textwidth]{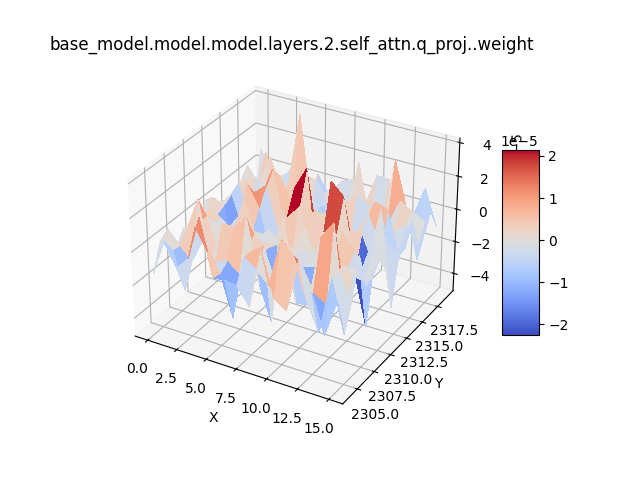}
     \end{subfigure}
     \begin{subfigure}[b]{0.245\textwidth}
         \centering
         \includegraphics[width=\textwidth]{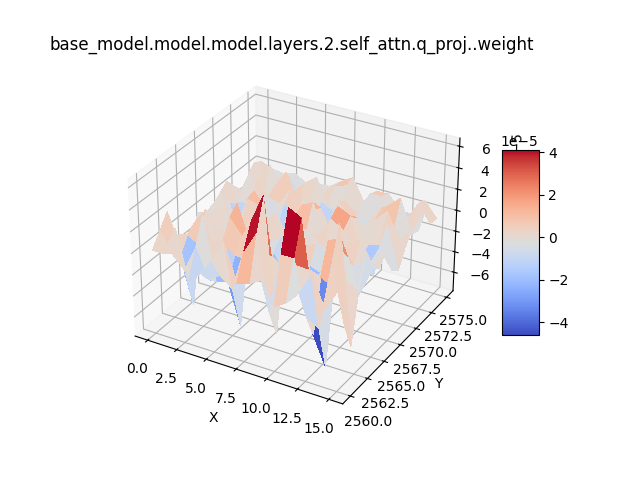}
     \end{subfigure}
     \begin{subfigure}[b]{0.245\textwidth}
         \centering
         \includegraphics[width=\textwidth]{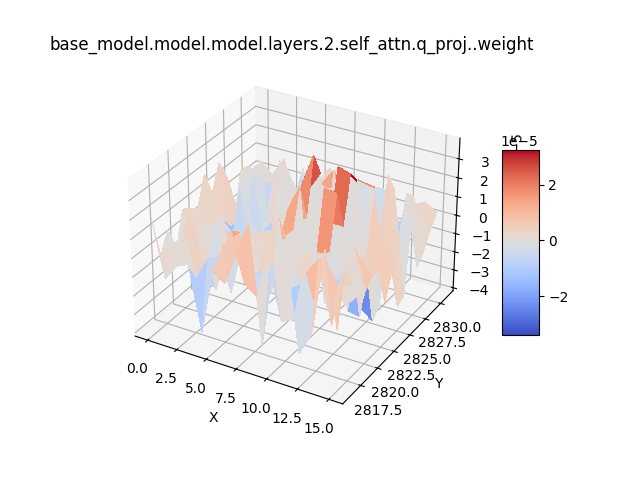}
     \end{subfigure}
     \begin{subfigure}[b]{0.25\textwidth}
         \centering
         \includegraphics[width=\textwidth]{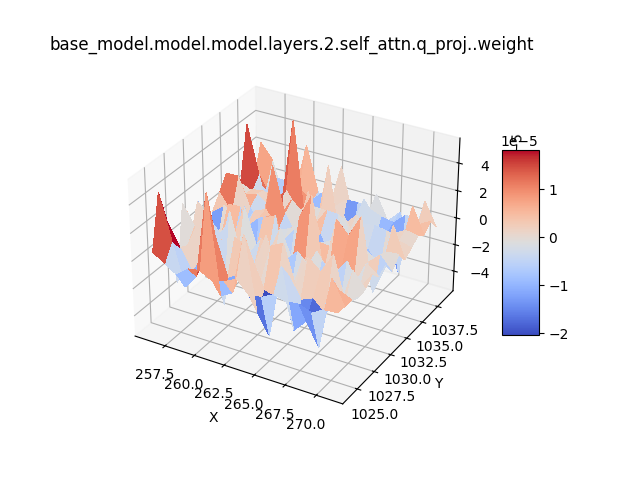}
     \end{subfigure}
     \begin{subfigure}[b]{0.245\textwidth}
         \centering
         \includegraphics[width=\textwidth]{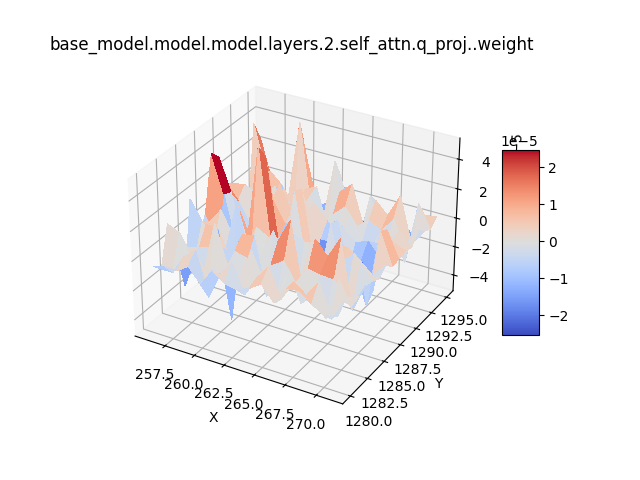}
     \end{subfigure}
     \begin{subfigure}[b]{0.245\textwidth}
         \centering
         \includegraphics[width=\textwidth]{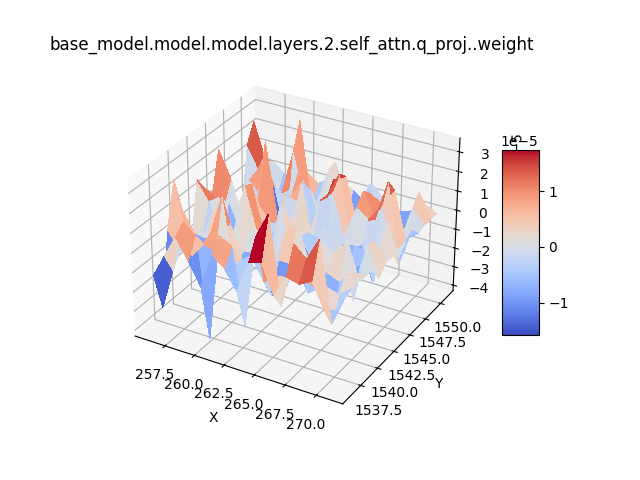}
     \end{subfigure}
     \begin{subfigure}[b]{0.245\textwidth}
         \centering
         \includegraphics[width=\textwidth]{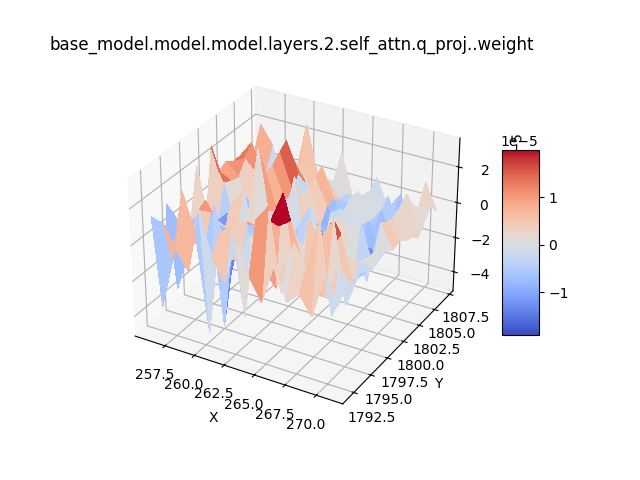}
     \end{subfigure}
     \begin{subfigure}[b]{0.25\textwidth}
         \centering
         \includegraphics[width=\textwidth]{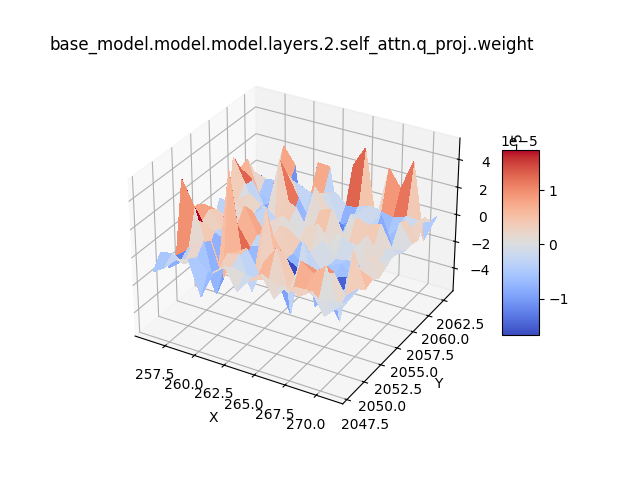}
     \end{subfigure}
     \begin{subfigure}[b]{0.245\textwidth}
         \centering
         \includegraphics[width=\textwidth]{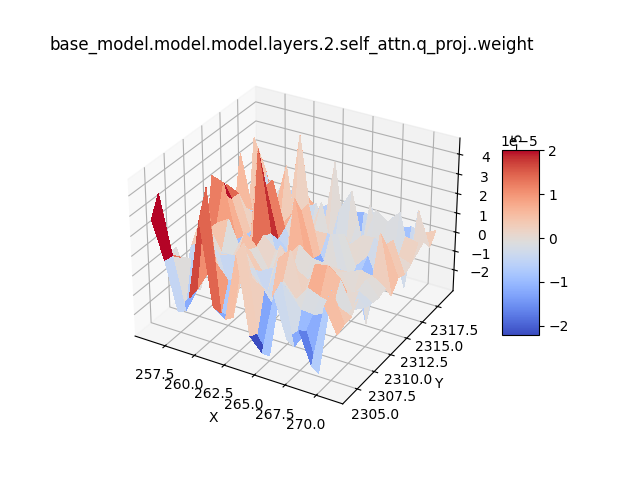}
     \end{subfigure}
     \begin{subfigure}[b]{0.245\textwidth}
         \centering
         \includegraphics[width=\textwidth]{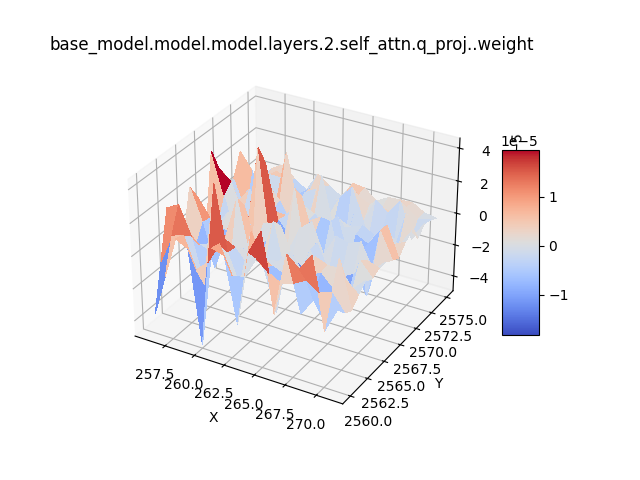}
     \end{subfigure}
     \begin{subfigure}[b]{0.245\textwidth}
         \centering
         \includegraphics[width=\textwidth]{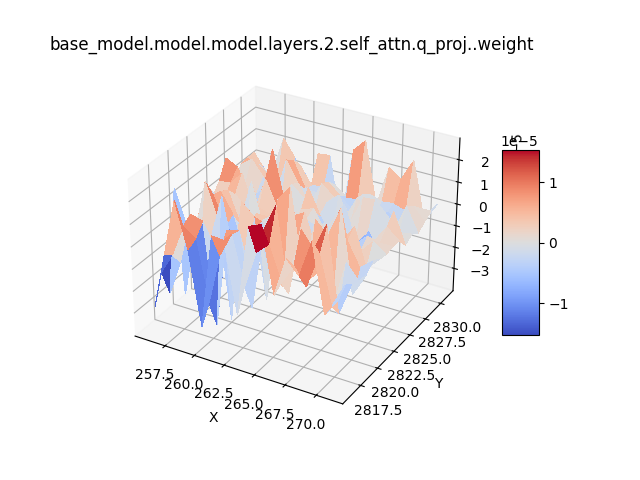}
     \end{subfigure}
     \begin{subfigure}[b]{0.25\textwidth}
         \centering
         \includegraphics[width=\textwidth]{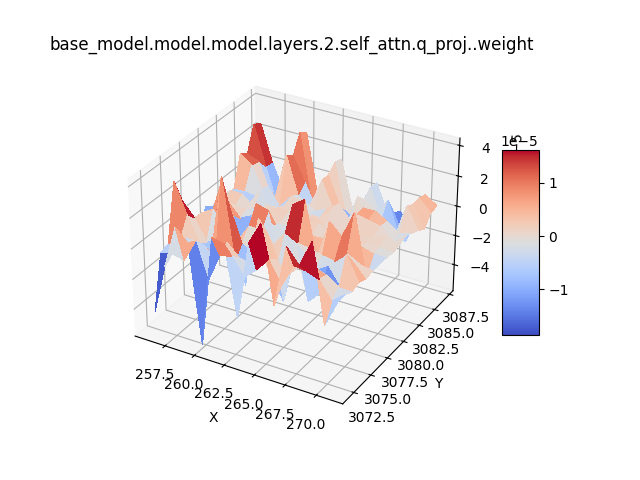}
     \end{subfigure}
     \begin{subfigure}[b]{0.245\textwidth}
         \centering
         \includegraphics[width=\textwidth]{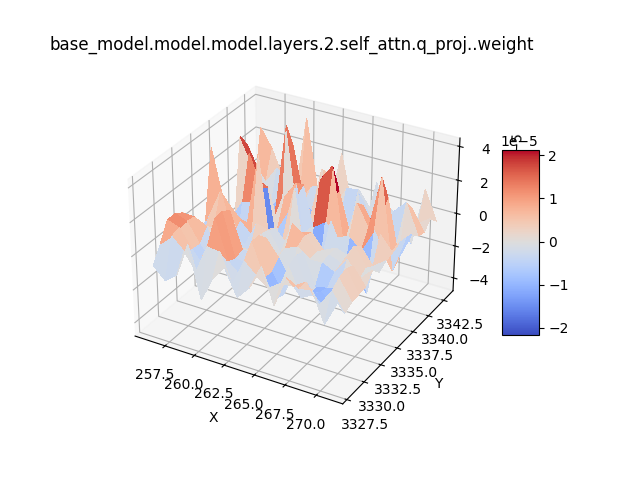}
     \end{subfigure}
     \begin{subfigure}[b]{0.245\textwidth}
         \centering
         \includegraphics[width=\textwidth]{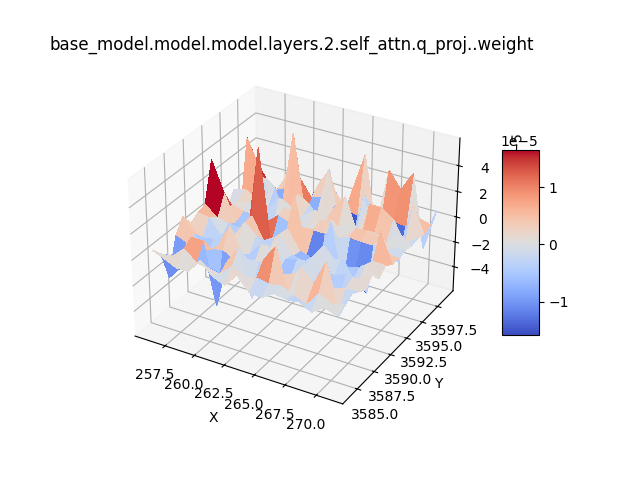}
     \end{subfigure}
     \begin{subfigure}[b]{0.245\textwidth}
         \centering
         \includegraphics[width=\textwidth]{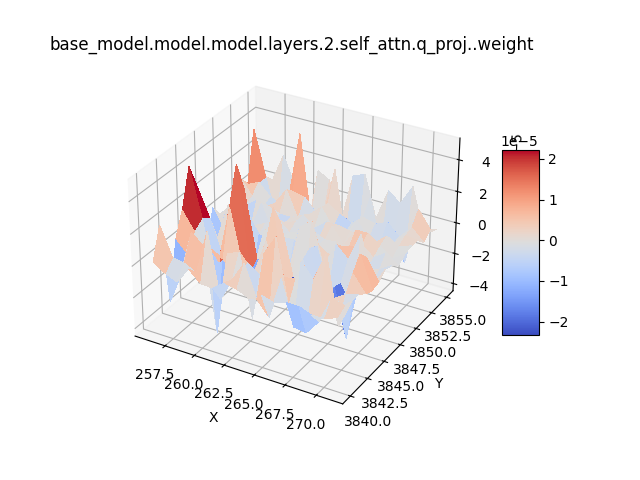}
     \end{subfigure}
\caption{The visualization of $\mathbf{B} \mathbf{A}$ with the low-rank decomposition of $\mathbf{Q}$ for layers $\{0,1,2,29,30,31\}$.}
\label{fig:Q_vis}
\end{figure*}

\end{document}